\documentclass[5p,times,10pt,authoryear]{elsarticle}
\usepackage{color}
\usepackage{times}
\usepackage{epsfig}
\usepackage{graphicx}
\usepackage{amsmath}
\usepackage{amssymb}
\usepackage{array}
\usepackage{soul}
\usepackage[ruled]{algorithm2e}
\usepackage{multirow}
\usepackage[noend]{algpseudocode}

\usepackage{amssymb}

\journal{Computer Vision and Image Understanding}

\begin{document}

\begin{frontmatter}



\title{Learning Deep Representations for Scene Labeling with Semantic Context Guided Supervision}


\author{Zhe~Wang,
        Hongsheng~Li, Wanli~Ouyang,
        and~Xiaogang~Wang}

\address{Department of Electronic Engineering, The Chinese University of Hong Kong, Shatin, New Territories, Hong Kong}
\address{\{zwang, hsli, wlouyang, xgwang\}@ee.cuhk.edu.hk}

\begin{abstract}
Scene labeling is a challenging classification problem where each input image requires a pixel-level prediction map. Recently, deep-learning-based methods have shown their effectiveness on solving this problem. However, we argue that the large intra-class variation provides ambiguous training information and hinders the deep models' ability to learn more discriminative deep feature representations. Unlike existing methods that mainly utilize semantic context for regularizing or smoothing the prediction map, we design novel supervisions from semantic context for learning better deep feature representations. Two types of semantic context, scene names of images and label map statistics of image patches, are exploited to create label hierarchies between the original classes and newly created subclasses as the learning supervisions. Such subclasses show lower intra-class variation, and help CNN detect more meaningful visual patterns and learn more effective deep features. Novel training strategies and network structure that take advantages of such label hierarchies are introduced. Our proposed method is evaluated extensively  on four popular datasets, Stanford Background (8 classes),  SIFTFlow (33 classes), Barcelona (170 classes) and LM+Sun datasets (232 classes) with 3 different networks structures, and show state-of-the-art performance. The experiments show that our proposed method makes deep models learn more discriminative feature representations without increasing model size or complexity.
\end{abstract}


\end{frontmatter}


\section{Introduction}
The task of scene labeling is to densely predict every pixel of an image into one of the pre-defined classes. One of the most popular ways is to take the image patch around the pixel of interest as input for a learning system. However, large intra-class variation, i.e., the variation of samples within the same class, is one of the key challenges in scene labeling. It is caused by various factors such as the change of illumination, posture, scale, viewpoint and background, and these factors are coupled together in images in nonlinear ways. It is difficult to design features that can be used to well classify samples with significantly different appearances into the same class. 

Recent works \citep{yang2014context, farabet2013learning, he2004multiscale, shotton2006textonboost, pinheiro2013recurrent, socher2011parsing} have shown that the performance of scene labeling can be improved by effectively exploiting rich contextual information, and image context is the most commonly used one, i.e., training classifiers that take large image patches as input and predicting class labels of pixels. It is shown that deep models which have large receptive fields contain more image contextual information and generally lead to higher segmentation accuracy \citep{pinheiro2013recurrent, farabet2013learning}. Semantic contextual information, on the other hand, is usually utilized to regularize or smooth the predicted label maps as a post-processing step \citep{he2004multiscale,liu2011nonparametric} or as final neural network layers in an end-to-end learning system \citep{zheng2015conditional,liu2015semantic}.

\begin{figure}[t]
\centering
   \includegraphics[width=\linewidth]{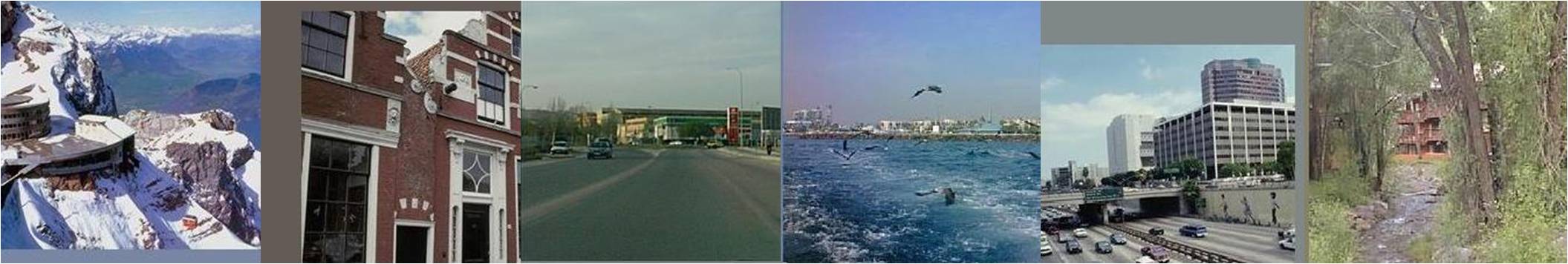}\\
   (a) ``Building'' samples with large appearance differences
 \includegraphics[width=\linewidth]{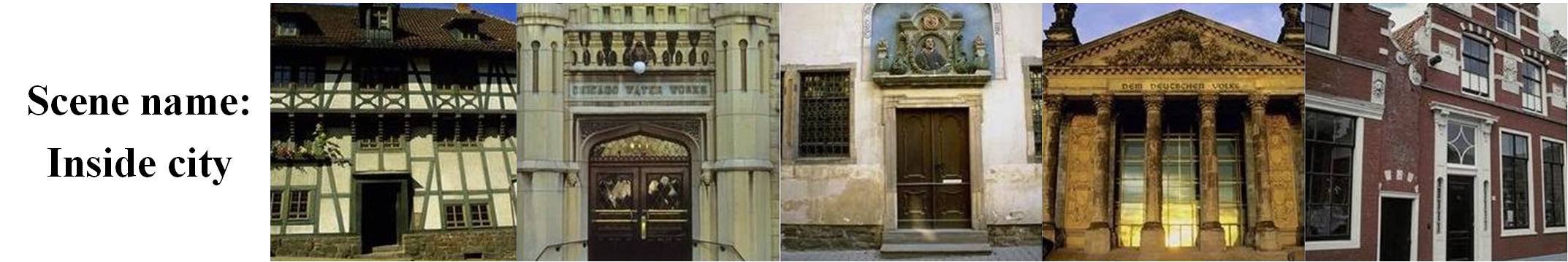}\\
 \includegraphics[width=\linewidth]{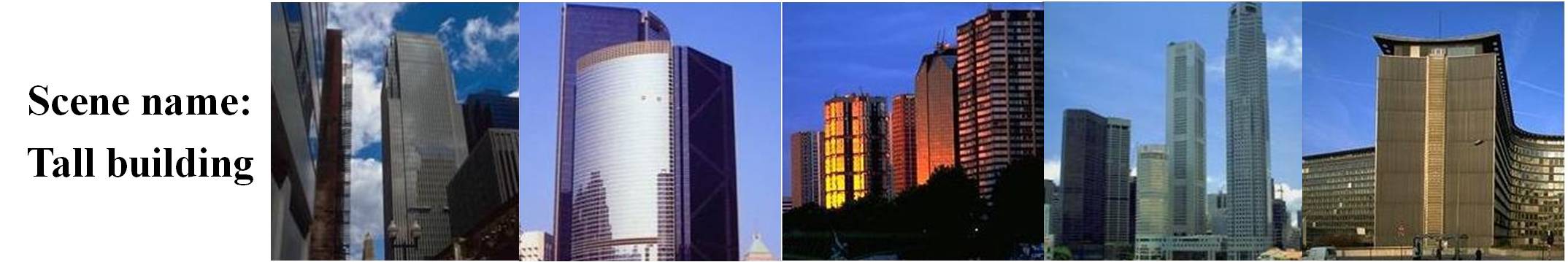}\\
         (b) Subclasses of ``building'' based on scene names
   \includegraphics[width=\linewidth]{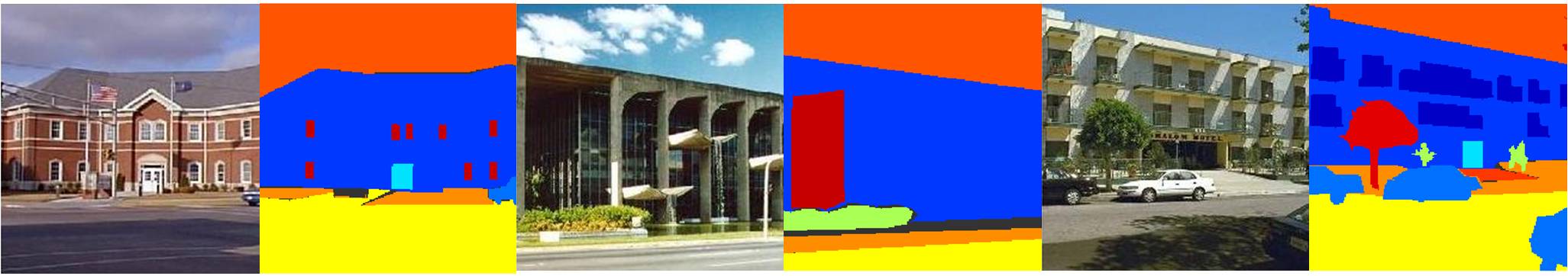}        
      \includegraphics[width=\linewidth]{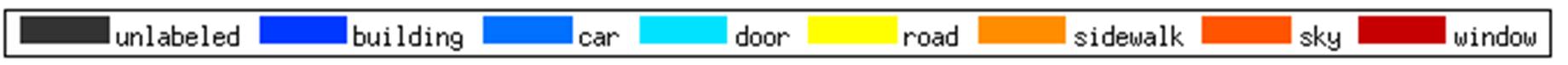}       
   \includegraphics[width=\linewidth]{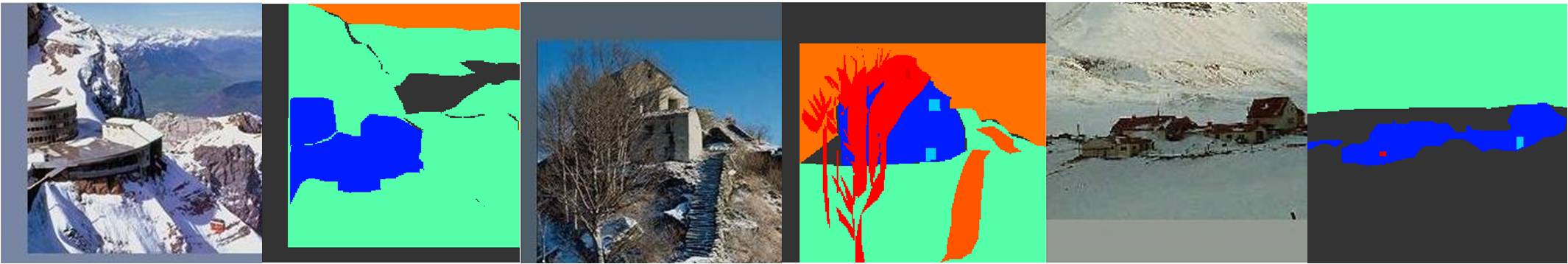}
         \includegraphics[width=\linewidth]{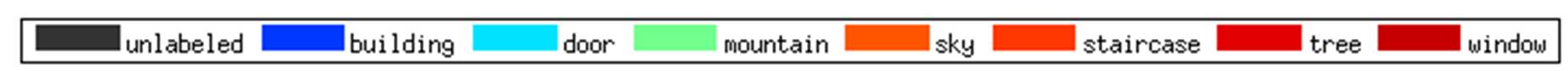}     \\
         (c) Subclasses of ``building'' based on label maps
   \caption{(a) Training image patches of the class "building" from the SIFTFlow dataset exhibit large apperance differences. (b) Based on scene names (top) ``inside city'' and (bottom) ``tall building'', subclasses of ``building'' with lower intra-class variation can be created. (c) Based on label map statistics, subclasses with lower intra-class variation can be created to represent ``buildings by the road'' and ``buildings in the mountains''. Uniformly-grey-colored areas represent padded mean pixels.}\label{fig:intro}
\end{figure} 

Limited by the learning capacity, conventional learning models such as SVM cannot effectively utilize useful information from large image context. In recent years, significant research progress \citep{grangier2009deep,farabet2013learning} on scene labeling has been achieved by using deep Convolutional Neural Networks (CNNs). It has been discovered that CNN pre-trained on a large-scale dataset such as the ImageNet has good generalization ability \citep{rirshick2014rich, yosinski2014transferable}. It provides a good initial point and can be further fine-tuned on the scene labeling datasets, being adapted to the new task. 

The success of CNN relies on its capability of learning highly discriminative deep feature representations from training image patches. Once the feature representations are effectively learned, even a simple linear classifier can well predict class labels based on that. Its neurons at different levels serve as detectors of various visual patterns at different scales. However, current deep learning approaches for scene labeling still face the challenge of large intra-class variation, especially at the stage of feature learning. As shown in Fig. \ref{fig:intro}(a), all the image patches with large diversity in appearance are assigned to the class ``building'' according to the labels of their centered pixels. It raises a lot of ambiguity, which confuses CNN when the weights of neurons are adjusted to detect meaningful visual patterns. It
would make the feature learning process much easier if the
training patches could be grouped into subclasses with better
visual consistency. However, more supervised information is needed in order to obtain such subclasses.  One could argue that such subclasses can be obtained by clustering image context. However, it requires discriminative feature representations learned by deep models in order to achieve satisfactory clustering results, which leads to a chicken-and-egg problem. 

State-of-the-art deep learning methods \citep{farabet2013learning,chen2014semantic} focus on using semantic context for smoothing or regularizing the predicted label maps, while ignoring other rich semantic context available in the datasets. In this paper, we exploit using two types of semantic context: scene names of images and label map statistics of image patches, as supervision signals for learning deep feature representations. In some scene labeling datasets, each image has a name which indicates the scene category. 
As shown in Fig. \ref{fig:intro}(b), image patches that have the same class label and are from the same scene category are likely to have similar appearance.
On the other hand, the label map statistics of image patches also provides crucial prior information on the appearance of patches, since they specify the spatial layout of the semantics in the surrounding region. As shown by the examples in Fig. \ref{fig:intro}(c), image patches with similar label map statistics show consistent appearance. Although all these patches belong to the class ``building'', the patches in the top row of Fig. \ref{fig:intro}(c) capture buildings by the road with sky on the top, while those patches in the bottom row contain buildings in the mountains. Such distinction can be well reflected by label map statistics.

In this paper, we create a two-level label hierarchy for each of the original classes by exploiting semantic context, and CNN is fine-tuned with the proposed label hierarchies. The deep feature representations learned in this way are more discriminative and can better predict the original classes. 
 Extensive experiments have shown the effectiveness of the proposed label hierarchy and training schemes on four datasets: SIFTFlow \citep{liu2008sift}, Stanford background \citep{gould2009decomposing}, Barcelona \citep{tighe2010superparsing} and LM+Sun \citep{tighe2010superparsing_jnl} datasets. Our proposed approaches outperform or achieve the state-of-the-art accuracies on all four datasets.

Our contributions can be summarized as three-fold:
\begin{itemize}
\item
We demonstrate the effectiveness of learning deep feature representations in scene labeling by creating label hierarchies from semantic context as the strong supervision. This useful information in semantic context was not explored in previous deep learning works for scene labeling.

\item
Two approaches are proposed to generate fine-grained subclasses based on scene names and clustering of label maps, respectively. Label hierarchy is built between the original classes and the new subclasses. Such label hierarchies decrease intra-subclass variation and is a perfect match to CNNs, whose loss functions are minimized via local optimization methods.

\item
To learn deep representations based on the proposed label hierarchy, we develop two distinct approaches to fine-tune an ImageNet pre-trained CNN. The first one fine-tunes the CNN on the newly created subclasses and the original classes sequentially. The second builds a novel hierarchical architecture for the CNN to optimize both the classification of the subclasses and original classes.
\end{itemize}

\section{Related Work}

\subsection{Scene labeling methods}
Current methods for scene labeling can be generally divided into two categories: parametric and non-parametric methods, depending on whether the categories are used for training in advance. 

\subsubsection{Parametric methods}
Parametric methods take image patches or whole images as input to train classifiers.

\paragraph{Conventional methods} 
Conventional methods mainly depend on hand-crafted features extracted from image patches and train classifiers such as multilayer perceptron \citep{he2004multiscale} and joint boosting \citep{shotton2006textonboost}. The hand-crafted features are designed by human experience and might not be robust enough for the large number of different scene types.  Since the label of each pixel is predicted independently, the boundary region may not be accurately pinpointed and there may exist isolated pixels with wrong labels. 
Various techniques have been proposed to generate spatially smooth label maps based on semantic contextual information.
Markov Random Field (MRF), Conditional Random Field (CRF) and other graphical models \citep{he2004multiscale, shotton2006textonboost, barinova2010geometric, ladicky2010graph, yao2012describing} are widely used to model the joint probabilities of neighboring labels. 
\paragraph{Deep-learning-based methods}

Recently, deep-learning-based methods \citep{pinheiro2013recurrent, socher2011parsing, long2014fcn, farabet2013learning, sharma2015deep, eigen2015predicting, dai2015boxsup, noh2015learning, liu2015semantic, qi2015semantic} drew increasing attention from the computer vision community because of their great power to model complex decision boundaries and  ability to learn discriminative features that outperform hand-crafted ones. 

 \cite{farabet2013learning} trained a multiscale deep CNN to extract dense features that encode regions of different sizes centered on each pixel.  \cite{pinheiro2013recurrent} relied on a recurrent CNN that consists of a sequential series of networks sharing the same set of parameters. Each recurrent component takes the RGB image and the predictions of the previous component as input of the network. These patch-based methods are not very efficient when feeding overlapping patches to CNN. More recently, \cite{long2014fcn} proposed the fully convolutional network (FCN), which utilizes whole images as training samples and directly output the whole prediction maps.
To further regularize the prediction maps of FCN and utilize the semantic contextual information, \cite{chen2014semantic} fed the output of FCNs to a fully connected CRF as a post-processing step. \cite{zheng2015conditional} converted the fully connected CRFs as a recurrent neural network (RNN), which is achieved by modeling the mean-field approximation as a stack of CNN layers. The parameters of the CRF can then be jointly optimized with the CNN in an end-to-end manner.  \cite{liu2015semantic} modeled the spatial context between objects as a graphic model, which includes high-order relations and mixture of label contexts. \cite{shuai2015dag} introduced a directed acyclic graphic recurrent neural networks (DAG-RNNs) to model long-range contextual information in the image. \cite{wang2016learnable} learned semantic histogram features in networks to integrate the global context information into prediction.

Although these approaches also exploit rich information in label maps, they have notable distinctions with our work. 
These methods aim to model pairwise, higher-order or long-range relations between pixels, while we focus on creating finer learning supervision for each pixel from two types of semantic context, scene names of images and label map statistics of image patches. Such useful information in semantic context was not explored in existing deep-learning-based methods. Since the proposed learning supervision is provided for each pixel, it can be flexibly applied to both patch-based CNNs and FCNs.

\subsubsection{Non-parametric methods}
Another category of scene labeling algorithms is the non-parametric methods \citep{liu2008sift, liu2011nonparametric,tighe2010superparsing, yang2014context, singh2013nonparametric,george2015image}. These methods do not require training a classifier. Instead, they retrieve similar images from the training dataset and transfer their labels to the test image. 
Some methods \citep{yang2014context} require several rounds of image retrieval and superpixel classification.   Generally, they use the global context for retrieving similar images and the local context for matching the query patch to its gallerias. The non-parametric methods are more suitable for the scenarios  where the training data is updated frequently because they do not need retraining the classifier. 

\subsection{Learning deep feature representations}
It has been shown that training deep neural network is essentially learning discriminative feature representations. However, training the deep neural network is a difficult problem, especially when the network is very deep. 
Researchers \citep{martens2012training,sutskever2013importance,dauphin2014identifying} have proposed various algorithms for training deep neural networks. For example, \cite{dauphin2014identifying} proposed the saddle-free Newton methods which can rapidly escape high dimensional saddle points. Other works \citep{srivastava2014dropout,ioffe2015batch,he2015delving}  design specific layers that alleviate the overfitting problem.   An interesting direction \citep{szegedy2014going,srivastava2015training,he2015deep}  recently proposed is to add specific ``skip'' layers to improve the information flow, and show impressive performance int the ImageNet Challenge. Different from the above works, we study learning effective feature representations from a new perspective, without modifying the network structure nor adding extra training data. This is achieved by grouping the training patches into subclasses with better visual consistency and learn the features with the subclass labels.   The subclasses by clustering may introduce some noise. Deep learning with noisy labels have shown inspiring results in different applications \citep{xiao2015learning,joulin2015learning}. We observe that our method can also benefit from the new subclass labels.

\subsection{Building label hierarchies}
There were works that also explored label hierarchies. \cite{xie2015hyper} proposed two data augmentation approaches which requires external data as the hyper-class of the original fine-grained labled data. Unlike their methods, we do not require external data but create the label hierarchies solely based on the original dataset. \cite{yan2015hd} built label hierarchies by grouping confusing fine categories into the same coarse categories. A hierarchical deep CNN is trained to first separate easy categories using a coarse classifier. Then challenging classes are routed downstream to fine category classifiers. Our network also has a hierarchical architecture but do the prediction reversely. The fine-grained subclass categories are predicted first, followed by the prediction of the original categories.
To tackle the contour detection problem, \cite{lim2013sketch} defined a set of so-called ``sketch token'' classes by clustering patches extracted from the binary contour images. Then a random forest classifier is trained to predict probability a patch belongs to each sketch token class.

\section{Methods}
We solve the problem of learning deep feature representations for scene labeling by creating finer learning supervisions for training convolutional neural networks. The original classes are split into more meaningful subclasses to create the label hierarchies. It imposes finer supervision for feature learning by creating more fine-grained classes from semantic context. The intra-class variation problem is mitigated since the samples in each subclass have more consistent appearance. Besides, the hierarchical relation is used to further improve the performance.  
We then train convolutional neural networks with the created label hierarchies. 

\subsection{Creating label hierarchy from semantic context}\label{subscn:data_preparation}
\subsubsection{Semantic context from scene names}\label{subsubscn:name_based_cluster}
Some scene labeling datasets (such as SIFTFlow \citep{liu2008sift}) provide a scene name for each image, which describes the scene category and can be viewed as global semantic context for all the pixels in the image. Examples of scene names from the SIFTFlow dataset are ``forest'', ``inside-city'', and ``coast''. In our approach, the scene names are only used in the training stage, but not in test. 

\begin{figure}[t]
\centering
   \includegraphics[width=\linewidth]{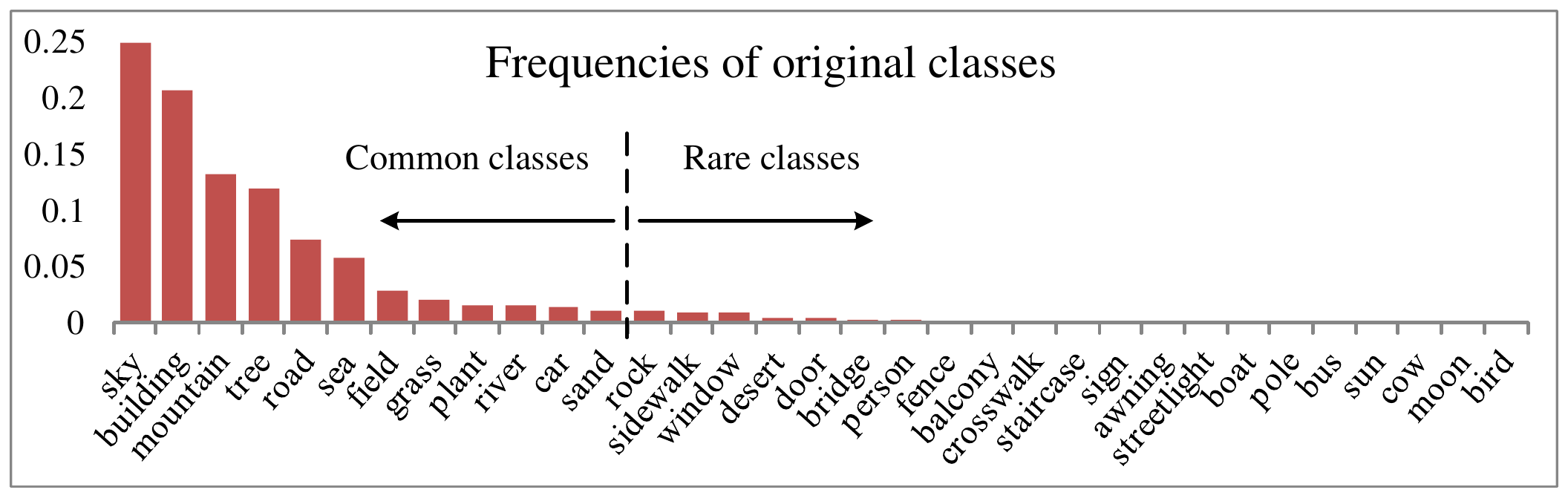}
   \includegraphics[width=\linewidth]{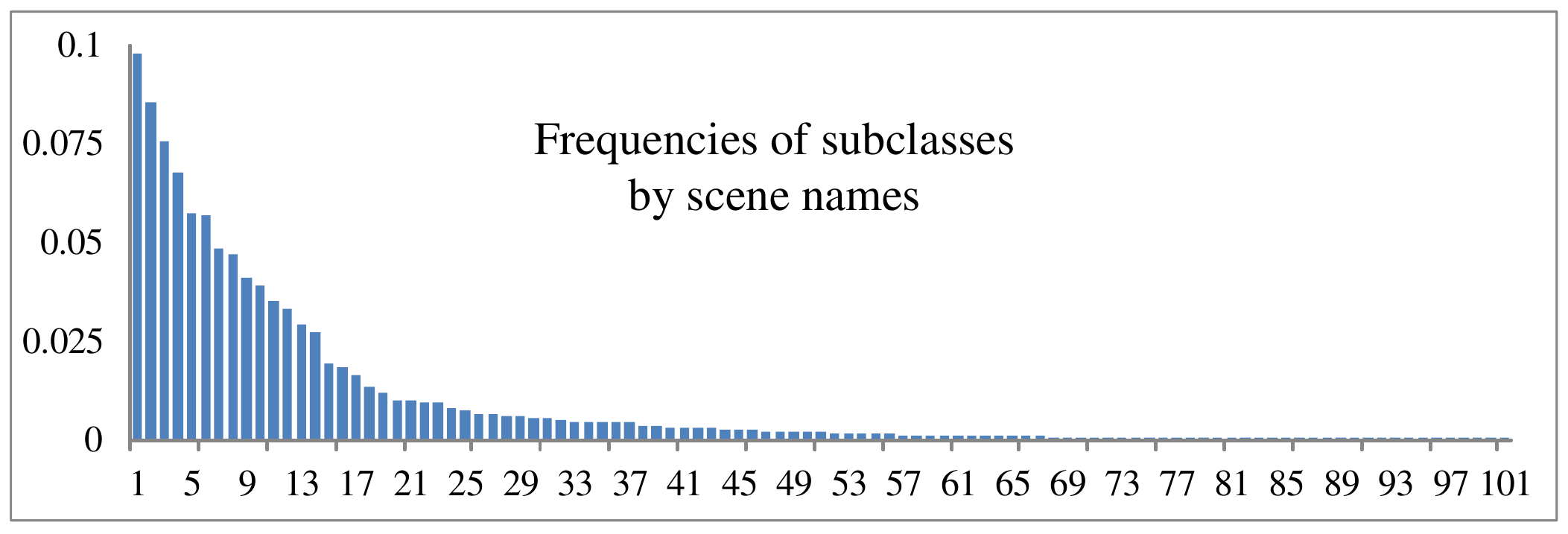}
   \includegraphics[width=\linewidth]{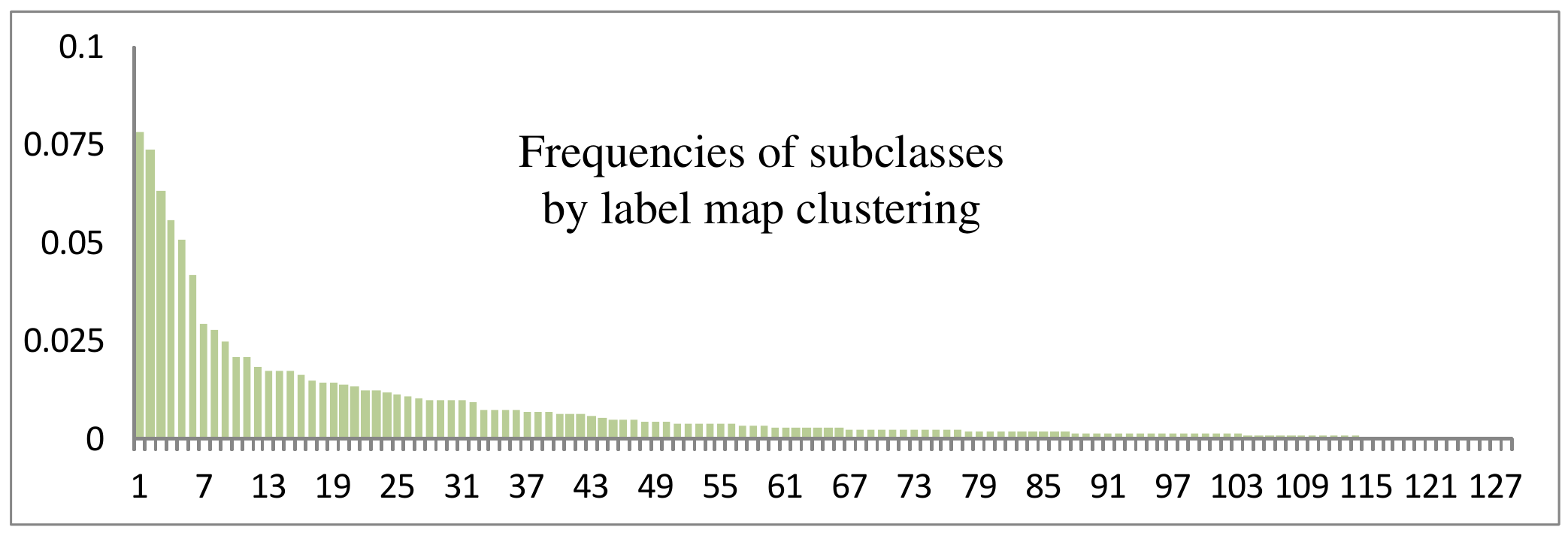}
   \caption{(Top) frequencies of original classes. With a threshold of $\rho=93\%$, the most frequent 11 classes can be chosen as common classes and the rest 22 classes are denoted as rare ones. (Middle) frequencies of 102 subclasses obtained based on scene names (Section \ref{subsubscn:name_based_cluster}). (Bottom) frequencies of 128 subclasses obtained via label map clustering (Section \ref{subsubscn:label_map_based_cluster}).}\label{fig:label_frequency}
\end{figure}

The provided scene names naturally split the images into $n_{name}$ subsets, where $n_{name}$ is the number of different scene names in the training dataset.
We introduce a simple method to create label hierarchy from class labels and scene names. Let $\left\lbrace x_i, l_i \right\rbrace$ be a training sample, where $x_i$ is a patch cropped from the image and $l_i \in \left\lbrace c_1,c_2,...,c_L \right\rbrace$ is its class label of the center pixel. 
 Each training sample  (image patch) now has an original class label and a subclass label. For instance, if an image patch with the original class label ``building'' is cropped from an image with the scene name ``highway'', then its subclass label is assigned as ``highway+building'', which can be intuitively regarded as ``a building patch in a highway scene image''.  
However, since the semantic labels of training samples usually follow a long-tail distribution, we do not split the original classes that have too few training samples. Similar to \cite{yang2014context}, we define rare classes by examining the superpixel distribution of the entire training set (see Fig. \ref{fig:label_frequency} (top) for the distribution of the SIFTFlow dataset.) We first sort the classes by their numbers of superpixels in a descending order, and then keep including the classes with most samples into the common classes until the total number of superpixels of the common classes becomes greater than a threshold $\rho$.  Only common classes are split to subclasses, while the rare classes directly inherit their original labels as their single subclass labels for consistency. 
Thus, for our first round of training with subclasses, we have $n_{SC}$ labels, where $L \leq n_{SC} \leq L\times n_{name}$.

The rare class ratio $\rho$ depends on the superpixel distribution of the training set but is not very sensitive. In our experiment we tried different $\rho$ values, but observe no significant differences on the final performance. (see Section \ref{subsec:siftflow_results})

In Figs. \ref{fig:label_frequency} (top) and \ref{fig:label_frequency} (middle), the frequencies of the original labels and new subclass labels in the SIFTFlow dataset are shown in a descending order. In the original label set, we specify 11 common classes with a threshold $\rho=93\%$. After the original classes are expanded to 101 subclasses by using 8 scene names provided by the dataset, we obtain a more balanced subclass set.

\subsubsection{Semantic context from label map statistics}\label{subsubscn:label_map_based_cluster}
\begin{figure}[t]
\centering
\begin{tabular}{c@{\hspace{-2mm}}c@{\hspace{1mm}}c@{\hspace{1mm}}c@{\hspace{1mm}}c}
   &
   \includegraphics[height=2cm]{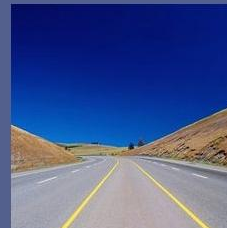}&
   \includegraphics[height=2cm]{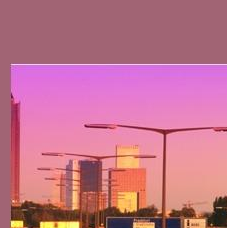}  &      
   \includegraphics[height=2cm]{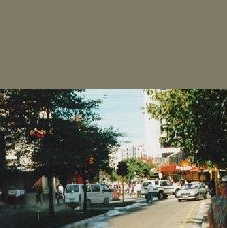}&
   \includegraphics[height=2cm]{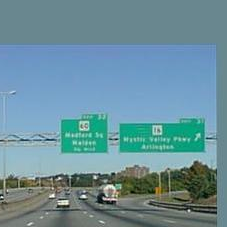}
\end{tabular}
   \caption{Four image patches of ``sky'' from the images sharing the same scene name ``highway'' in the SIFTFlow dataset. They have distinct apperances because of their different semantic layouts.}\label{fig:sky}
\end{figure}

Using scene names as semantic context has several limitations. Not all the datasets provide meaningful or accurate scene names. Even if each image has an accurate scene name, the number of images sharing the same scene name could be imbalanced, and thus some subclasses may  have too many or too few images. 
Some image patches still have large variance in appearance even though they have the same class labels and scene names, because scene names cannot accurately reflect the semantic layout of the surrounding region of a pixel. Fig. \ref{fig:sky} shows such examples, where there are four image patches of ``sky'' from images sharing the same scene name ``highway'' in the SIFTFlow dataset. However, because of their different semantic layouts, they have distinct appearances.
Here we introduce our second approach of using label map statistics to create label hierarchy.

\begin{figure}[t]
\centering
   \includegraphics[width=\linewidth]{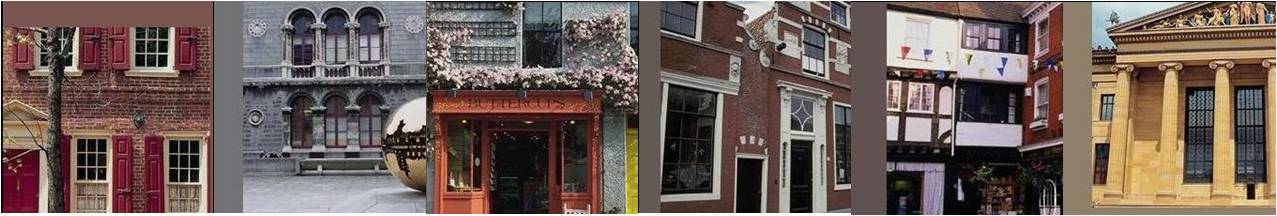}
   \includegraphics[width=\linewidth]{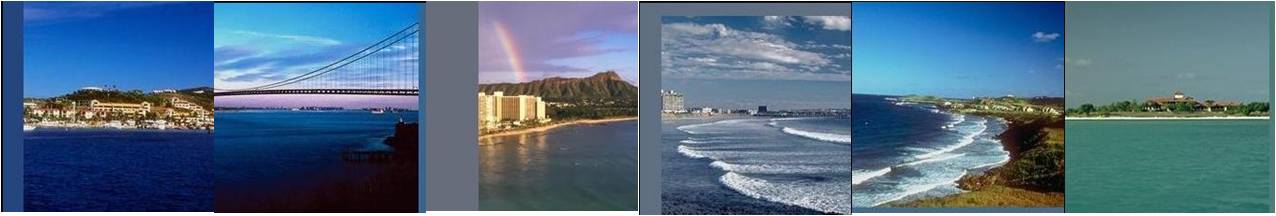}        
   \includegraphics[width=\linewidth]{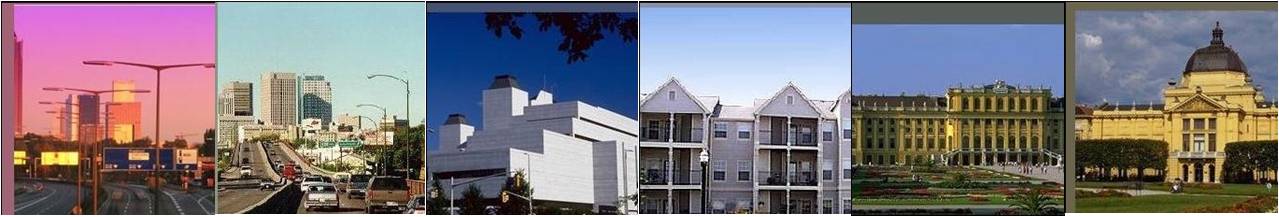}
   \includegraphics[width=\linewidth]{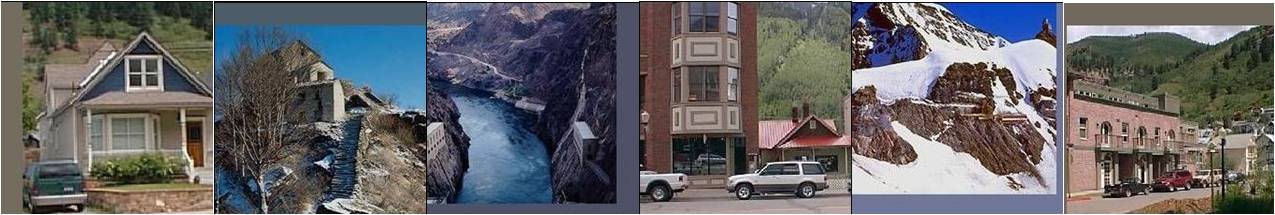}
   \caption{Example image patches of the ``building'' subclasses in the SIFTFlow dataset via label map clustering. (First row) parts of buildings. (Second row) far-away buildings at the horizon between sky and sea. (Third row) buildings under the sky. (Fourth row) buildings in the mountains. Uniformly-colored areas represent padded pixels.}\label{fig:building_subclass}
\end{figure} 

The scene labeling task provides rich label information. 
Given a training image patch, in the conventional pipeline, it is assumed to be classified to the class of its center pixel.
But the semantic context from the training label map is ignored. We propose to cluster the label maps of image patches to create subclasses for each original class. In this way, a more balanced and meaningful subclass dataset could be created (the bottom row in Fig. \ref{fig:label_frequency}). 
 
For each patch we compute the histogram $h_i$ of labels in the region of size $R\times R$ around the centered pixel from the label map. $h_i$ is normalized as $\tilde{h}_i = h_i/\Vert h_i\Vert_2$ to compensate for the cases when the patch is beyond the image boundary. $\tilde{h}_i$ is of dimension $L$, i.e., it has $L$ bins. For all the samples of class $j$,  k-means is applied to normalized histograms to obtain $K_j$ cluster centers.  Let $c_{j,1},..c_{j,K_j}$ be the cluster centers of class $j$, and can be obtained by 
\begin{equation}
\arg \min_S \sum_{k=1}^{K_j} \sum_{h\in S_{j,k}} \| h - c_{j,k} \|^2,
\end{equation}
where $S_{j,k}$ stores the set of samples that are assigned to the nearest cluster center $c_{j,k}$.
Each cluster center is denoted as a new subclass, and each sample $\tilde{h}_i$ is assigned with a new subclass label. The number of clusters $K_j$ for each class is chosen by taking the following factors into account. It should be positively related to the number of training samples of classes $j$, and should not be too large in case there are too few training samples in the new subclasses. 
Again we first partition the original classes into common classes and rare classes, in the way we defined in Section \ref{subsubscn:name_based_cluster}. The rare classes still inherit their labels from the original labels and we only find clusters for each common class. 
The algorithm for finding the optimal $K_j$ for each common class is summarized in Algorithm \ref{algorithm_1}, 
which tests different $K_j$ values 1) to avoid those values that the algorithm fails to converge in 100 iterations and 2) to create more balanced subclasses. 
We also verified the sensitivity of $K_j$, and observed that the final accuracy is insensitive to the parameters $K_j$ (Section \ref{subsec:siftflow_results}). 

\begin{algorithm}
\caption{Computing the number of clusters $K_j$ for common class $j$}
\label{algorithm_1}
\KwIn{All the samples $\tilde{h}_i$ of common class $j$\; The number of samples in the largest rare class $n^*$\;}
\textbf{Initialization:} $K_j$ = 2\\
\For{$i$ = 2 to 15}{
1) Set the cluster number to be $i$, and run the k-means algorithm for 100 iterations\\
2) \If{the k-means algorithm in step 1)                                              does not converge}{Skip step 3)}
3) \eIf {all subclasses have more than $n^*$ samples}{$K_j=i$}{Terminate iterations} 
return $K_j$}
\end{algorithm} 

Fig. \ref{fig:building_subclass} shows some training samples of different subclasses of the original class  ``building''. The original class ``building'' has large intra-class variation because of the differences in scale, position, shape, and view point. 
By clustering their label maps, we group building patches of the same ``type'' in a more meaningful way. The images in the first row are taken from a close view point, and thus the buildings occupy most of the image patches. In the second row, we can see the horizons splitting the samples into two halves: ``sky'' and ``sea'', while the buildings look tiny because they are far away. The third row shows different buildings under the sky, where ``sky'' and ``building'' occupy about half of the image patch, respectively. The last row shows buildings in mountains.

The surrounding region used to compute the histogram is called ROI. 
By setting different sizes of ROI, the histogram of a label patch can act either as a local contextual descriptor or a global contextual descriptor. When ROI is within the patch, the local context in the center part of the patch is used for clustering. On the other extreme, we could also compute the histogram of the whole label map as the feature for clustering ($R=\infty$). Then the histograms describe the global context. Training samples from the images of similar scenes would be grouped together, disregarding where they are located in the images.  

\subsection{Learning deep feature representations with label hierarchy}
Since the optimization of the convolutional neural network object function is highly non-convex, a good initial point is crucial for recovering a good local minimum with back propagation. We finetune our CNN models from a starting point pre-trained on ImageNet dataset similar to \cite{farabet2013learning}. 
Two different methods are investigated to utilize the proposed label hierarchy for fine-tuning CNN and learn discriminative deep feature representations. 

\subsubsection{Sequential subclass-class fine-tuning}\label{subscn:from_hard_to_easy}
\begin{figure}[t]
	\centering
	\begin{tabular}{c@{\hspace{-1.5mm}}c}
		&\includegraphics[scale=0.43]{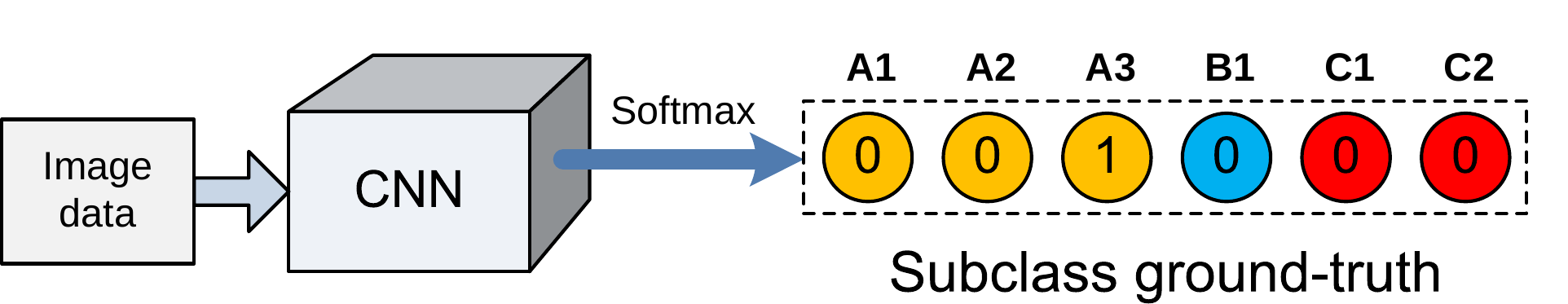}\\
		&\includegraphics[scale=0.43]{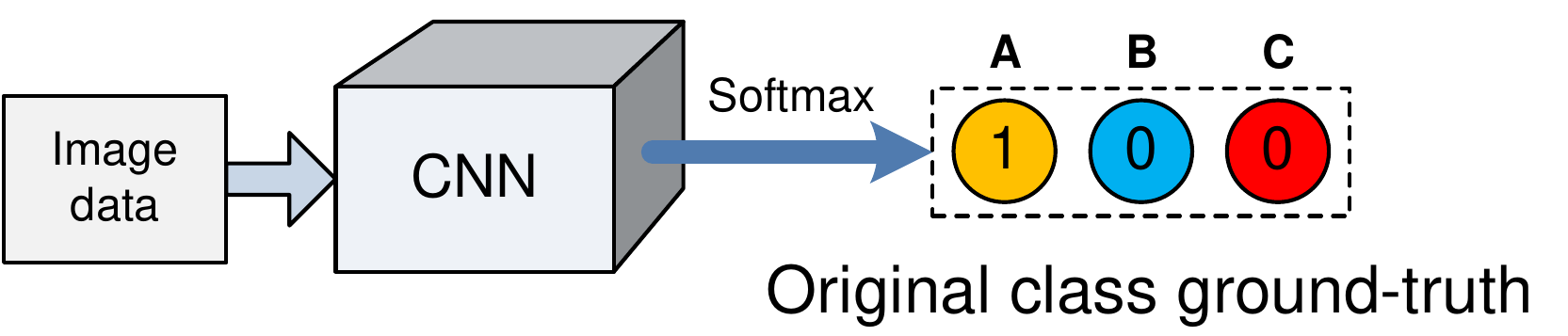}
	\end{tabular}
   \caption{Illustration of the sequanatial subclass-class fine-tuning strategy described in Section \ref{subscn:from_hard_to_easy}. (Top) The CNN is first fine-tuned with the subclass labels in the 1) and 2) steps. (Bottom) the CNN is then fine-tuned with the original class labels in the 3) and 4) steps.}\label{fig:sequential}
\end{figure}

With the original labels and their fine-grained subclass labels obtained by either scene names or clustering label maps, we adapt the deep CNN models to the scene labeling task in four steps. 1) The model is fine-tuned on the fine-grained subclass labels obtained by one of the above mentioned approaches. Only the last fully connected layer is randomly initialized and all the lower layers are inherited from the ImageNet classification model. We fix the lower layers and only update the parameters in the last fully connected layer. This is because the lower layers can be viewed as generic image  descriptors and can be applied to scene labeling datasets. We directly use these features to train the last fully connected layer, which can be regarded as a linear classifier.  2) After our model in the first step  converges, we further fine-tune it by updating the parameters of the whole network.  3) We switch back to the original task by replacing the last fully connected layer with a randomly initialized new one to classify the original classes. In this step, we fix the lower layers again and only update the new  fully connected layer. 4) All the layers are optimized on the original task jointly. The sequential fine-tuning strategy is illustrated in Fig. \ref{fig:sequential}.

We qualitatively investigated the effects of each of the steps. In practice, we observe that the third step is not indispensable. Only marginal performance drop is observed if it is skipped. However, if we skip the first step, the final performance would drop for about $0.5\%$ on the SIFTFlow dataset. This may be explained by the fact that the data distribution between the ImageNet and the scene labeling task is quite different. Skipping the first step is similar to setting a bad initialization point for the CNN model. Directly updating all the parameters at once may make the CNN easily converge to a bad local minimum. Fixing the lower layers provides a mild way for the optimization so that it is not too much influenced by the randomly initialized last fully connected layer. In comparison, transferring from the new subclasses to the original classes in the third step is much easier because both tasks are highly related and is based on similar data.

\subsubsection{Hierarchical multi-label fine-tuning}\label{subscn:hierarchical_multi_label}
\begin{figure}[t]
\centering
	\includegraphics[width=\linewidth]{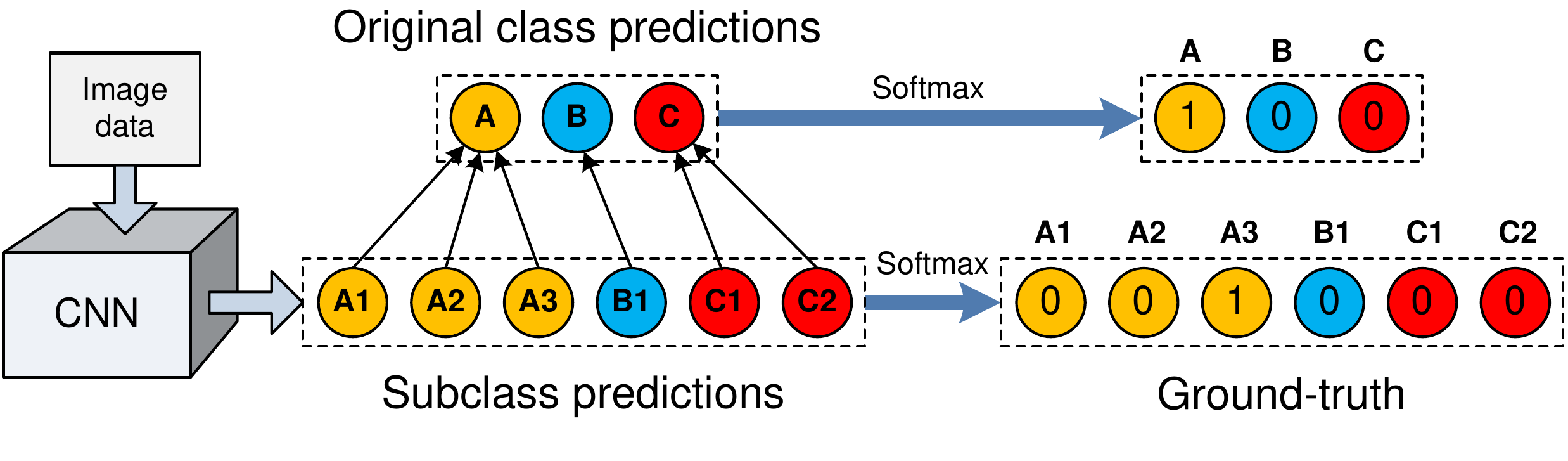}	
   \caption{Illustration of the hierarchical multi-label fine-tuning strategy described in Section \ref{subscn:hierarchical_multi_label}. A fully connected layer is introduced, which sums the predictions of subclasses as the predictions of the original classes. Training losses are back-propagated through both output layers to the CNN.}\label{fig:hierarchicla_model}
\end{figure}

In Section \ref{subscn:from_hard_to_easy}, we treat fine-tuning on the two label sets as two highly related problems, but CNN is optimized on the two tasks in a sequential manner. This would ignore the hierarchical relations between the original classes and their fine-grained subclasses. To make use of this relationship, we propose to simultaneously optimize the deep model for both tasks at different layers.  This is achieved by adding a new fully connected layer on top of the pre-trained CNN model. For a given training sample, our model first predicts the probability distribution on the subclasses and feeds the scores to the newly added fully connected layer to compute the probability of the original classes. In this layer, we explicitly set the scores of the original classes to be the sum of those of their corresponding subclasses. This is achieved by setting the weight matrix $W$ of size $L\times n$ to be a sparse matrix, where  $L$  and $n$ are the numbers of original classes and  new subclasses, respectively. The $j$th row of $W$ corresponds to the original class $j$, where only the entries corresponding to the subclasses of $j$ are 1, while other entries are 0. The errors of predicting the original labels can be back-propagated to lower layers together with the errors of predicting the new subclasses. In Fig. \ref{fig:hierarchicla_model}, the ground truth of the training sample is subclass $A_3$, which is a child of class $A$. We compute the scores of the subclasses at the second from the top layer and convert them to probabilities of the subclasses through the softmax operation. Meanwhile, the scores of subclasses that have the same parent are summed up to be the score of their parent class, which is used for calculating the error on the original label. Thus, the two tasks are optimized jointly while their hierarchical relations are utilized in the optimization.

The fine-tuning contains two steps. In the first step, we fix $W$, so that the score of the original class is forced to be the sum of those of its children subclasses. After training CNN in the first step converges, we further fine-tune our model by updating the parameters in $W$ with all other layers. In this way, we start from a good initialization point and learn the true relationship between the scores of the original classes and the subclasses.  The loss function for the hierarchical multi-label loss can be written as 
\begin{equation}
Loss(p_i, \hat{l}_i,\tilde{l}_i;\theta) = 
\dfrac{1}{m} \left\{ \sum_{i=1}^m L(p_i,\hat{l}_i) + \alpha L(q_i ,\tilde{l}_i) \right\}+ \dfrac{\beta}{2}  \| \theta \|_2^2,\label{eqn:Loss}
\end{equation}
where $m$ is the batch size, $p_i \in R^{n\times 1}$  is the scores of the subclasses of the $i$th patch, $\hat{l}_i$ and $\tilde{l}_i$ are the subclass label and the original label of the $i$th patch, respectively,  $\alpha$ is the weight between the losses of the new subclass and the original class. $\beta \| \theta \|_2^2$ is the weight decay term to punish large weights,  $q_i$ is scores of the original classes obtained by the \textit{softmax} function
\begin{equation}
q_{i,j} =  \dfrac{e^{W_jp_i}}{\sum_{k} e^{W_kp_i} },
\end{equation} 
$L$  is the loss function
\begin{equation}
L(p,l) = -\sum_{j=1}^L l_j \log q_{j},
\end{equation}
where the label $l_j=1$, if the patch is labeled $j$, and $0$ otherwise.
All the filters are updated by the gradients from the two loss layers, except the weights $W$ in the last inner-product layer, or the $1\times 1$ convolutional layer in the case of FCN.

\section{Experiments}

\subsection{Experimental setup}
\paragraph{Evaluations on different models}
Our proposed methods on learning better deep feature representations for scene labeling are not limited to specific CNN architectures. We tested their effectiveness on three popular CNN models, the Clarifai \citep{zeiler2014visualizing}, the OverFeat \citep{sermanet2013overfeat} and the FCN \citep{long2014fcn} models, which are all pre-trained on the ImageNet dataset. The former two models are patch-based while FCN takes the whole image as input. The three CNN models without using the label hierarchy are utilized as the baselines. 
During training, we adopted the mini-batch Stochastic Gradient Descent (SGD) to optimize the CNN models and the mini-batch sizes are 64, 64 and 10 for the three models, respectively. The learning rate is initialized as 0.001 and decreased by a factor of 10 with a stepsize of 20000. During testing, the efficient pixelwise forward-propagation algorithm \citep{li2014highly} was adopted for the Clarifai and the OverFeat models. For each $256\times256$ image, it takes $6s$ and $9s$ respectively on a modern GPU to generate the final label map, and $3s$ for FCN. 
 
\paragraph{Evaluation metrics}
The evaluation metrics are per-pixel and per-class accuracies. Let $c_i$ be the number of pixels correctly labeled as class $i$, $t_i$ denote the total number of pixels in total, and $L$ be the number of classes. We compute 
\begin{itemize}
\item per-pixel accuracy: 
$\displaystyle \sum_{i=1}^L c_i / \sum_i^L t_i$,
\item per-class accuracy: 
$\displaystyle (1/L) \sum_{i=1}^L c_i/t_i$
\end{itemize}
to evaluate the compared algorithms.

\paragraph{Hyper parameters}
In all experiments, we fixed the weights $\alpha=1$ and $\beta = 0.00025$ in Equation (\ref{eqn:Loss}). We explore the influence of the rare class ratio $\rho$ in Section \ref{subsec:siftflow_results} and the region size $R$ in Sections \ref{subsec:siftflow_results} and \ref{subsec:stanford_results}. In other experiments we fixed $\rho=93\%$ and $R=129$.

\paragraph{Data preparation}
We evaluated the compared methods on four scene labeling datasets, the SIFTFlow \citep{liu2008sift},  Stanford background \citep{gould2009decomposing},  Barcelona \citep{tighe2010superparsing} and  LM+Sun \citep{tighe2010superparsing_jnl} datasets. 
For each dataset, we augmented the training set by randomly scaling, rotating and horizontally flipping each training image for $5$ times.  The scaling factors and the rotation angles were randomly chosen in the ranges of $[0.9, 1.1]$ and $[-8^{\circ}, 8^{\circ}]$, and the random flipping probability was $50\%$. For the Clarifai and OverFeat models each augmented training image was first divided into superpixels ($\sim300$ pixels per superpixel) using the VLFeat open source library \citep{vedaldi08vlfeat}, and $227\times227$ image patches centered at the superpixels were then cropped as training samples. To ensure that image patches cropped at the boundaries of the training image are of the same size, the mean value of each image was padded outside the image boundaries.

\subsection{Results on the SIFTFlow dataset}
\label{subsec:siftflow_results}
The SIFTFLow dataset consists of 2488 training images and 200 test images. All the images are of size $256\times 256$ and contain 33 semantic labels. We tested the Clarifai, OverFeat and FCN baseline models on this dataset. 

As shown in Fig. \ref{fig:label_frequency}, we identify 11 out of the 33 classes as common classes with a rare class ratio $\rho=93\%$. For creating the label hierarchy based on scene names, since there are $8$ scene names in the training dataset, each common class is divided into $8$ subclasses, while each rare class is the subclass of itself. Thus we divided the original classes into $101$ subclasses. For creating the label hierarchy via label map clustering, we divide the original classes into $128$ subclasses in total. we created $3$ sets of subclasses by varying the $R$ value: $R=129, R=227$ and $R=\infty$, respectively. 
We compared our method with state-of-the-art methods, which includes both deep-learning-based \citep{farabet2013learning, pinheiro2013recurrent, long2014fcn, sharma2015deep, eigen2015predicting, shuai2015dag, caesar2016region, wang2016learnable}  and non-deep-learning-based methods \citep{tighe2010superparsing, liu2008sift, yang2014context}. The accuracies by different methods are recorded in Table \ref{tbl:siftflow_result}.


\begin{table}[t]
\centering
\begin{tabular}{|l|c|c|}
\hline
~~~~~~~~~~~~~~~~~~Method & Per-pixel & Per-class \\ \hline \hline
\cite{tighe2010superparsing}    & 0.769     & 0.294     \\
\cite{liu2008sift}         & 0.748     & n/a        \\
\cite{farabet2013learning}                      & 0.785     & 0.296     \\ 
\cite{pinheiro2013recurrent}          & 0.777     & 0.298     \\ 
\cite{sharma2015deep}         & 0.796     & 0.336     \\ 
\cite{yang2014context}                  & 0.798     & 0.487     \\
\cite{eigen2015predicting} & 0.868 & 0.464 \\
\cite{shuai2015dag} & 0.853 & 0.557 \\
\cite{caesar2016region} & 0.843 & \textbf{0.64} \\
\cite{wang2016learnable} &  \textbf{0.879} & 0.5 \\
\hline \hline
Clarifai baseline             & 0.832     & 0.426     \\
Clarifai+S+SN                 & 0.839     & 0.427     \\
Clarifai+S+PC ($R=129$)       & 0.842     & 0.438     \\ \hline \hline
OverFeat baseline             & 0.839     & 0.446     \\
OverFeat+S+SN                 & 0.85      & 0.447     \\
OverFeat+S+LC ($R=129$)    & 0.854     & 0.469     \\
OverFeat+S+LC ($R=227$)    & 0.854     & 0.459     \\
OverFeat+S+LC ($R=\infty$) & 0.856     & 0.467     \\
OverFeat+H+LC ($R=129$)  & 0.857     & 0.446     \\ \hline\hline
FCN \citep{long2014fcn}                  & 0.851     & 0.517 \\ 
FCN+S+LC ($R=129$)  & 0.875 & 0.522\\
FCN+H+LC ($R=129$)  & \textbf{0.878} & 0.52\\
\hline
\end{tabular}
\caption{Per-pixel and per-class accuracies on the SIFTFlow dataset by different methods. (The best accuracy by a single model and the best overall accuracy are marked in bold. S = Sequantial strategy, H = Hierarchical loss,  SN = Scene Name, LC = Label map Clustering).}\label{tbl:siftflow_result}
\end{table}

As shown by Table \ref{tbl:siftflow_result}, 
both ways of creating label hierarchies (Model+S+SN and Model+S+LC) lead to accuracy improvements over the baseline models. Here the single letters S and H refer to sequential and hierarchical fine-tuning strategies, and SN and LC refer to label hierarchies created by scene name and label map histogram clustering, respectively.  Note that the network trained by sequential strategy, i.e., Model+S+LC, has the same architecture as the baseline model. This demonstrates that, by decreasing intra-subclass variation, the proposed label hierarchies do help the CNN learn more discriminative deep feature representations without increasing the model complexity. Compared with creating label hierarchy based on scene names (Model+S+SN), creating label hierarchy via label map clustering (Model+S+LC) leads to better labeling accuracy since label map clustering is able to generate more meaningful subclasses as explained in Section \ref{subsubscn:label_map_based_cluster}. Comparing the results by the two training strategies that utilizes label hierarchy (Table \ref{tbl:siftflow_result}(a) and (d)), we observe that the hierarchical fine-tuning strategy leads to better labeling accuracy. Modeling the hierarchical relations between classes and their subclasses by network structure provides more useful supervision to learn effective deep feature representations. Such a strategy also reduces the training time since it requires only a two-step fine-tuning instead of the four-step fine-tuning in the sequential strategy. We also observe that the results are not sensitive to the $R$ values. When $R$ equals the patch size (i.e. $R=227$) or the whole image ($R=\infty$), the performance is stable.

Our FCN baseline model is from \cite{long2014fcn}, which is based on the VGG16 \cite{chatfieldreturn} net. Following \cite{long2014fcn}, we first train the coarse FCN-32s net and then use its parameters to initialize the FCN-16s net. The upsampling deconvolutional layers are initialized by bilinear interpolation but are allowed to be learned during training. By applying the sequential finetuning method, we successfully boosted the per-pixel accuracy from 0.851 to 0.875, which shows that our method can be utilized for different networks. Our result by FCN+S+LC ($R=129$) outperforms that by  \cite{eigen2015predicting}, which was obtained by adding more convolutional layers to the VGG network, while our method did not modify the network structure. 


We further study the effectiveness of other parameters of our method,  the rare classes ratio $\rho$ and  the number of clusters $K_j$ for each class $j$. In the previous experiment setting as in Table \ref{tbl:siftflow_result}, we fix $\rho=93\%$ and $\widehat{K}_j$ is obtained by Algorithm \ref{algorithm_1}. Note $\rho$  determines how many classes being split into subclasses. $K_j$ is the target number of clusters for the k-means algorithm for each common class $j$ for creating our second label hierarchy. In this experiment, we set $\rho=93\%$ and $\widehat{K}_j$ as the baseline, and see if the proposed method is stable when one of the parameters is changed. The results are shown in Table \ref{tbl:ablation}. We did not observe significant differences between these settings. So the conclusion can be drawn that the proposed method is not sensitive to these two parameters in our tested ranges. 
\begin{table}[]
\centering
\label{my-label}
\begin{tabular}{|l|l|l|}
\hline
                     & per-pixel & per-class \\ \hline
$\rho=93\%$, $K_j=\widehat{K}_j$     & 0.854     & 0.469     \\ \hline
$\rho=90\%$, $K_j=\widehat{K}_j$     & 0.853     & 0.454     \\ \hline
$\rho=95\%$, $K_j=\widehat{K}_j$     & 0.854     & 0.465     \\ \hline
$\rho=93\%$, $K_j=\widehat{K}_j\times 0.8$ & 0.853     & 0.461     \\ \hline
$\rho=93\%$, $K_j=\widehat{K}_j\times 1.5$ & 0.852     & 0.453     \\ \hline
\end{tabular}
\caption{Accuracies of OverFeat+S+LC ($R=129$) with different rare class  ratios $\rho$ and the numbers of subclasses $K_j$ on SIFTFlow dataset. $\rho = 90\%/93\%/95\%$ corresponds to $9/11/13$ common classes out of 33 classes in total. }\label{tbl:ablation}
\end{table} 

\begin{figure}[t]
\centering
   \includegraphics[width=\linewidth]{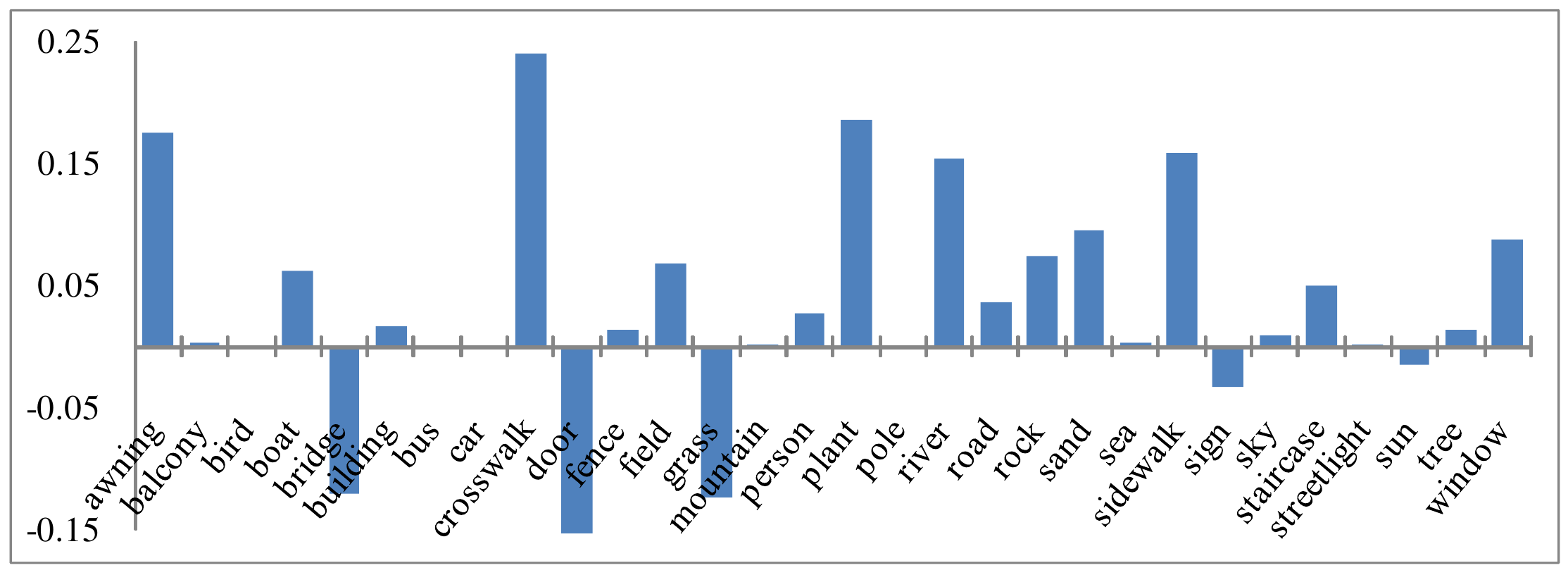}
   \caption{Per-class improvements of OverFeat+S+LC ($R$=227) over the OverFeat baseline model. A positive value means the class mean accuracy is improved.}\label{fig:per_class_accuracy}
\end{figure}

We qualitatively evaluate and analyse the accuracy improvements by our proposed method.
The per-class improvements by our proposed label hierarchy and sequential strategy, OverFeat+S+LC ($R=227$) in Table \ref{tbl:siftflow_result}(b), over the OverFeat baseline model is shown in Fig. \ref{fig:per_class_accuracy}. In Fig. \ref{fig:siftflow_many_result}, example labeling results by the two models are shown. Among the classes that are improved most, we can see that they are rare classes, i.e., classes that are not divided into subclasses, and are part of some other common classes. Their samples are more likely to be classified into the common classes. For example, ``awning'' is a part of ``building'', and ``crosswalk'' and ``sidewalk'' are accessories of ``roads''. By decreasing intra-subclass variation, the CNN can detect more meaningful visual patterns and learn better deep representations, and thus distinguish such rare classes more effectively.
Although most classes' accuracies are improved, three classes show notable performance drop, which are ``bridge'', ``door'' and ``grass''. This is because the baseline model has high precision but low recall on the three classes, i.e., the baseline model tends to classify pixels of many other classes into the three classes. The higher accuracies of the three classes were actually obtained by impairing other classes' accuracies. Take the bottom-right case in Fig. \ref{fig:siftflow_many_result} as an example, the baseline model prefers ``door'' to ``building'', where quite a few ``building'' pixels are wrongly classified. In our improved CNN model, the increasing accuracies of the other classes lead to lower precision but higher recall of the three classes.

\def\imagetop#1{\vtop{\null\hbox{#1}}}
\begin{figure*}[!ht]
	\centering
	\setlength{\tabcolsep}{0.3em}
	\begin{tabular}{c@{\hspace{-1mm}}c@{\hspace{1mm}}c@{\hspace{1mm}}c@{\hspace{1mm}}c@{\hspace{1mm}}c@{\hspace{1mm}}c@{\hspace{1mm}}c@{\hspace{1mm}}c@{\hspace{1mm}}c}
	& \imagetop{\includegraphics[width=2.32cm]{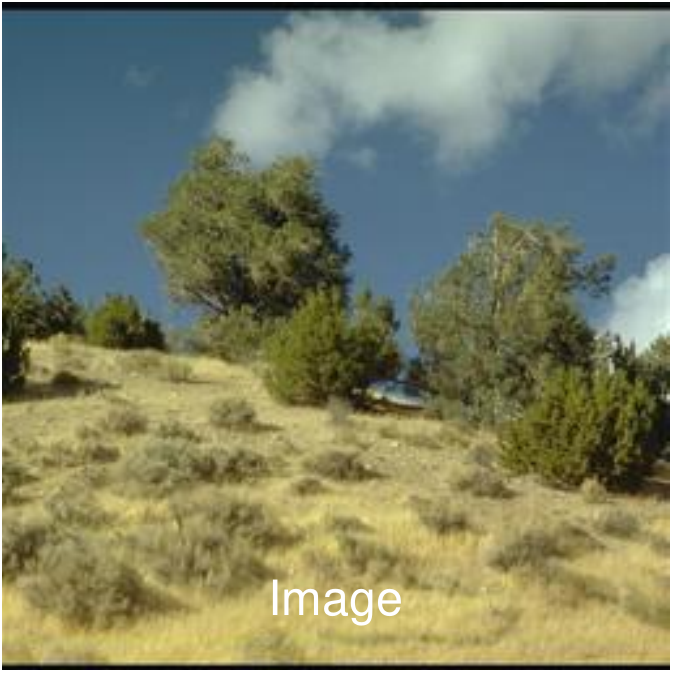}}
	& \imagetop{\includegraphics[width=2.32cm]{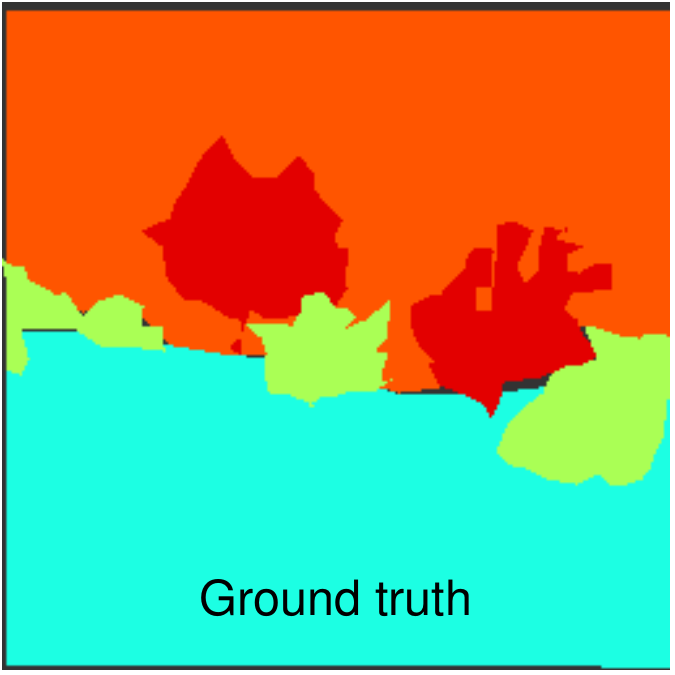}}
	& \imagetop{\includegraphics[width=1.3  cm]{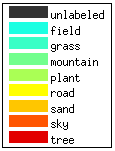}}
	& \imagetop{\includegraphics[width=2.32cm]{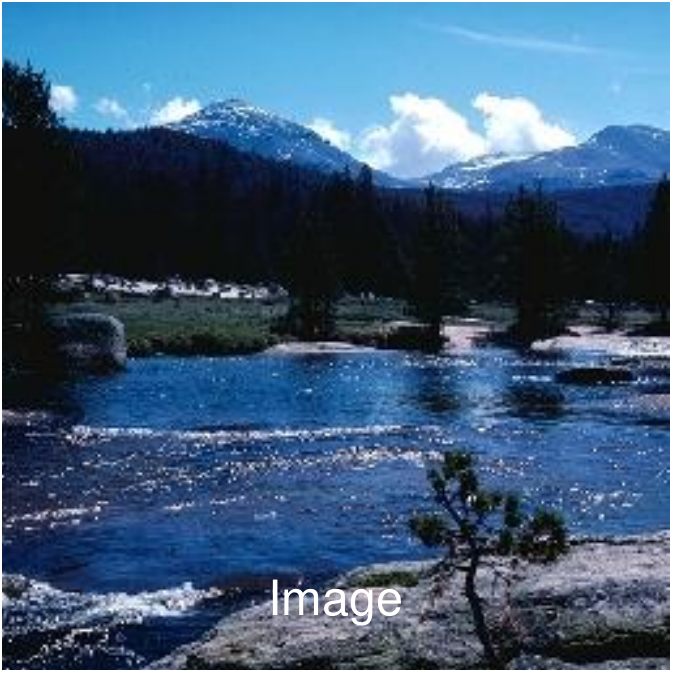}}
	& \imagetop{\includegraphics[width=2.32cm]{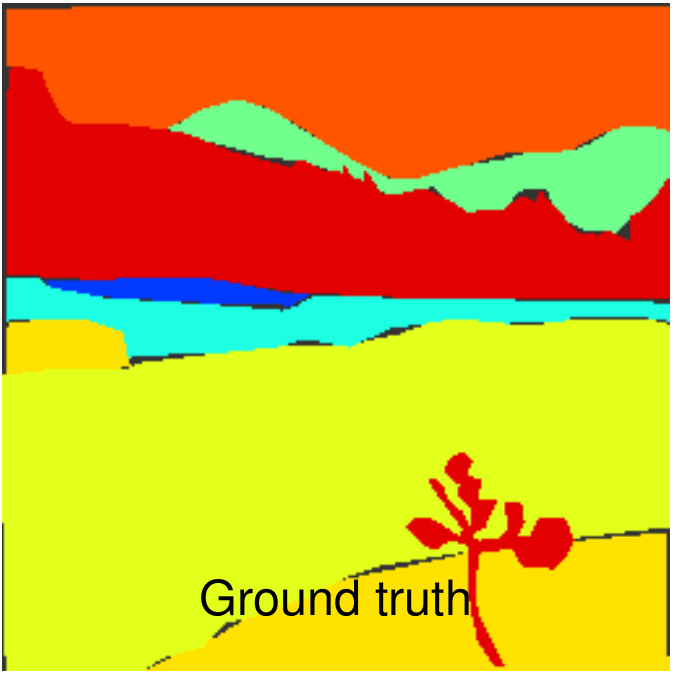}}
	& \imagetop{\includegraphics[width=1.3  cm]{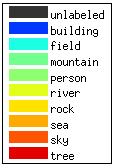}}
	& \imagetop{\includegraphics[width=2.32cm]{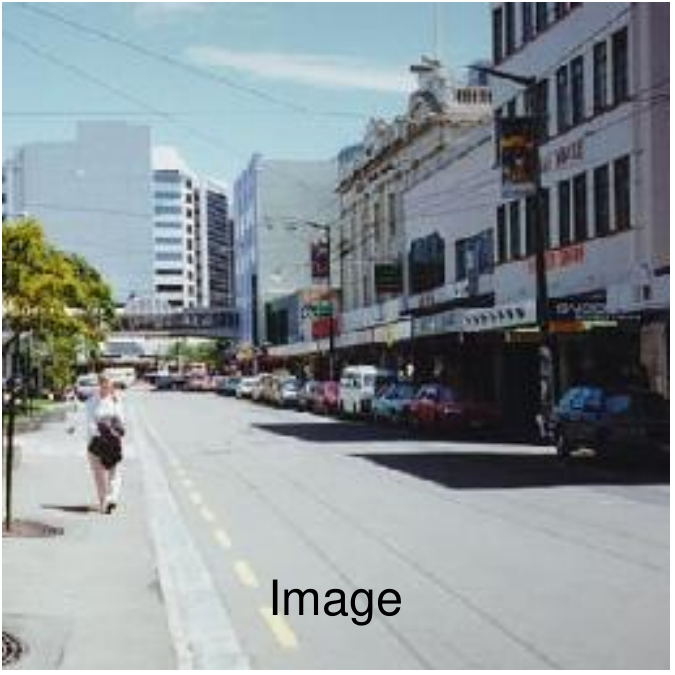}}
	& \imagetop{\includegraphics[width=2.32cm]{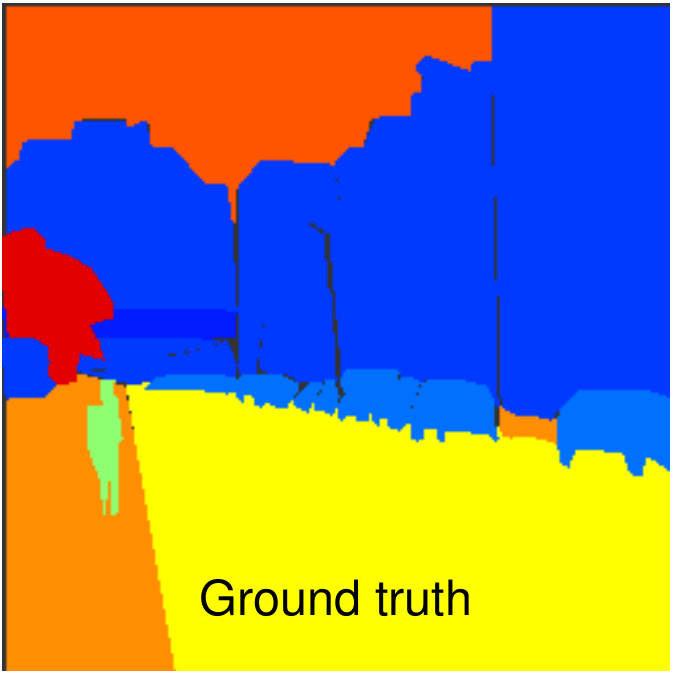}}
	& \imagetop{\includegraphics[width=1.3  cm]{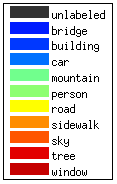}}\\[-5pt]
	& \imagetop{\includegraphics[width=2.32cm]{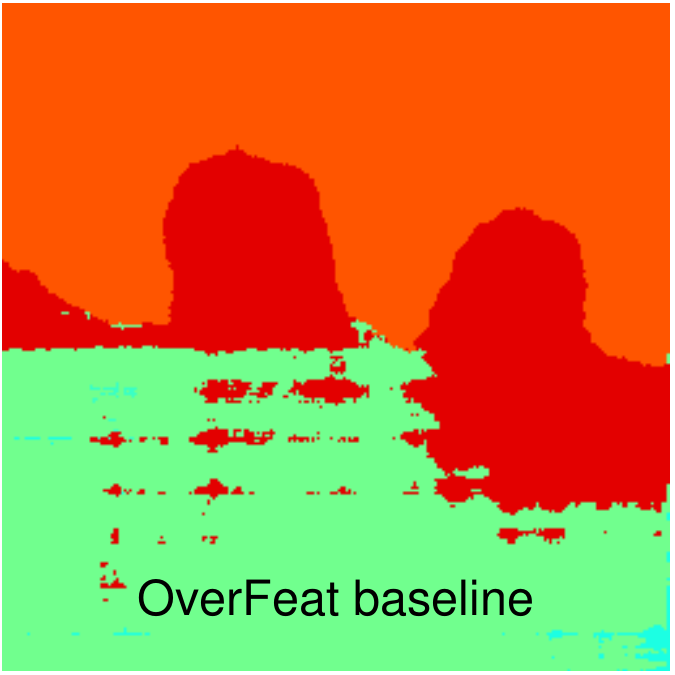}}
	& \imagetop{\includegraphics[width=2.32cm]{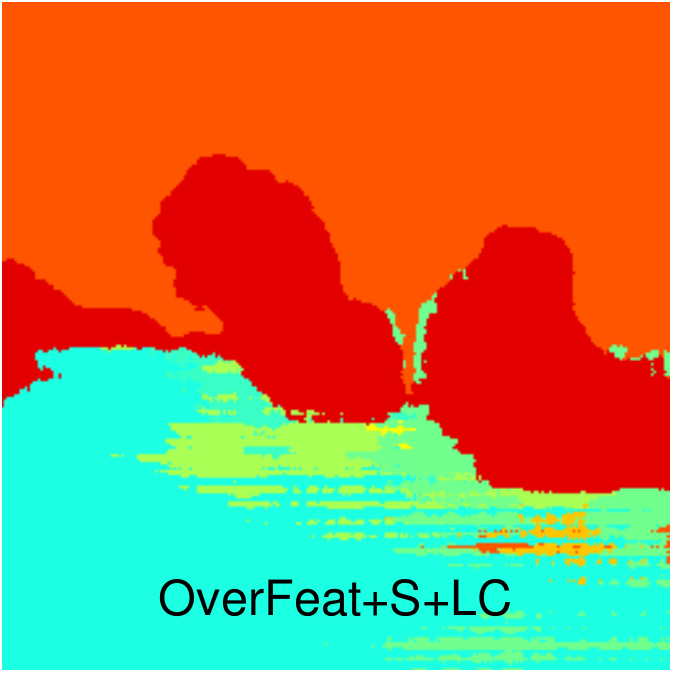}}
	&
	& \imagetop{\includegraphics[width=2.32cm]{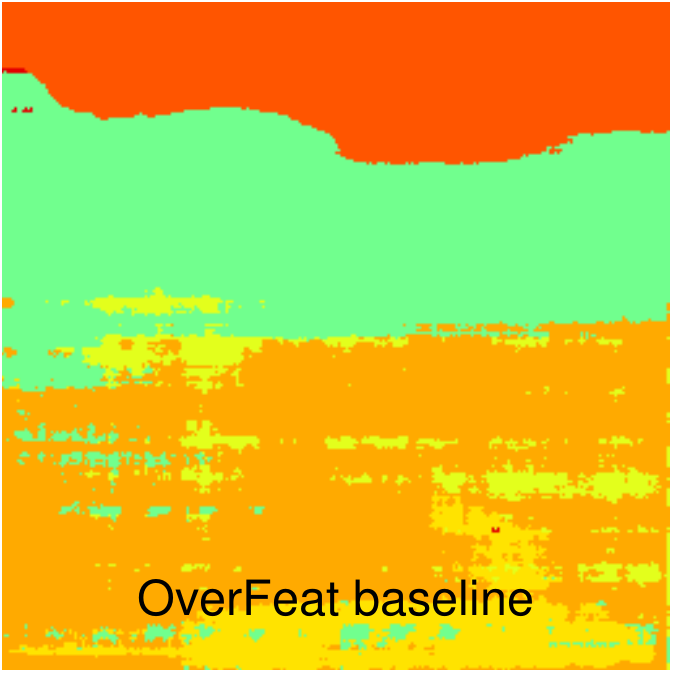}}
	& \imagetop{\includegraphics[width=2.32cm]{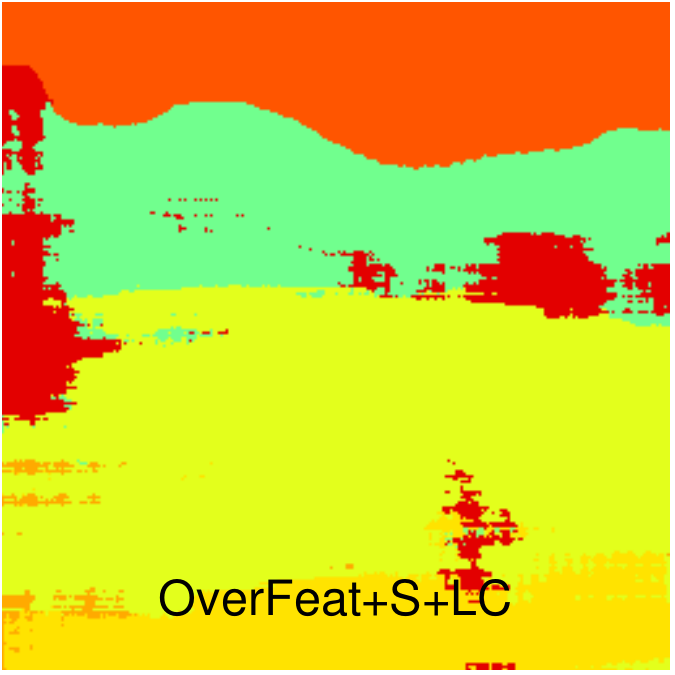}}
	&
	& \imagetop{\includegraphics[width=2.32cm]{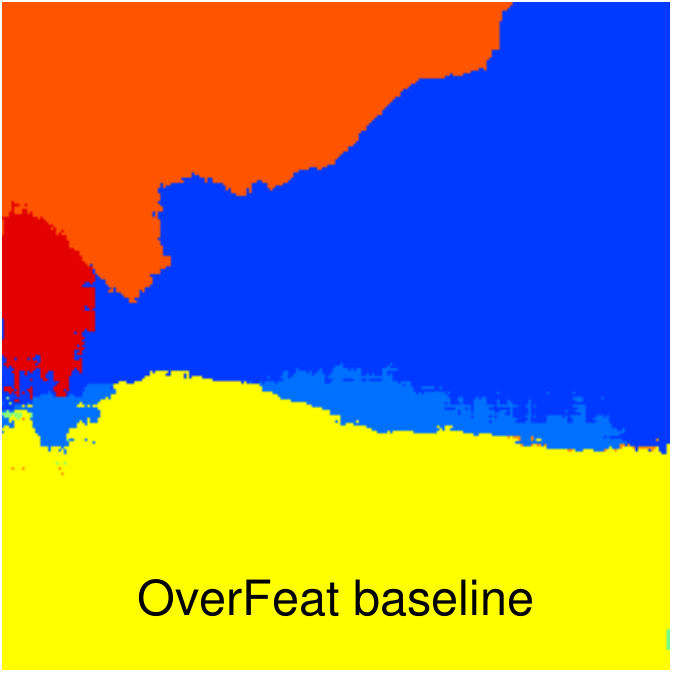}}
	& \imagetop{\includegraphics[width=2.32cm]{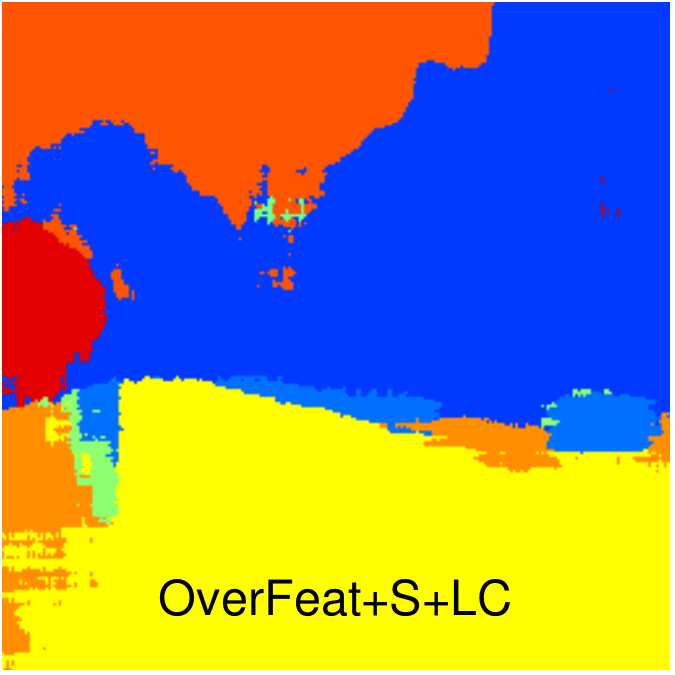}}\\
	& \imagetop{\includegraphics[width=2.32cm]{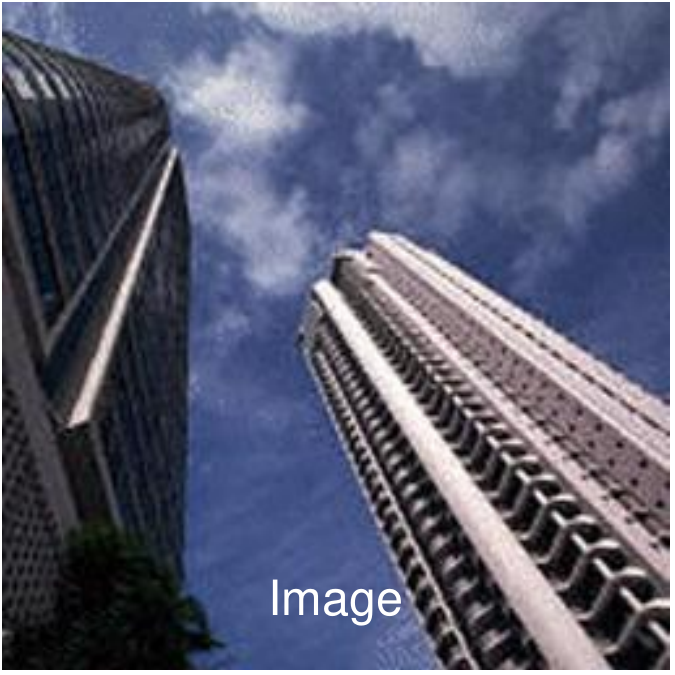}}
	& \imagetop{\includegraphics[width=2.32cm]{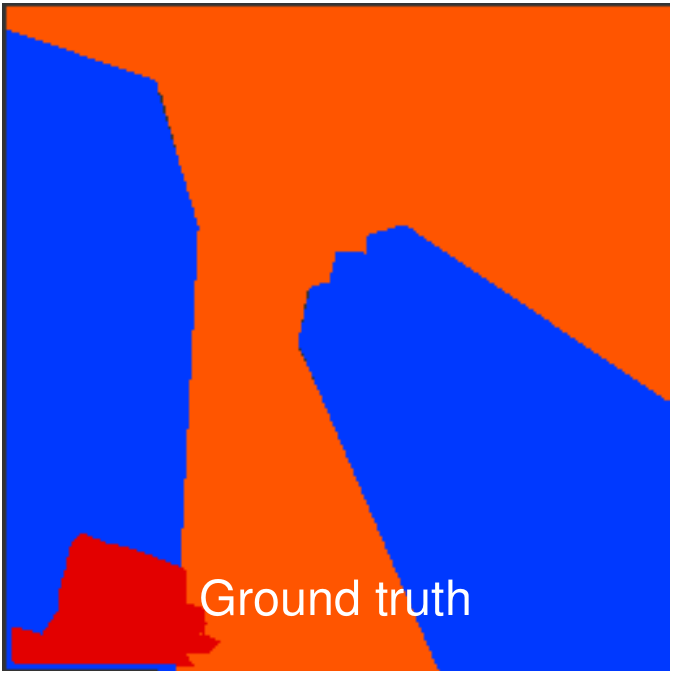}}
	& \imagetop{\includegraphics[width=1.3  cm]{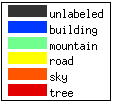}}
	& \imagetop{\includegraphics[width=2.32cm]{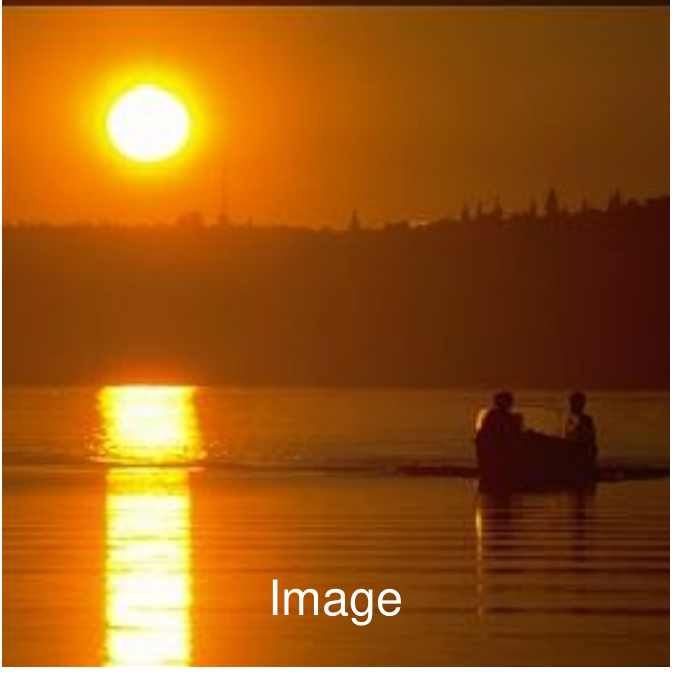}}
	& \imagetop{\includegraphics[width=2.32cm]{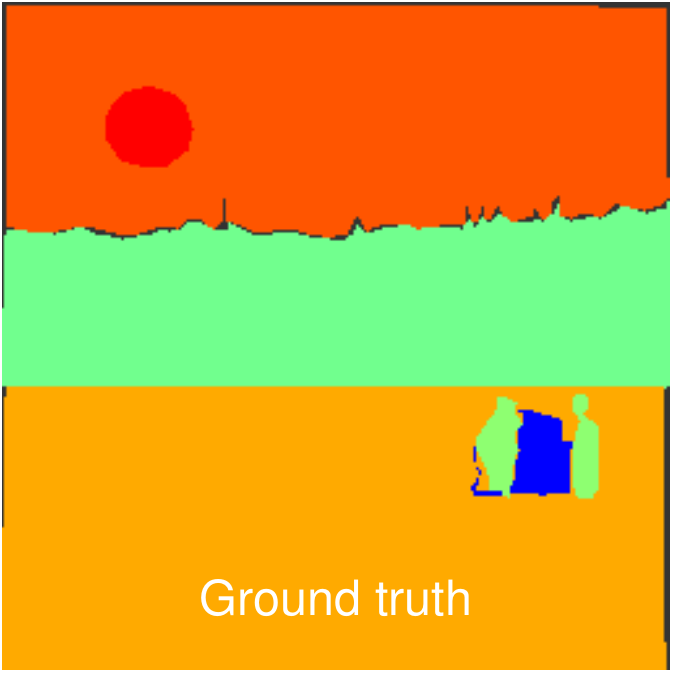}}
	& \imagetop{\includegraphics[width=1.3  cm]{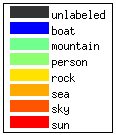}}
	& \imagetop{\includegraphics[width=2.32cm]{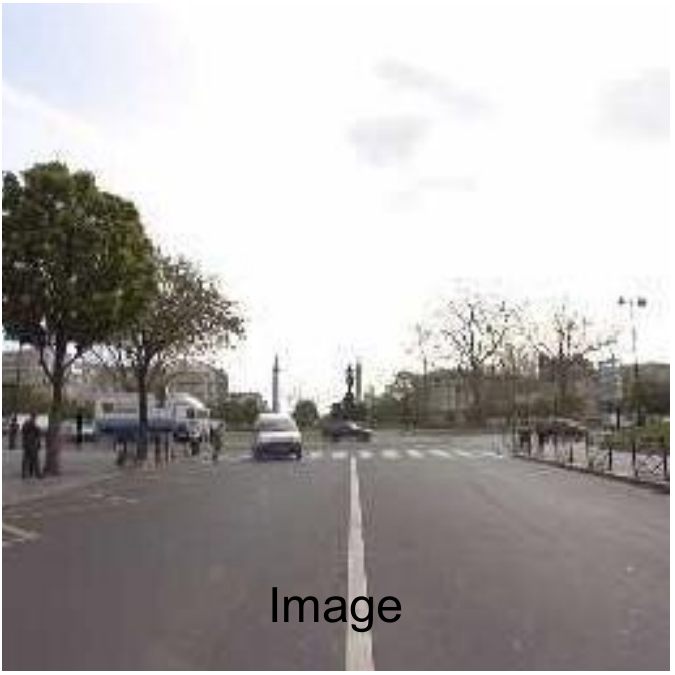}}
	& \imagetop{\includegraphics[width=2.32cm]{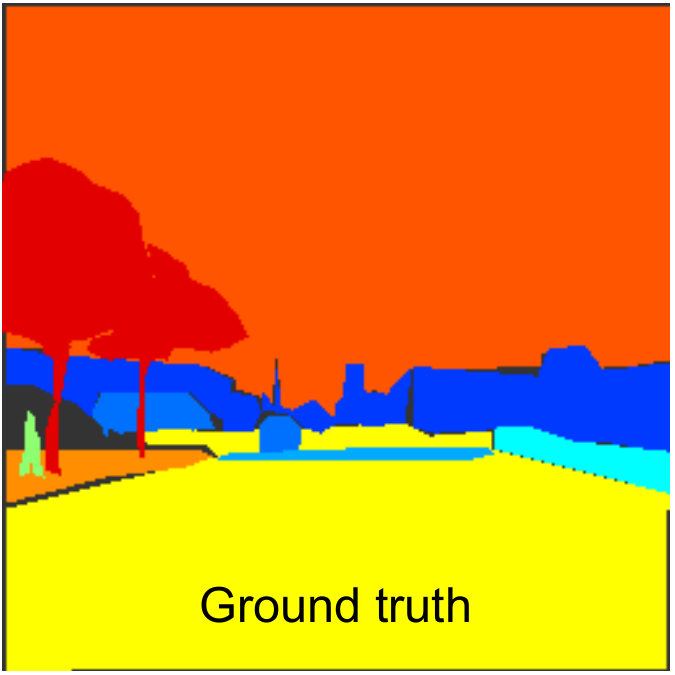}}
	& \imagetop{\includegraphics[width=1.3  cm]{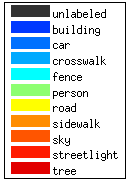}}\\[-5pt]
	& \imagetop{\includegraphics[width=2.32cm]{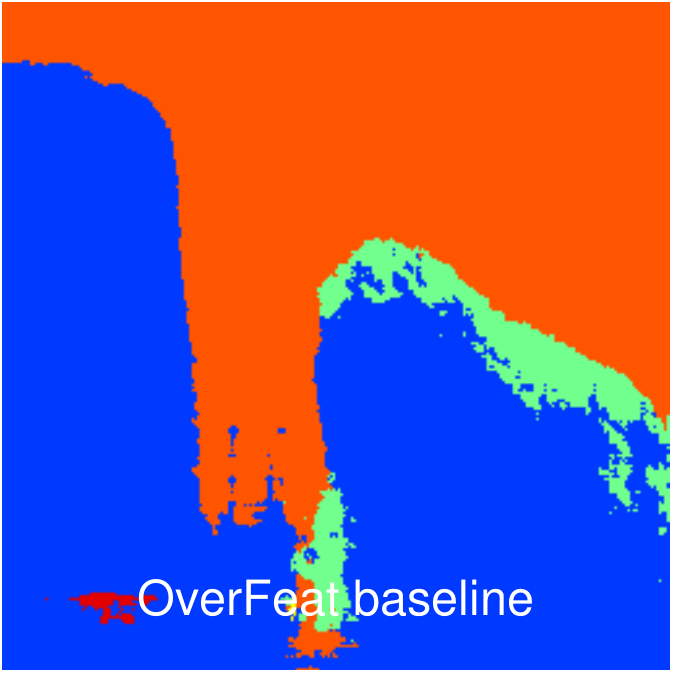}}
	& \imagetop{\includegraphics[width=2.32cm]{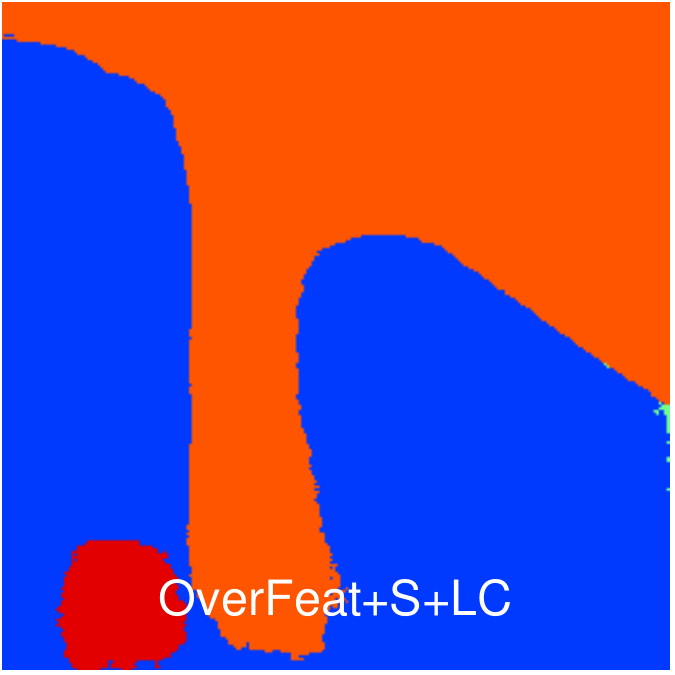}}
	&
	& \imagetop{\includegraphics[width=2.32cm]{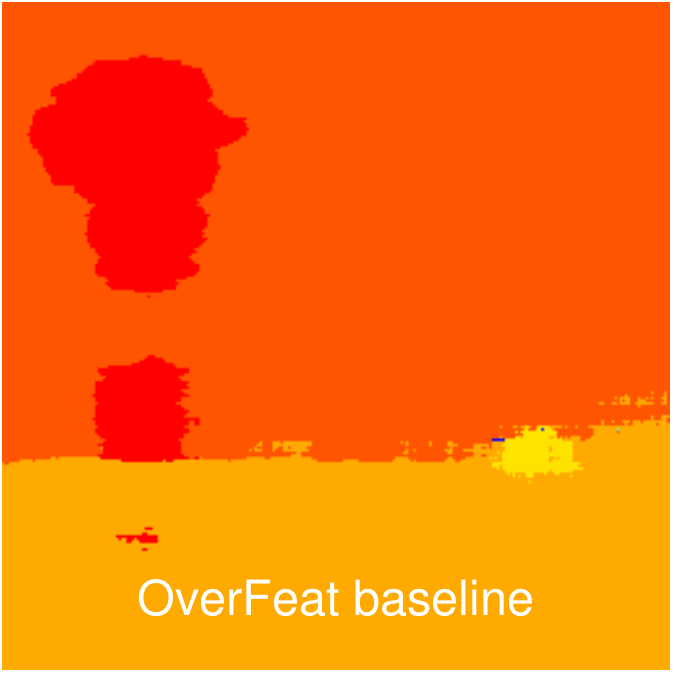}}
	& \imagetop{\includegraphics[width=2.32cm]{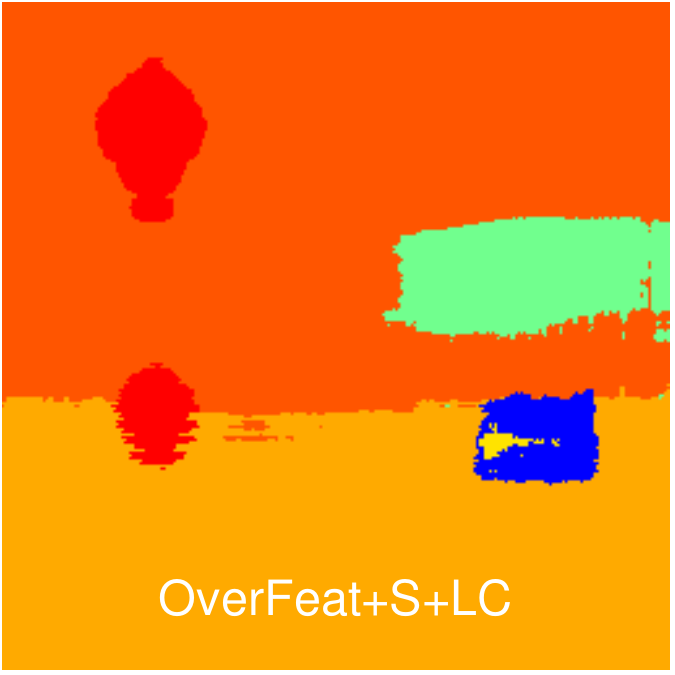}}
	&
	& \imagetop{\includegraphics[width=2.32cm]{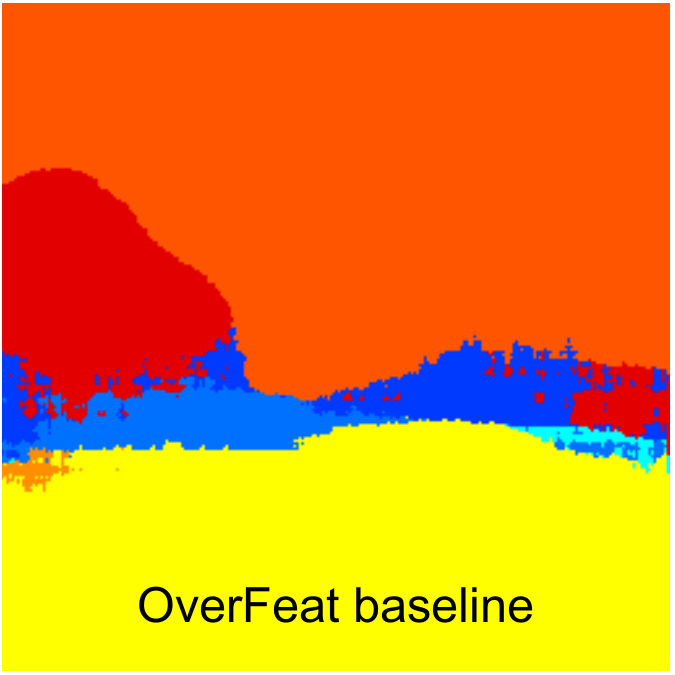}}
	& \imagetop{\includegraphics[width=2.32cm]{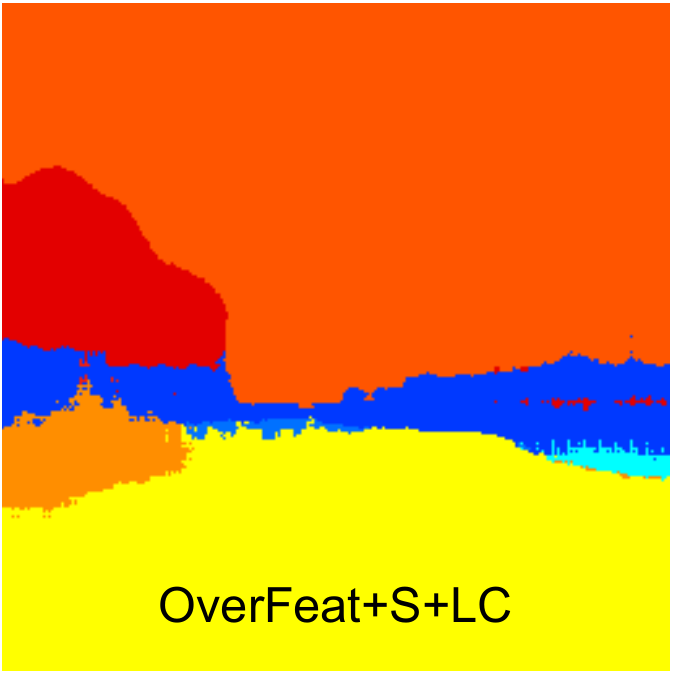}}\\
	
	& \imagetop{\includegraphics[width=2.32cm]{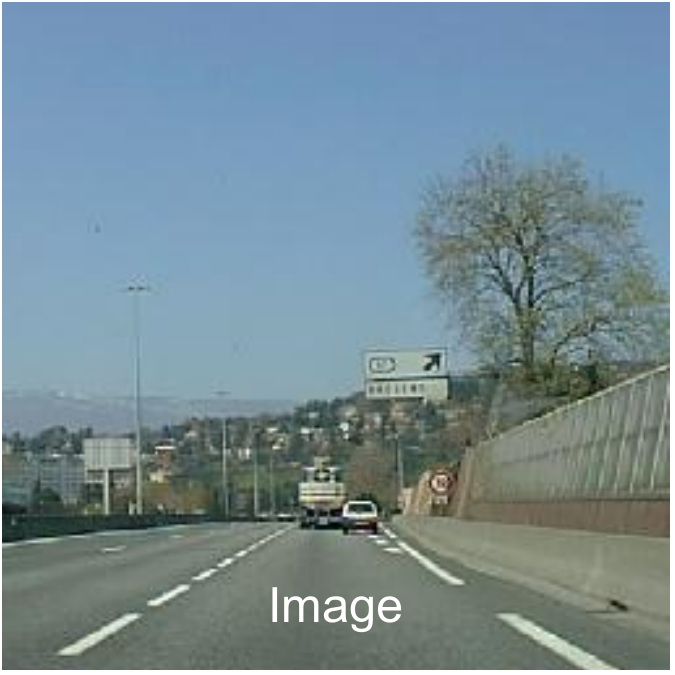}}
	& \imagetop{\includegraphics[width=2.32cm]{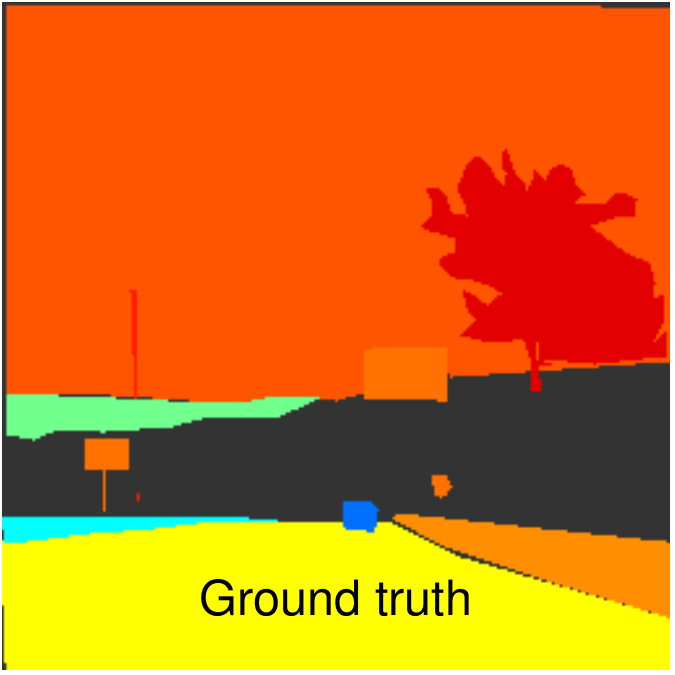}}
	& \imagetop{\includegraphics[width=1.3  cm]{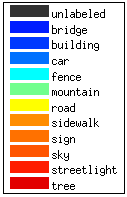}}
	& \imagetop{\includegraphics[width=2.32cm]{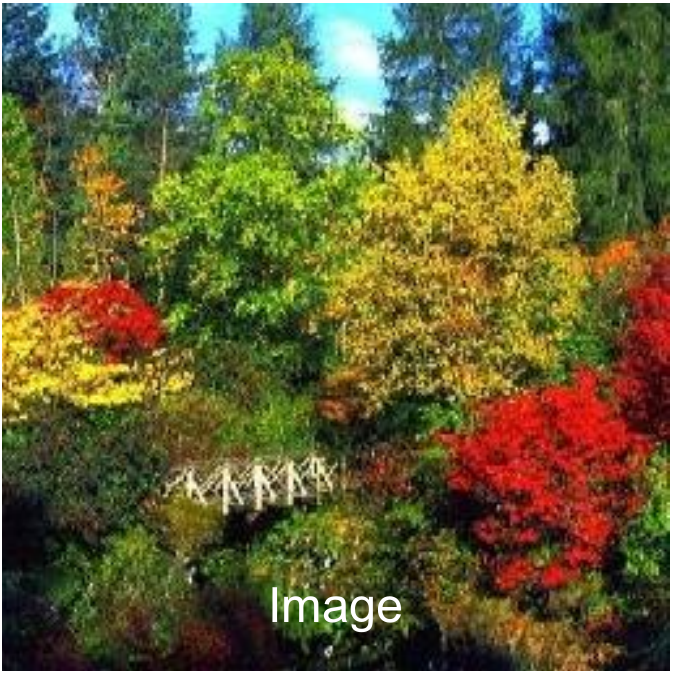}}
	& \imagetop{\includegraphics[width=2.32cm]{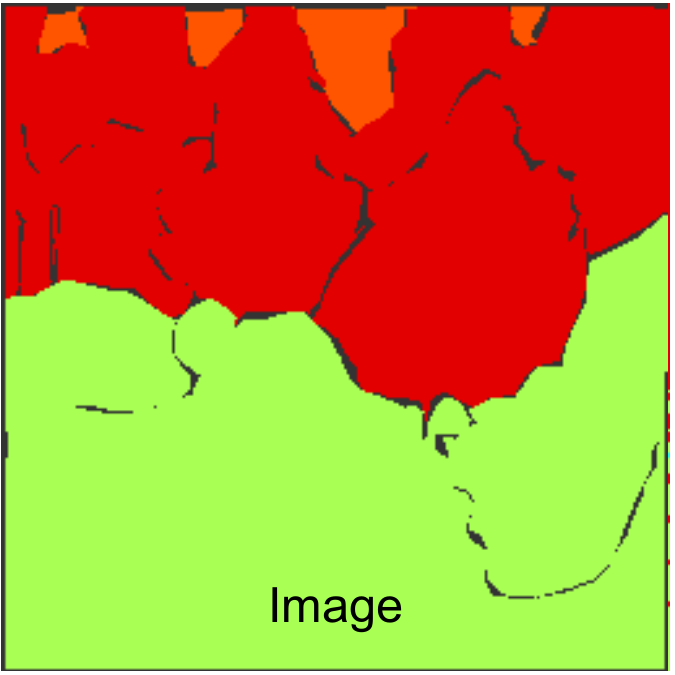}}
	& \imagetop{\includegraphics[width=1.3  cm]{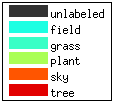}}
	& \imagetop{\includegraphics[width=2.32cm]{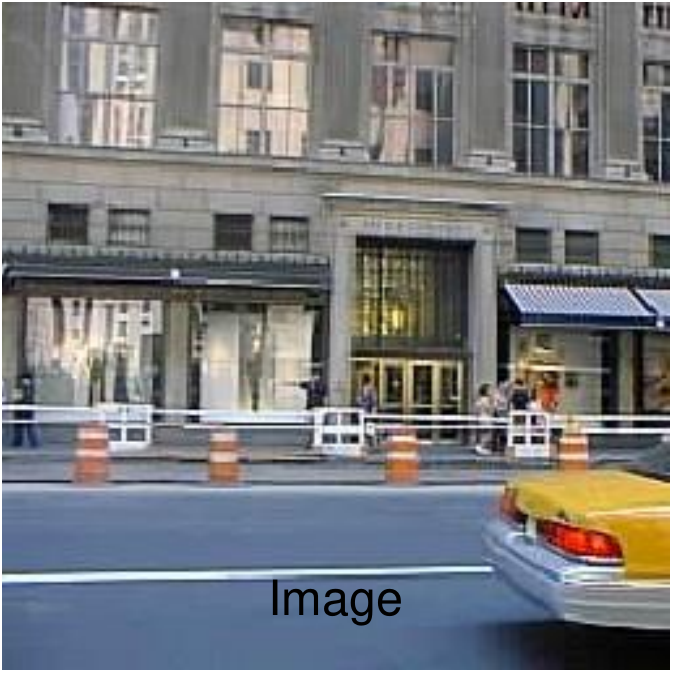}}
	& \imagetop{\includegraphics[width=2.32cm]{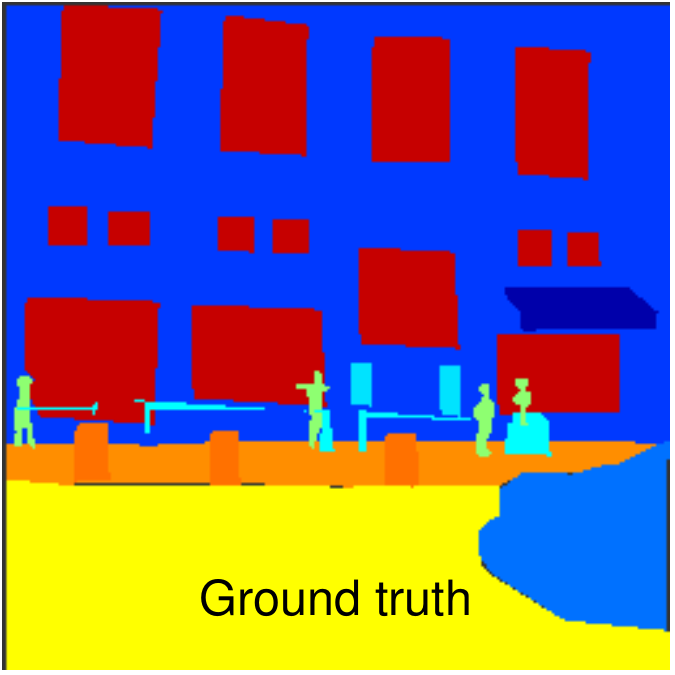}}
	& \imagetop{\includegraphics[width=1.3  cm]{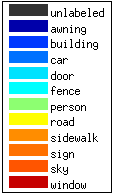}}\\[-8pt]
	& \imagetop{\includegraphics[width=2.32cm]{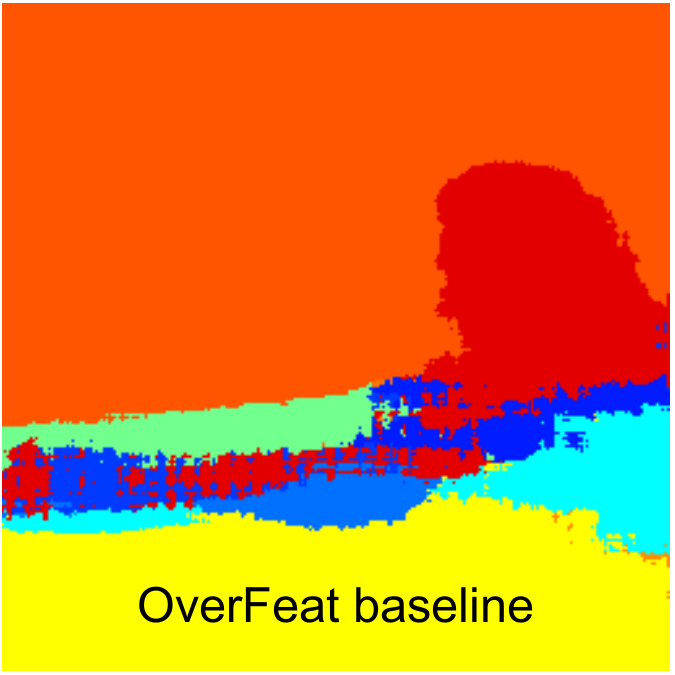}}
	& \imagetop{\includegraphics[width=2.32cm]{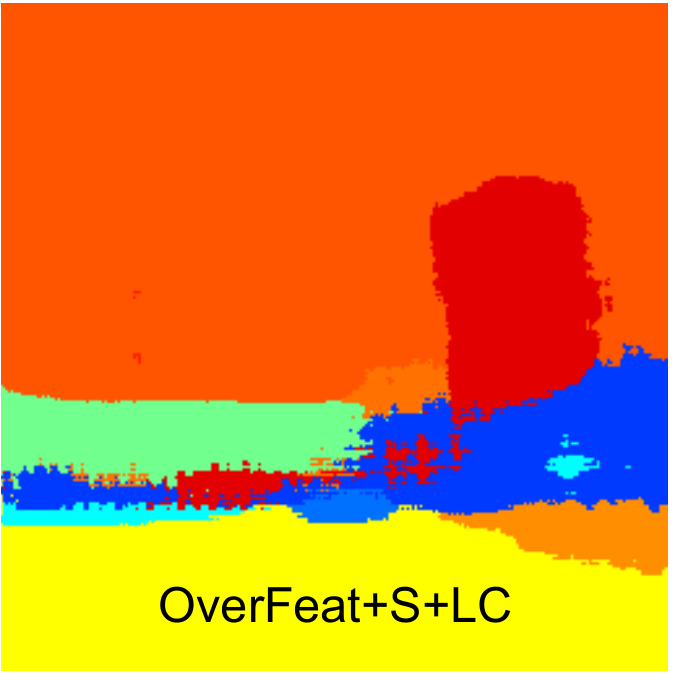}}
	&
	& \imagetop{\includegraphics[width=2.32cm]{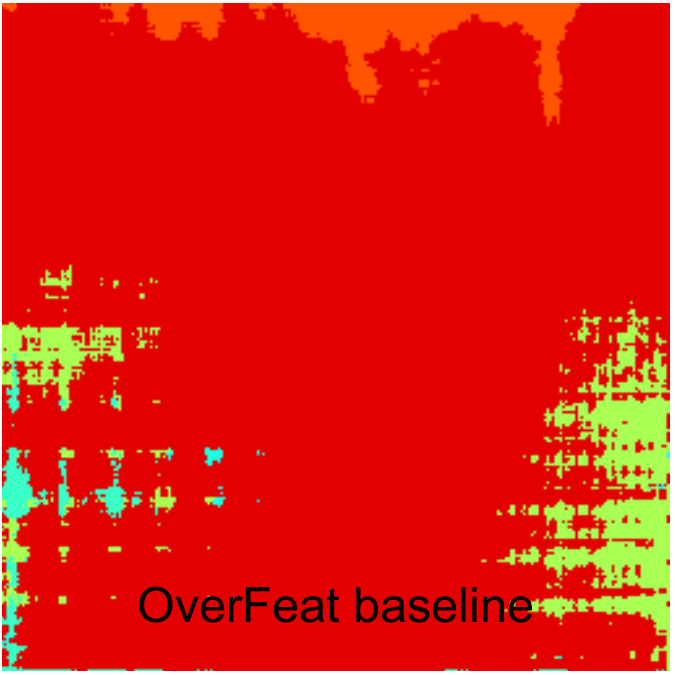}}
	& \imagetop{\includegraphics[width=2.32cm]{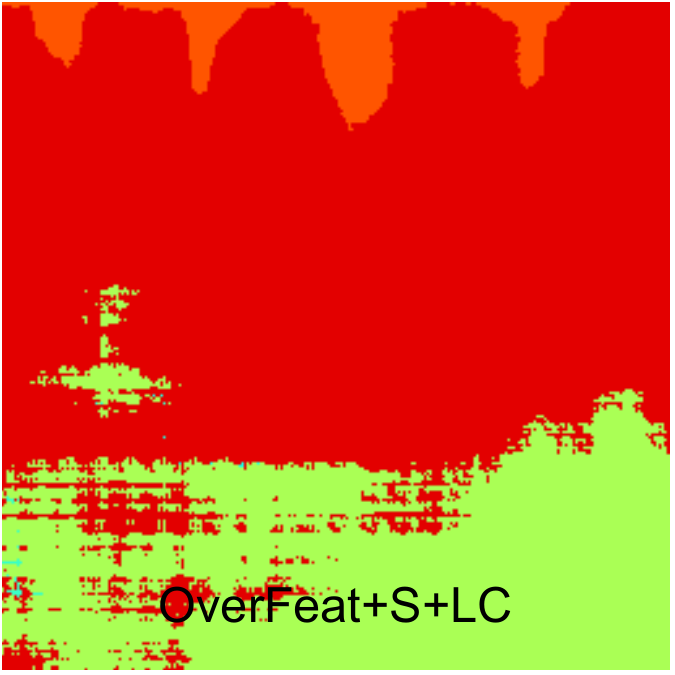}}
	&
	& \imagetop{\includegraphics[width=2.32cm]{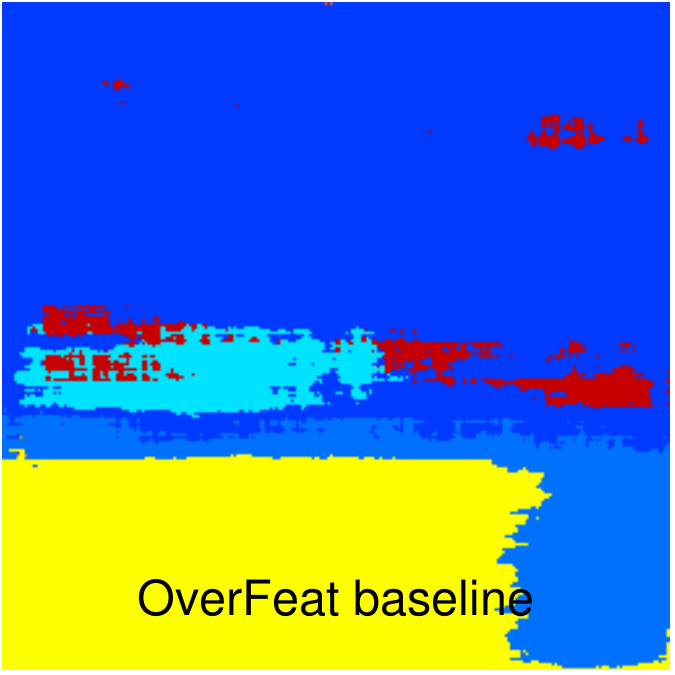}}
	& \imagetop{\includegraphics[width=2.32cm]{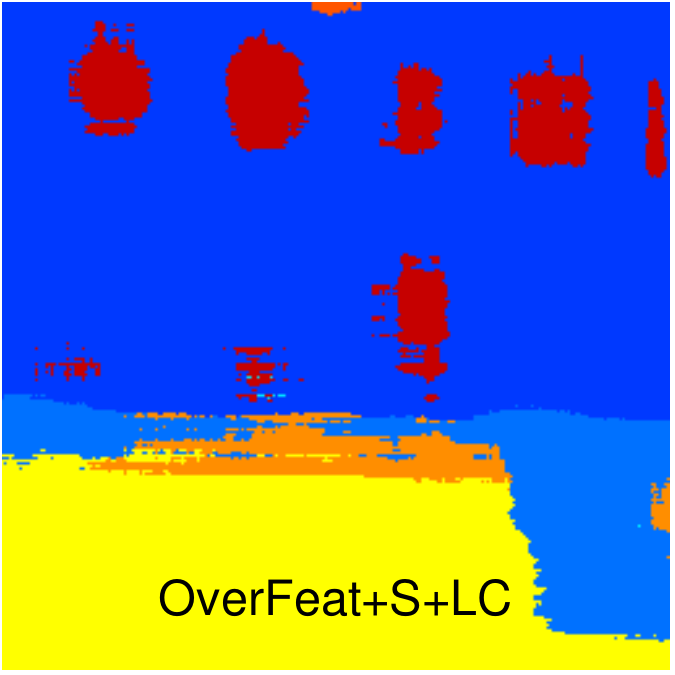}}			
	\end{tabular}
	\vspace{2pt}
	\caption{Some example scene labeling results on the SIFTFlow dataset by the proposed OverFeat+S+LC ($R=227$) and the OverFeat baseline models.  Best viewed in color.}\label{fig:siftflow_many_result}
\end{figure*}

\subsection{Results on the Stanford background dataset}
\label{subsec:stanford_results}
The Stanford background dataset contains $715$ images of outdoor scenes composed of $8$ classes. Each image have approximated $320\times 240$ pixels, where at least one foreground object is presented. Following \cite{socher2011parsing}, we selected $572$ images as the training set and the rest $143$ images as the test set.
Since the Stanford background dataset does not provide meaningful scene names for each image, we created label hierarchy via only label map clustering. Each original class is divided into $3$ to $6$ subclasses based on the criterion detailed in Section  \ref{subsubscn:label_map_based_cluster}, and $30$ subclasses are obtained in total at last.
We built $2$ sets of label hierarchy by setting $R=129$ and $R=227$, respectively.
\begin{table}[t]
\centering
\begin{tabular}{|l|c|c|}
\hline
~~~~~~~~~~~~~~~~~~Method & Per-pixel & Per-class \\ \hline \hline
\cite{gould2009decomposing}     & 0.764     & n/a       \\
\cite{tighe2010superparsing}    & 0.775     & n/a     \\
\cite{socher2011parsing}    & 0.781     & n/a     \\
\cite{lempitsky2011pylon}  & 0.819     & 0.724     \\
\cite{farabet2013learning}                      & 0.814     & 0.76    \\ 
\cite{pinheiro2013recurrent}          & 0.802     & 0.699     \\ 
\cite{sharma2015deep}         & 0.823     & 0.791    \\
\cite{wang2016learnable} & 0.871 & 0.837 \\ \hline \hline
OverFeat baseline             & 0.837     & 0.744     \\
OverFeat+S+IC ($R = 227$)           & 0.838     & 0.751     \\
OverFeat+S+LC ($R=129$)    & 0.841    & 0.75    \\
OverFeat+S+LC ($R=227$)    & 0.841   &  0.743    \\
OverFeat+H+LC ($R=227$)  & 0.842   & 0.754    \\ \hline\hline 
FCN \citep{long2014fcn}  & 0.851 & 0.811 \\ 
FCN+S+LC ($R = 129$) & 0.88 & 0.838 \\
FCN+H+LC ($R = 129$) & \textbf{0.882} & \textbf{0.843} \\\hline 
\end{tabular}
\caption{Per-pixel and per-class accuracies on the Stanford background dataset by compared methods. (The best accuracy by a single model and the best overall accuracy are marked in bold. S = Sequantial strategy, H = Hierarchical loss, LC = Label map Clustering, IC = Image context Clustering).}\label{tbl:stanford_result}
\end{table}

We trained the OverFeat model with our proposed methods on the dataset. The accuracies by different methods or training strategies are shown in Table \ref{tbl:stanford_result}.  
We also added another baseline method for comparison, where the original classes were divided into subclasses based on image context instead of semantic context. More specifically, we use the GIST features to group training images into different clusters, and each pixel is then assigned a cluster center and its original class label as its subclass label. As shown by its result (OverFeat+S+IC), such a strategy cannot generate subclasses with clear semantic explanations and therefore leads to almost the same accuracy as the baseline model.

For the FCN pipeline, our models trained by both strategies outperform the state-of-the-art methods. FCN+S+LC has exactly the same network architecture as the baseline FCN but achieves better performance, which proves that the gains come from  better feature representations through the learning of label hierarchies, other than simply adding more parameters.

\subsection{Results on the Barcelona and LM+Sun datasets}
\label{subsec:barcelona_results}
We further evaluate our method on two larger and more challenging datasets. 
The Barcelona dataset \citep{tighe2010superparsing} consists of 13649 images (13370 training images and 279 test images) and 170 labels. Its test set consists of street scenes from Barcelona, while the training set ranges in scene type but has no street scenes from Barcelona.
The LM+Sun dataset \citep{tighe2010superparsing_jnl} contains 45676 images (45176 training images and 500 test images) and 232 labels. 
The test images are selected at random that have 90$\%$ of their pixels labeled. However, only $38.4\%$ of the pixels of the training images on average are labeled ($49.4\%$ for the Barcelona dataset). 
This is quite different from the SIFTFlow and Stanford Background datasets ($91.6\%$ and $99.5\%$, respectively). 

The large number of unlabeled pixels would make the clustering result of label map patches unreliable. If we simply do not count the unlabeled pixels, we may not be able to extract the true semantic contexts from the patch. We show such an example in Fig. \ref{fig:clustering_counterexample}. From top to bottom we show three image patches, their label maps and the normalized histograms.  The image patches in (a) and (b) both contain the catogories ``person'' and ``table''. However in label map (b), only the pixels of the person are labeled, while the pixels of the table and the box are not. If the unlabeled pixels are ignored, the clustering algorithm would group the patches in (b) and (c) into the same subclass, which is not very meaningful.
We tested label hierarchy while ignoring unlabeled pixels. The result of using such label hierarchy on the OverFeat model with our sequential finetuning strategy (OverFeat+S+LC ($R=129$) (Naive)) is shown in Table \ref{tbl:barcelona_result}, which is only a slight improvement over the baseline (per-pixel accuracy improvement around 0.4$\%$). This is because the clustering result based on the incomplete label map is not very meaningful. The network cannot obtain strong supervision from the semantic contexts effectively. 

We proposed a preprocessing step before creating the label hierarchy. We first use the baseline model to predict the labels of the unlabeled pixels in the training images. For computational efficiency, the pixelwise forward propagation algorithm \citep{li2014highly} is adopted. Only the training images that have fewer than 90$\%$ pixels labeled are processed. This leaves us 8673 and 24682 training images in the Barcelona and LM+Sun datasets. Then the unlabeled pixels in the label maps are replaced with the predictions by the baseline model to create the label hierarchy via label map clustering. All the experiments are then carried on based on the new label maps.

\def\imagetop#1{\vtop{\null\hbox{#1}}}
\begin{figure}[t]
	\centering
	\setlength{\tabcolsep}{0.3em}
	\begin{tabular}{c@{\hspace{-1mm}}c@{\hspace{1mm}}c@{\hspace{1mm}}c}
	& \imagetop{\includegraphics[width=0.32\linewidth]{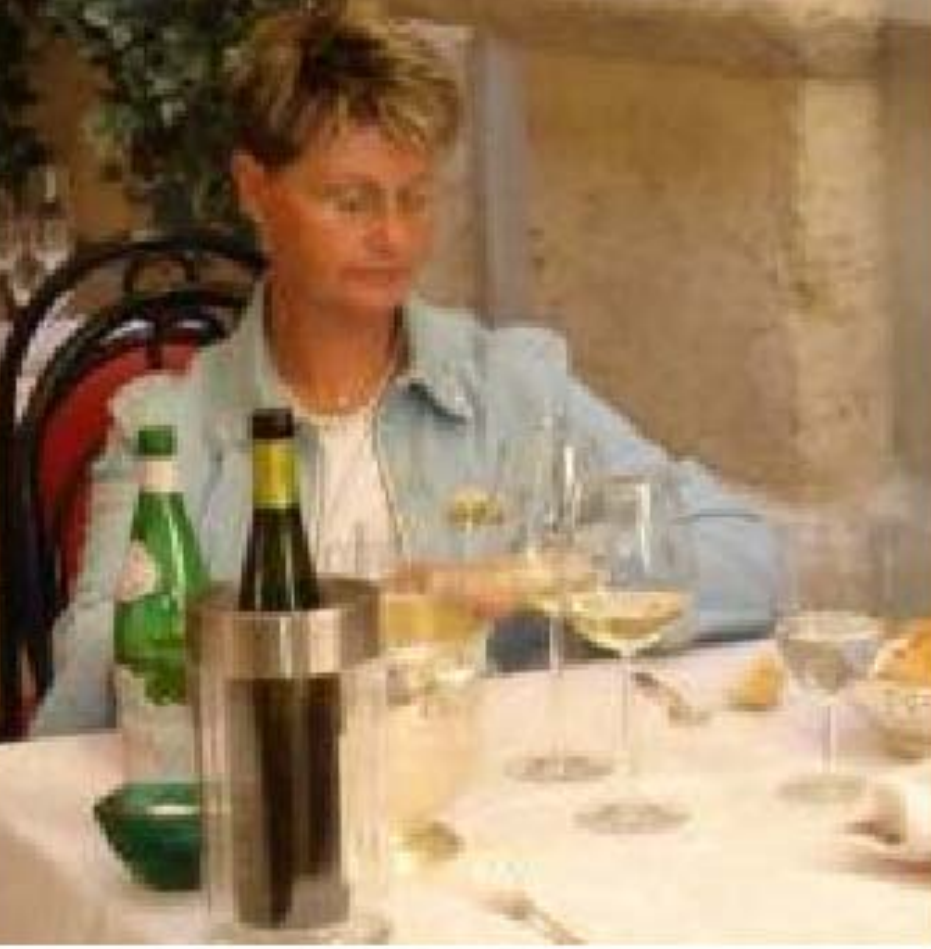}}
	& \imagetop{\includegraphics[width=0.32\linewidth]{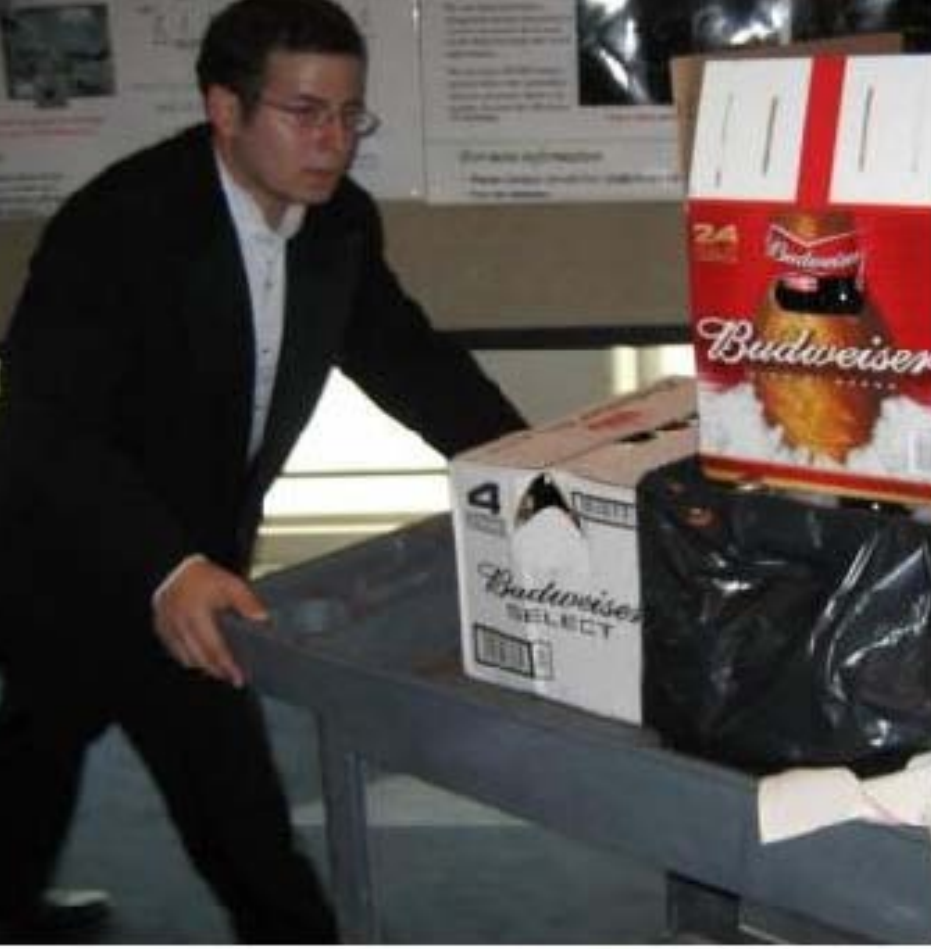}}
	& \imagetop{\includegraphics[width=0.32\linewidth]{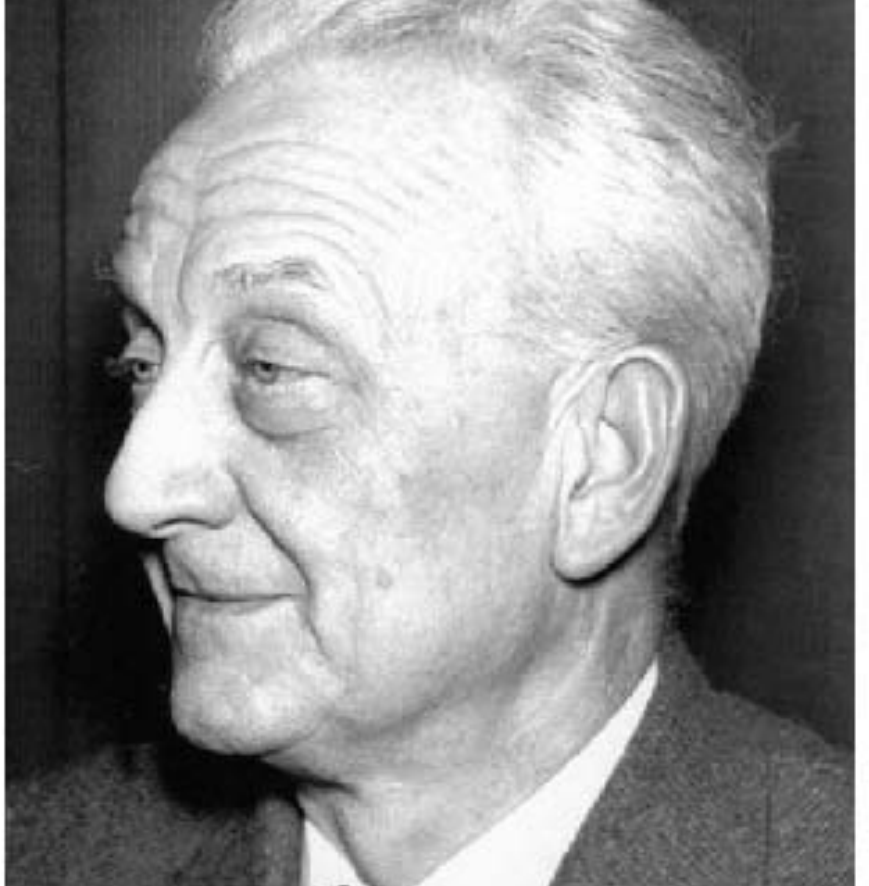}}\\
	& \imagetop{\includegraphics[width=0.32\linewidth]{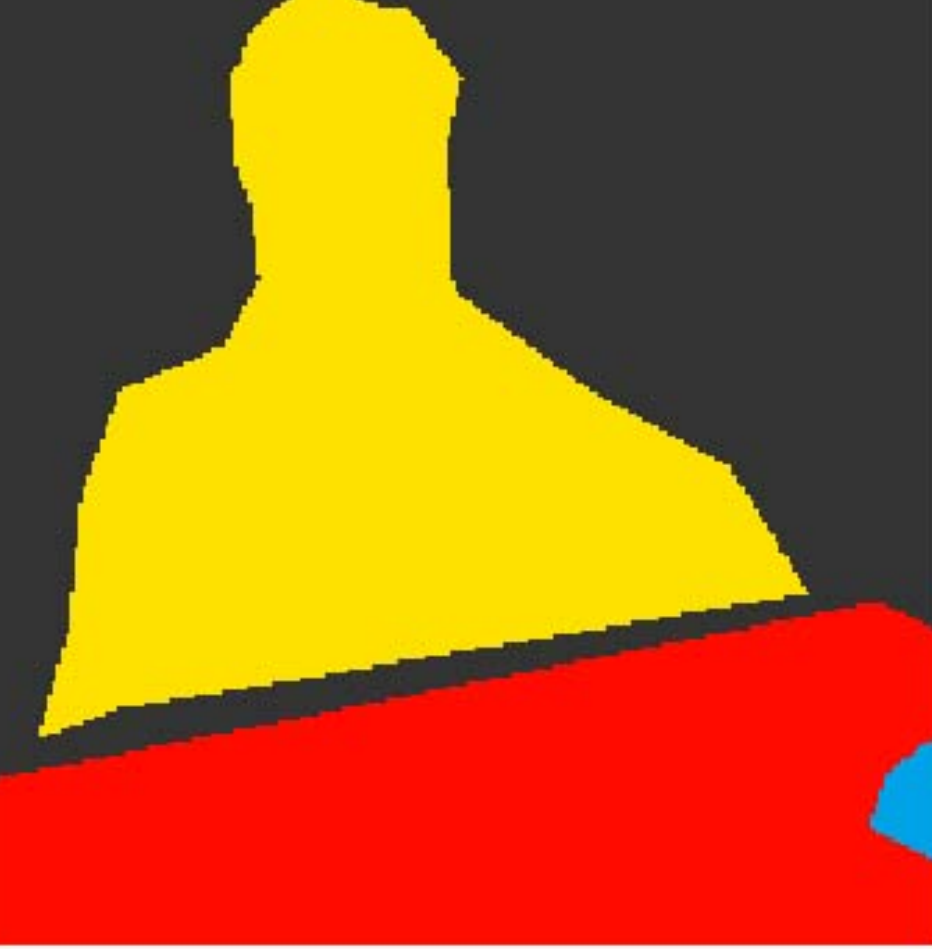}}
	& \imagetop{\includegraphics[width=0.32\linewidth]{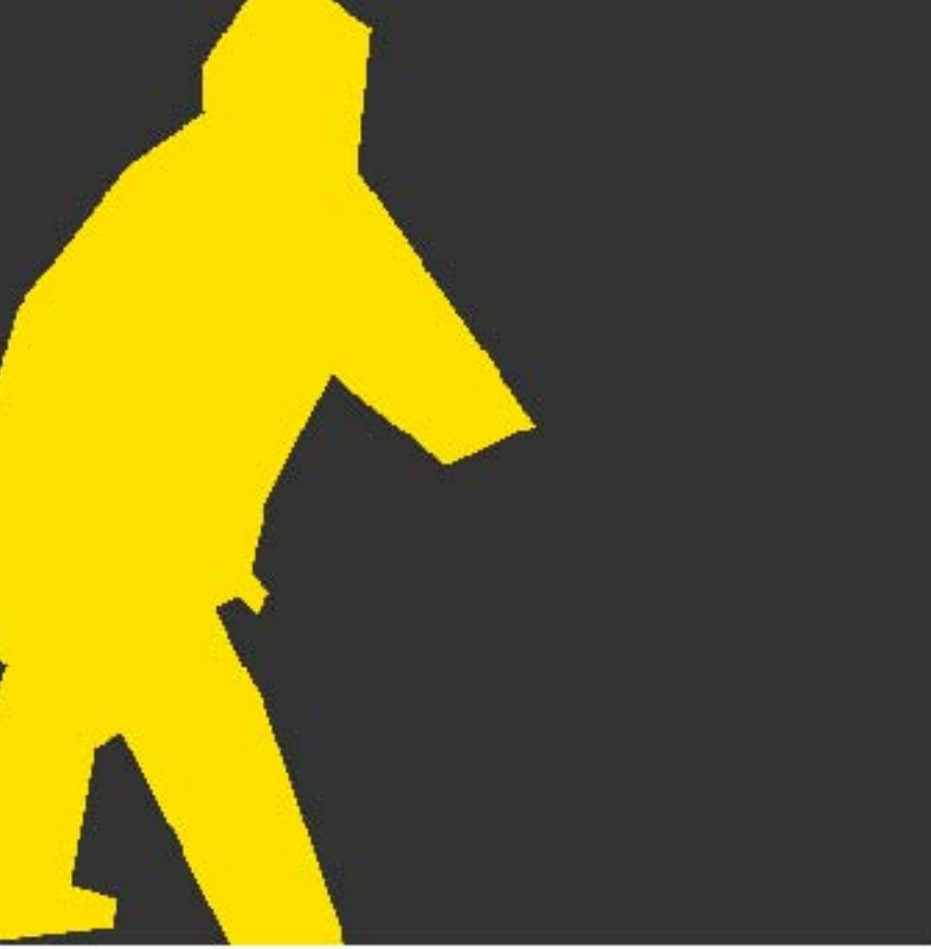}}
	& \imagetop{\includegraphics[width=0.32\linewidth]{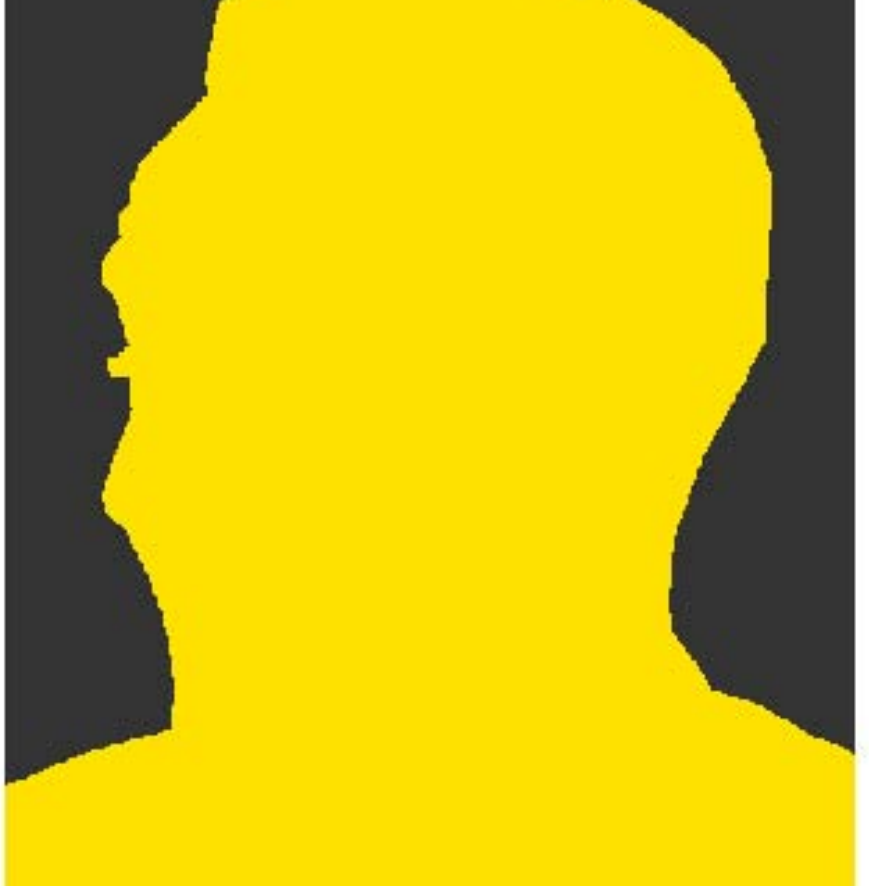}}\\
	& \imagetop{\includegraphics[width=0.32\linewidth]{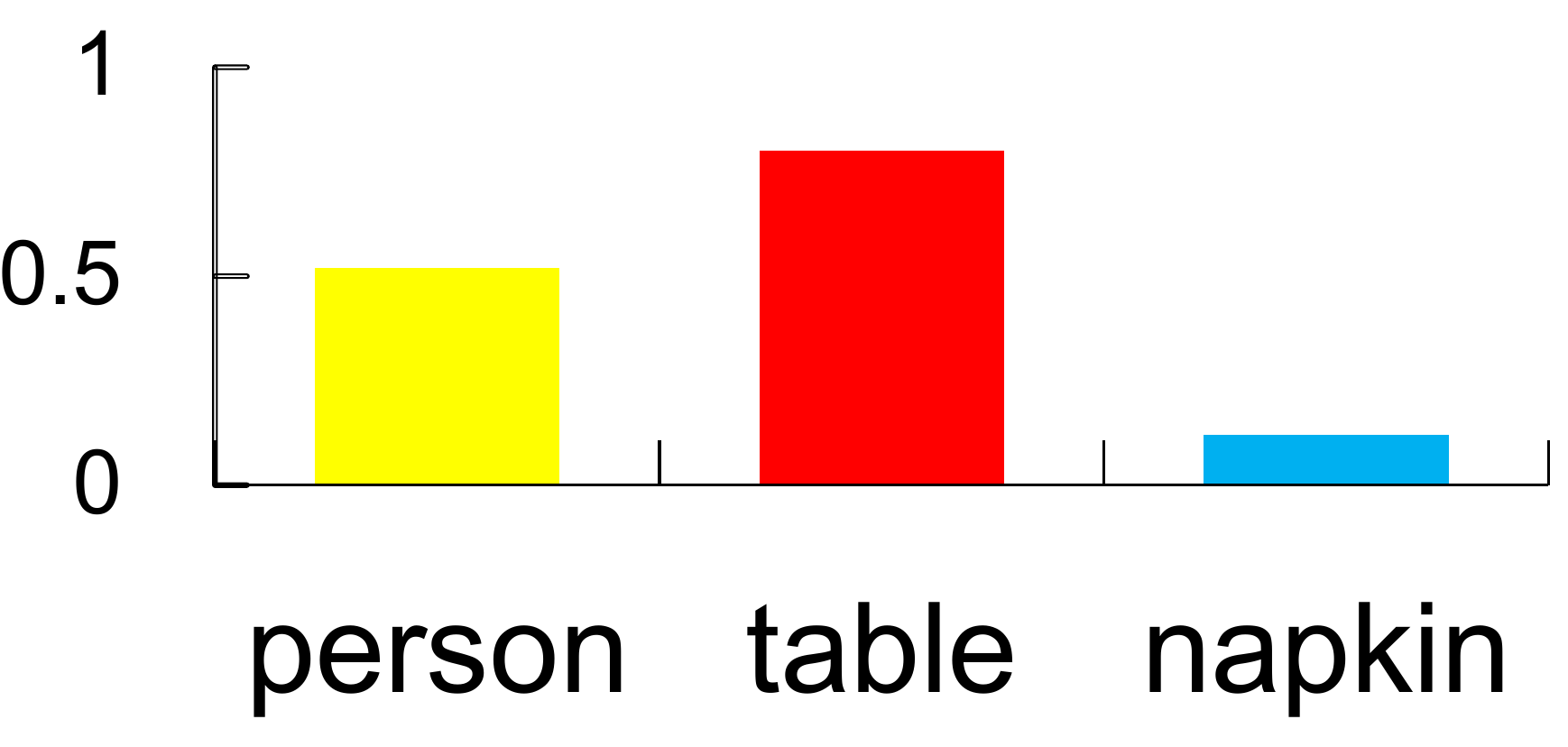}}
	& \imagetop{\includegraphics[width=0.32\linewidth]{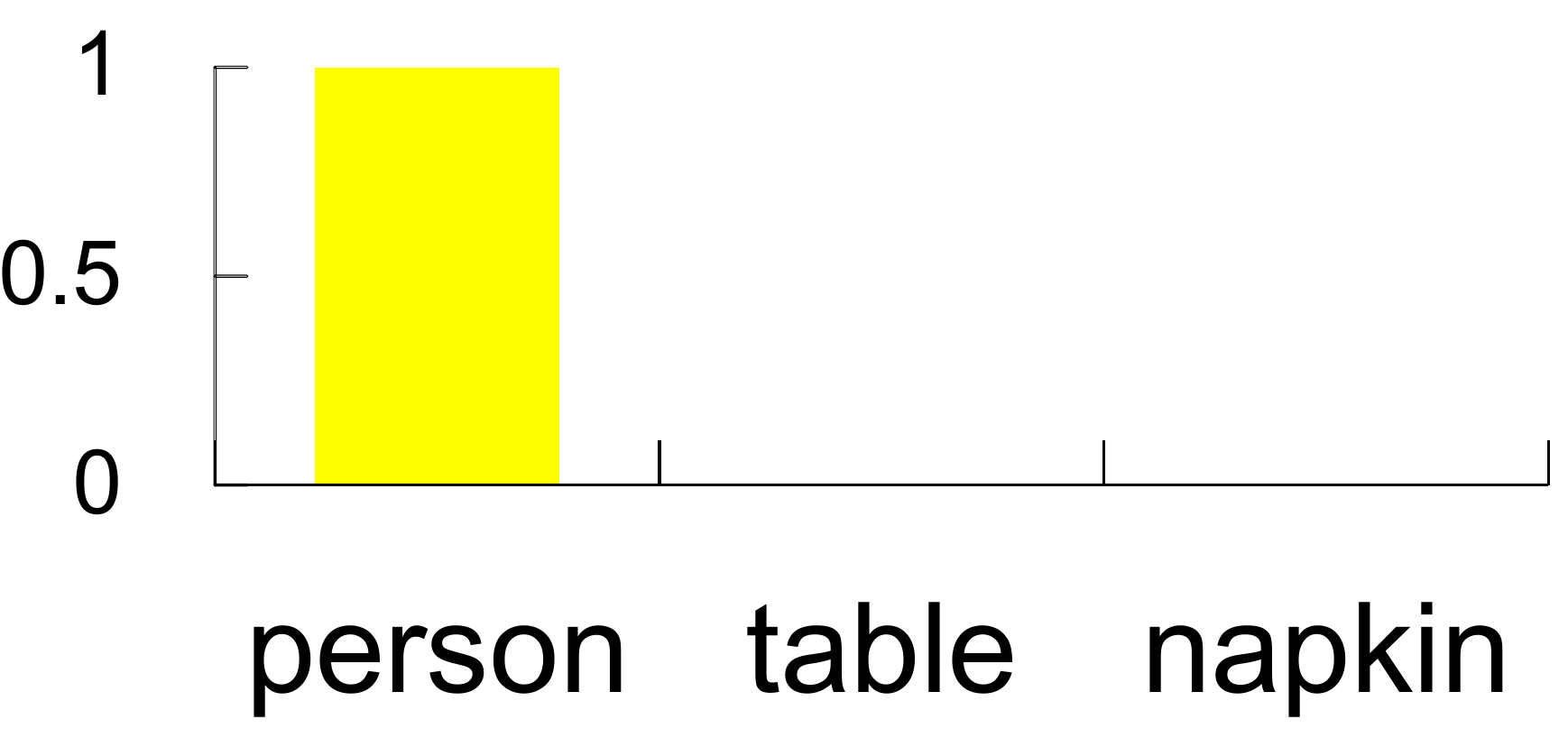}}
	& \imagetop{\includegraphics[width=0.32\linewidth]{./hist4.pdf}}\\	
	& \imagetop{\includegraphics[width=0.32\linewidth]{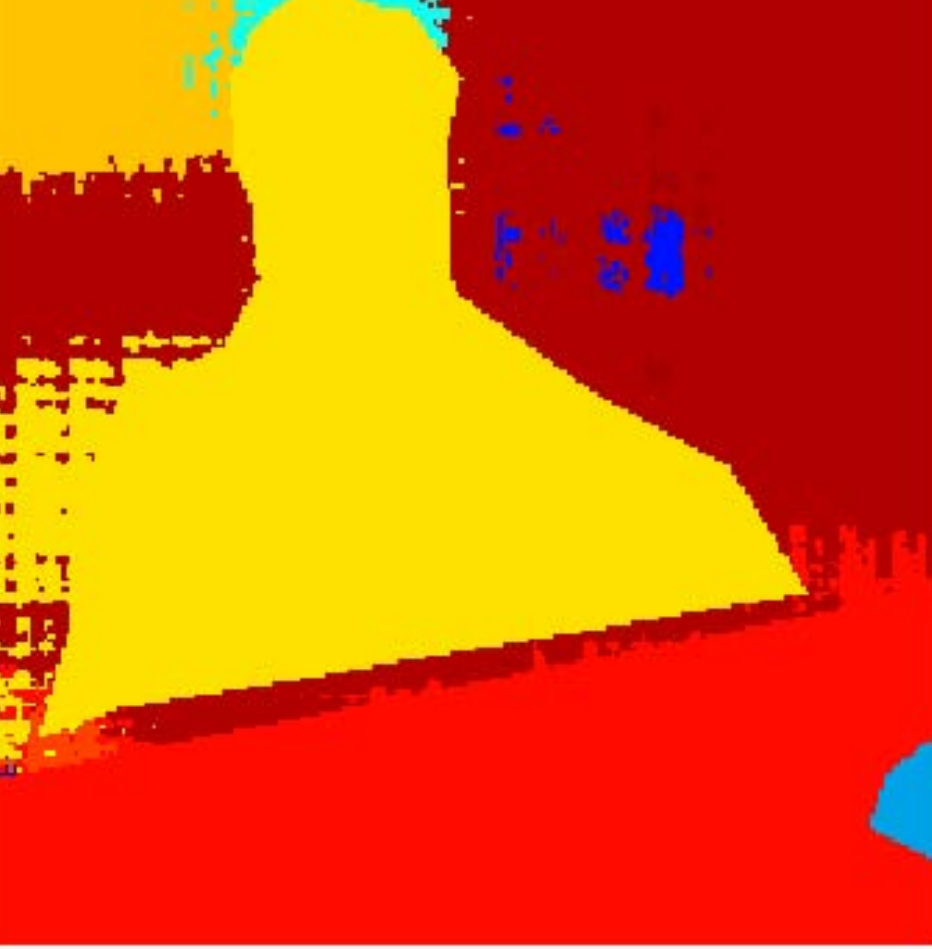}}
	& \imagetop{\includegraphics[width=0.32\linewidth]{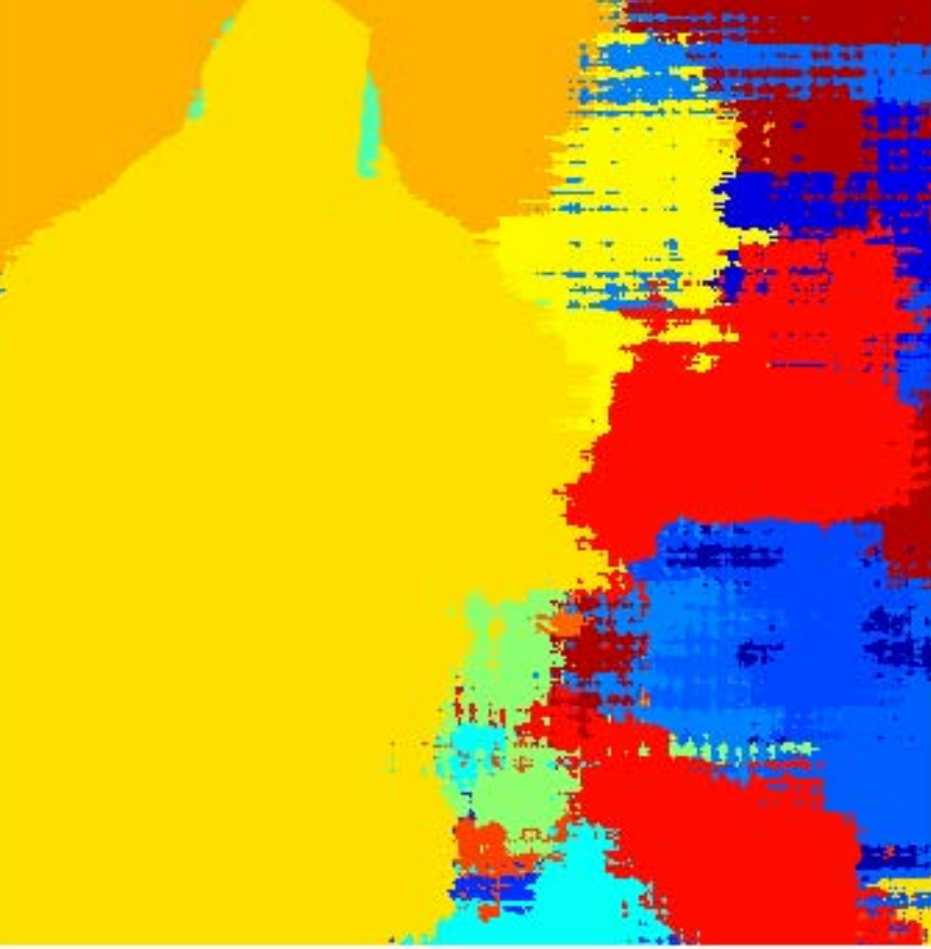}}
	& \imagetop{\includegraphics[width=0.32\linewidth]{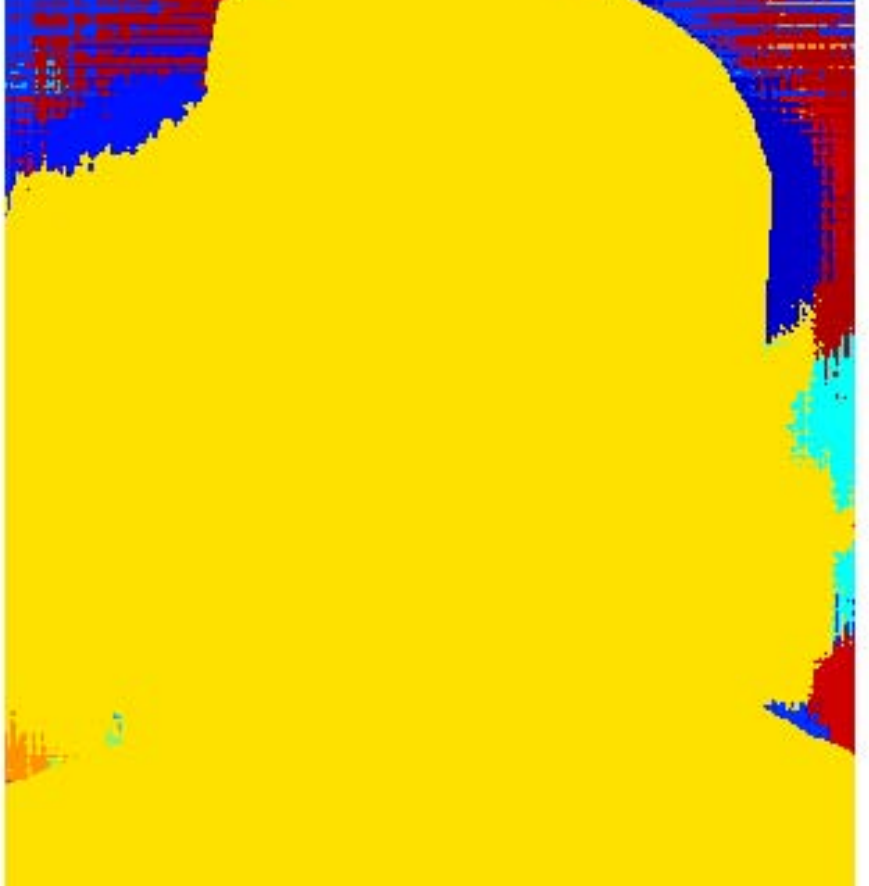}}\\	
	\end{tabular}
	\vspace{+2pt}
	(a)~~~~~~~~~~~~~~~~~~~~~(b)~~~~~~~~~~~~~~~~~~~~~(c)
   \caption{From top to bottom we show three image patches from the Barcelona dataset, their corresponding label maps, the normalized histograms and the updated label maps by the baseline model. Note the label map in (b) has only the pixels of the person labeled. While its appearance and semantic context is more similar to (a), it is grouped into the same cluster with (c). The baseline model successfully predicts the existence of the table in the updated label map, making the histogram of (b) closer to (a). Best viewed in color.}
   \label{fig:clustering_counterexample}
\end{figure}

\def\imagetop#1{\vtop{\null\hbox{#1}}}
\begin{figure*}[!ht]
	\centering
	\setlength{\tabcolsep}{0.3em}
	\begin{tabular}{c@{\hspace{-1mm}}c@{\hspace{-1mm}}c@{\hspace{1mm}}c@{\hspace{1mm}}c@{\hspace{1mm}}c@{\hspace{1mm}}c}
 & Image & Ground truth & OverFeat baseline & OverFeat+S+LC (R=129) &  \\
	\vspace{-0.2cm}
&  \imagetop{\includegraphics[width=0.19\linewidth]{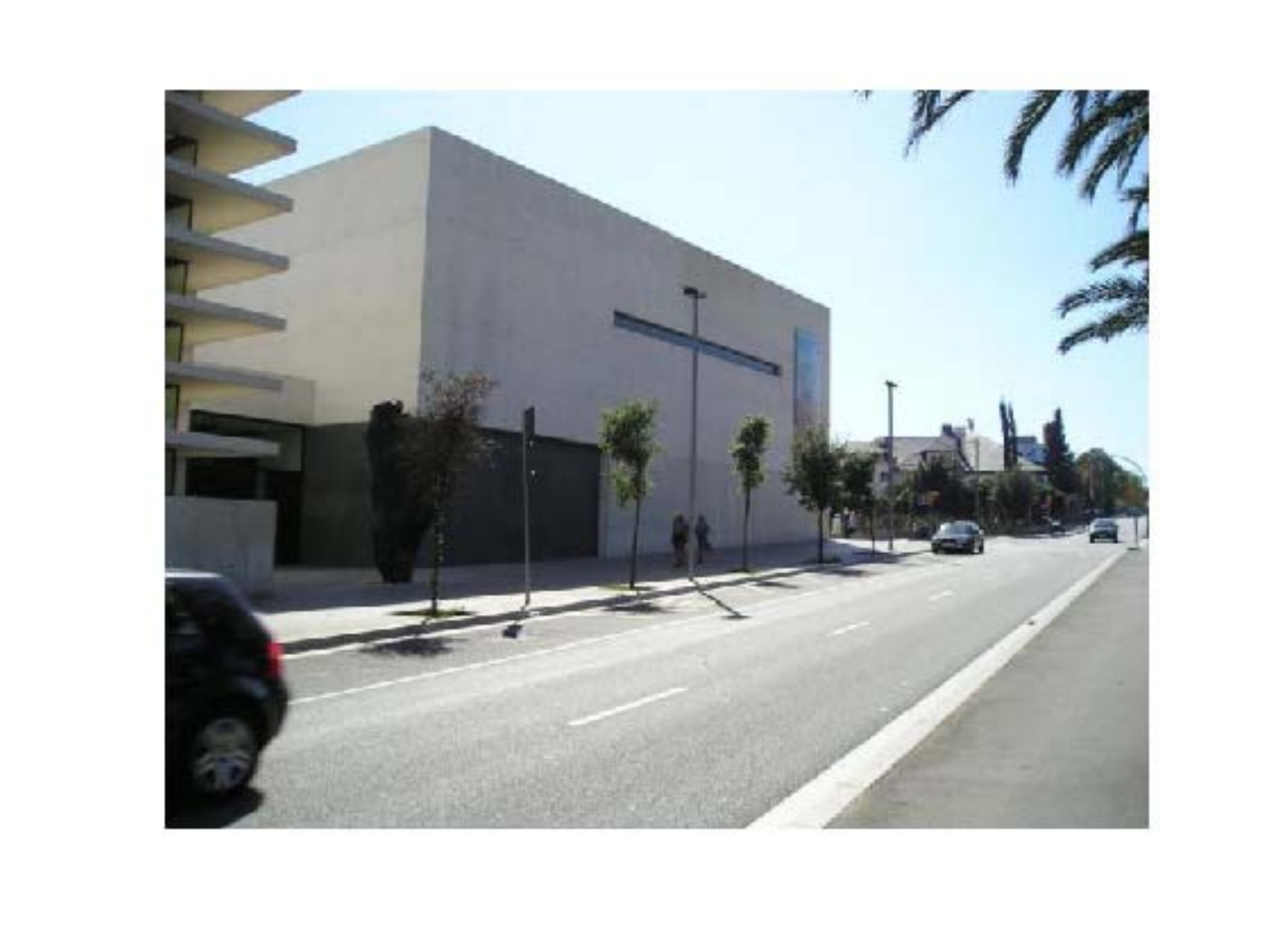}}
	& \imagetop{\includegraphics[width=0.19\linewidth]{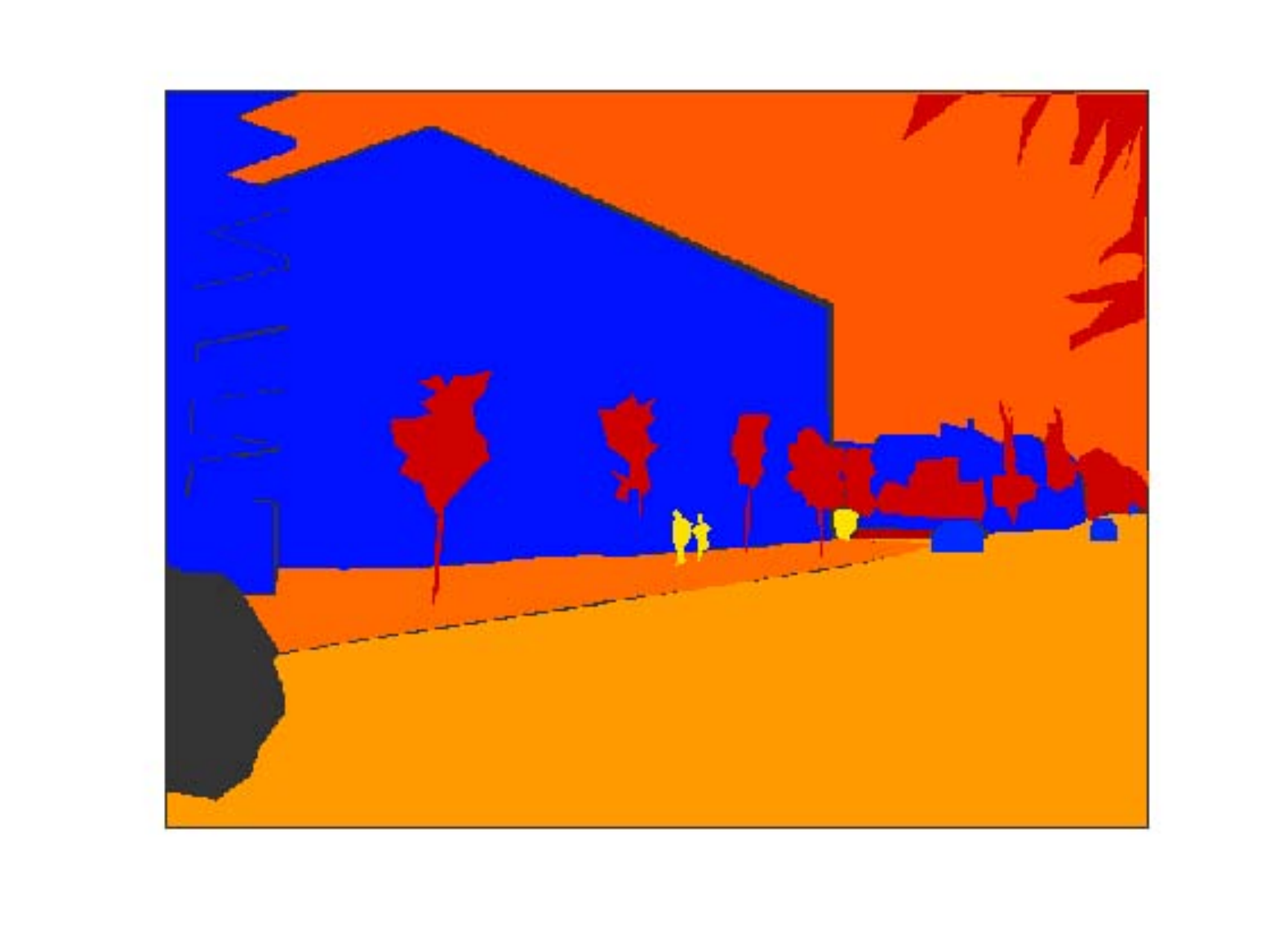}}
	& \imagetop{\includegraphics[width=0.19\linewidth]{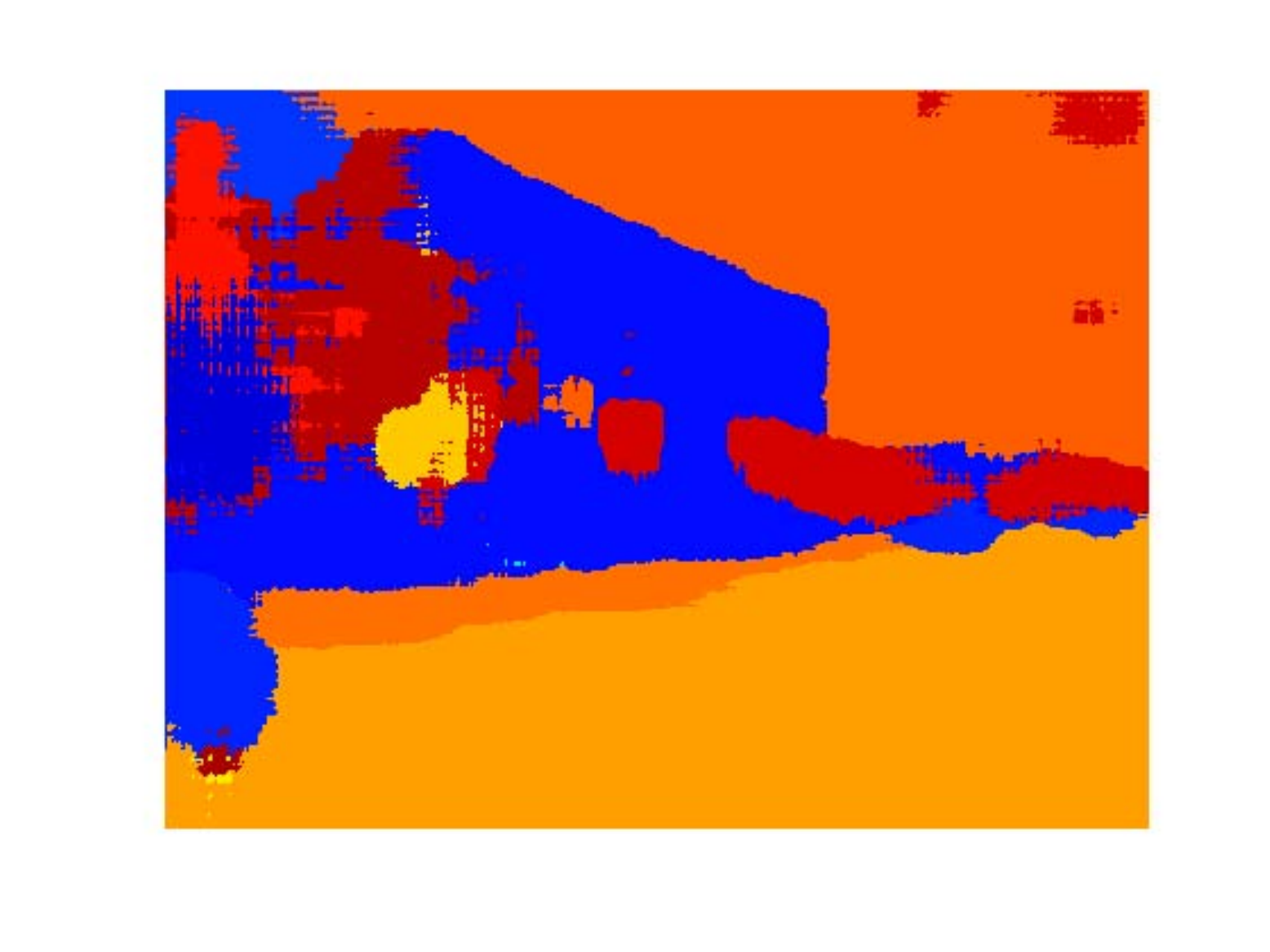}}
	& \imagetop{\includegraphics[width=0.19\linewidth]{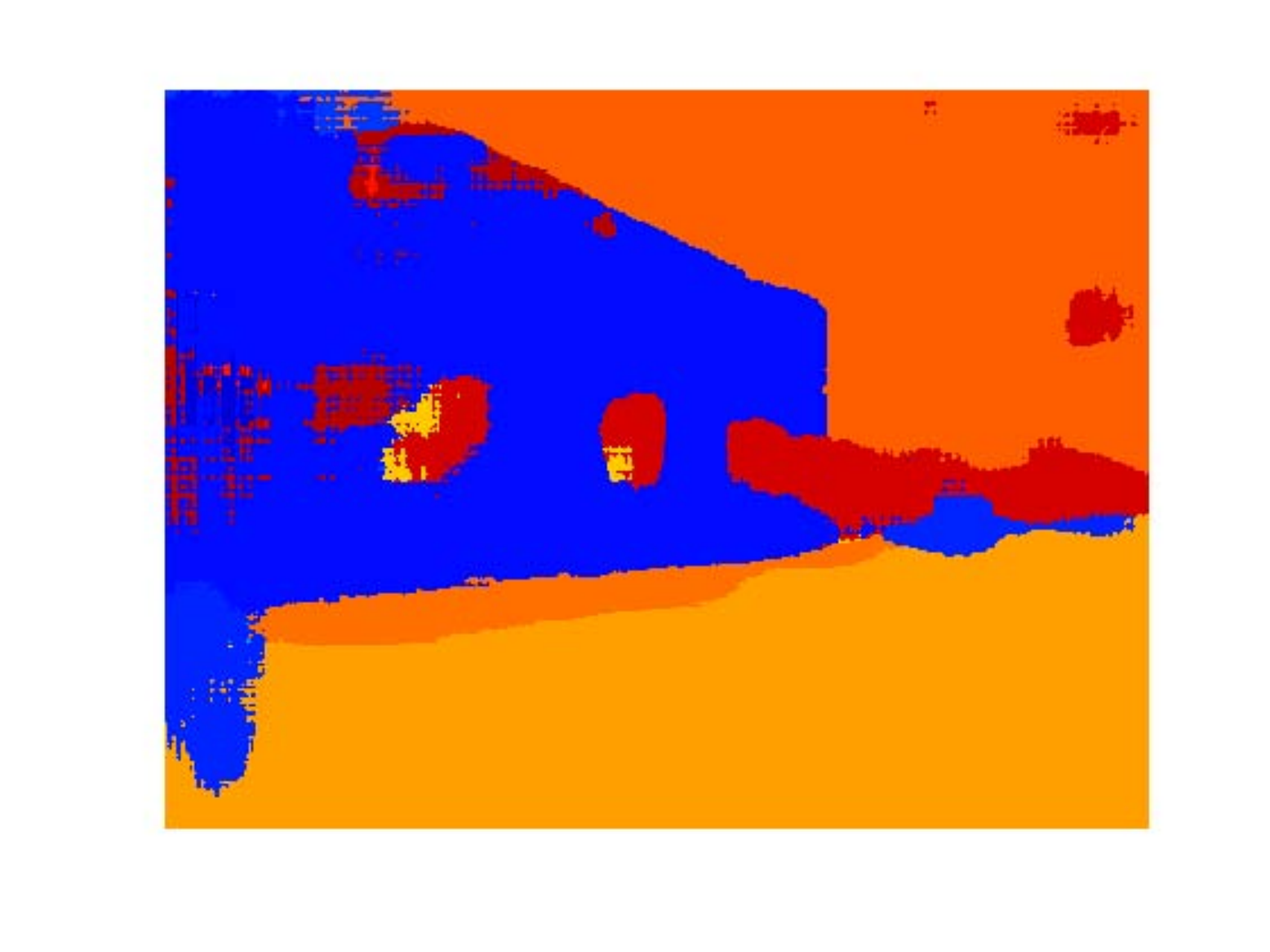}}
	& \imagetop{\includegraphics[width=0.07\linewidth]{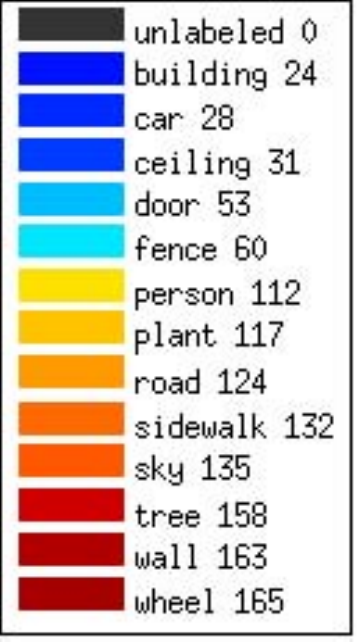}}	\\
	\vspace{-0.2cm}
& \imagetop{\includegraphics[width=0.19\linewidth]{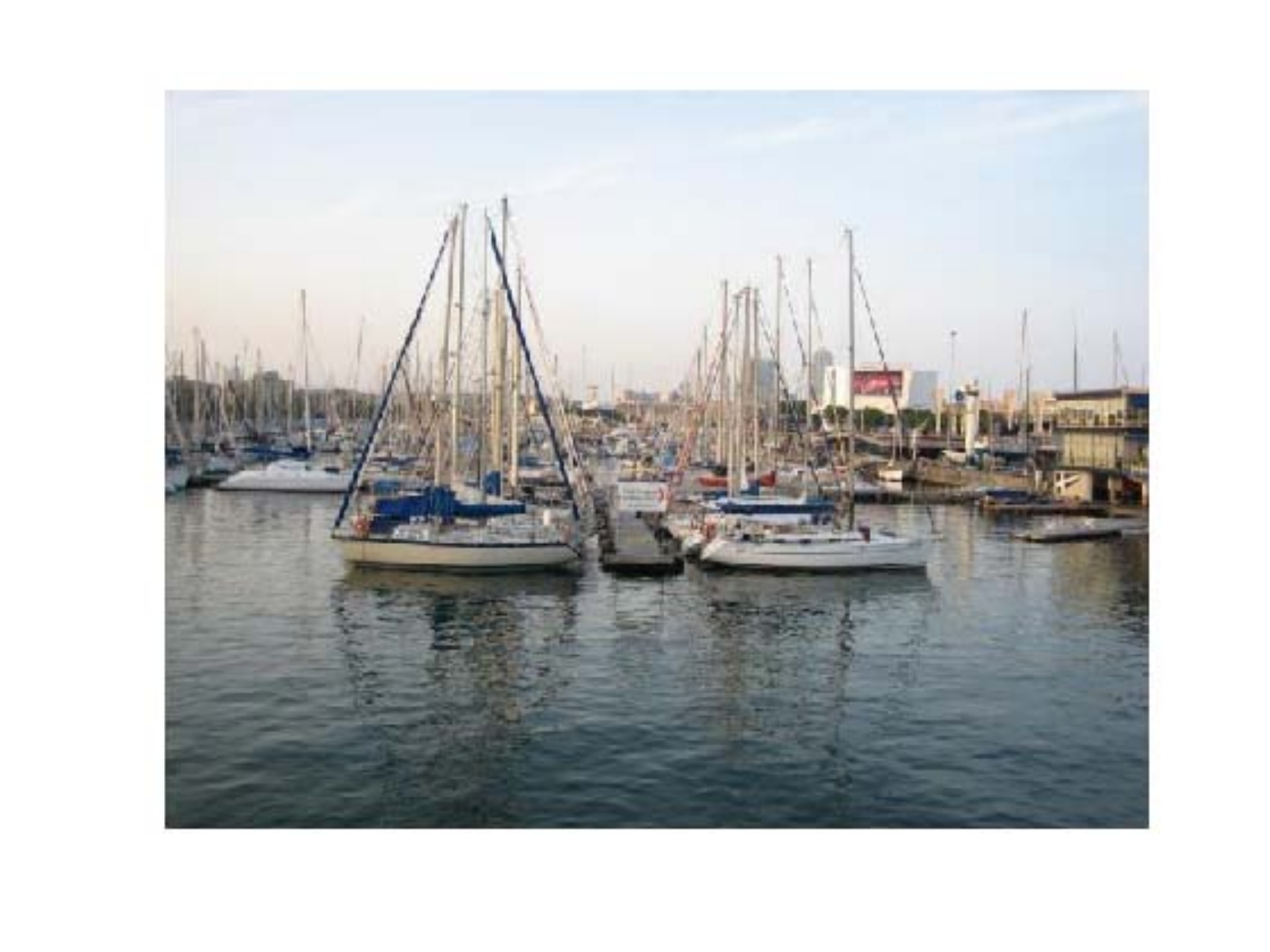}}
	& \imagetop{\includegraphics[width=0.19\linewidth]{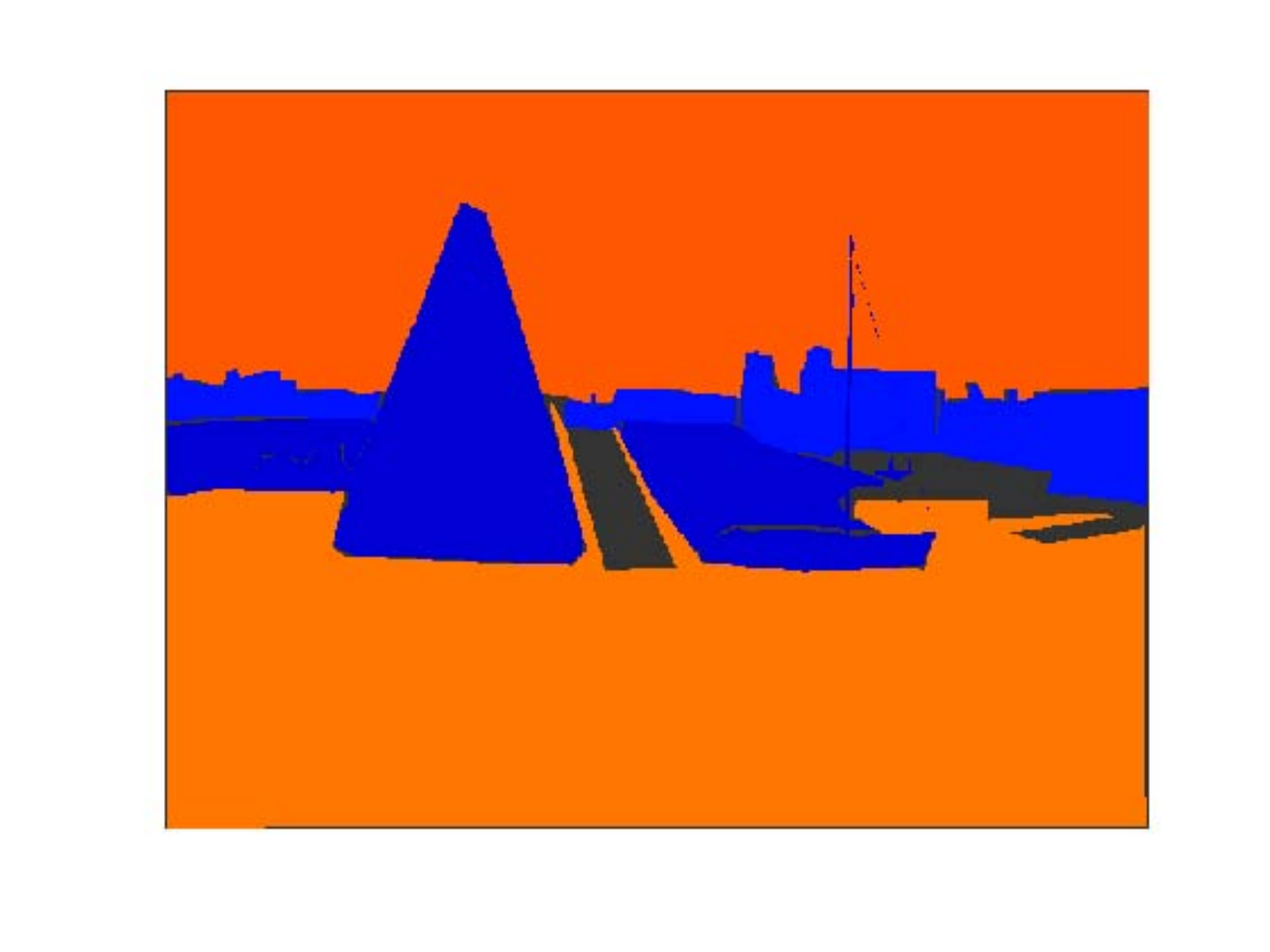}}
	& \imagetop{\includegraphics[width=0.19\linewidth]{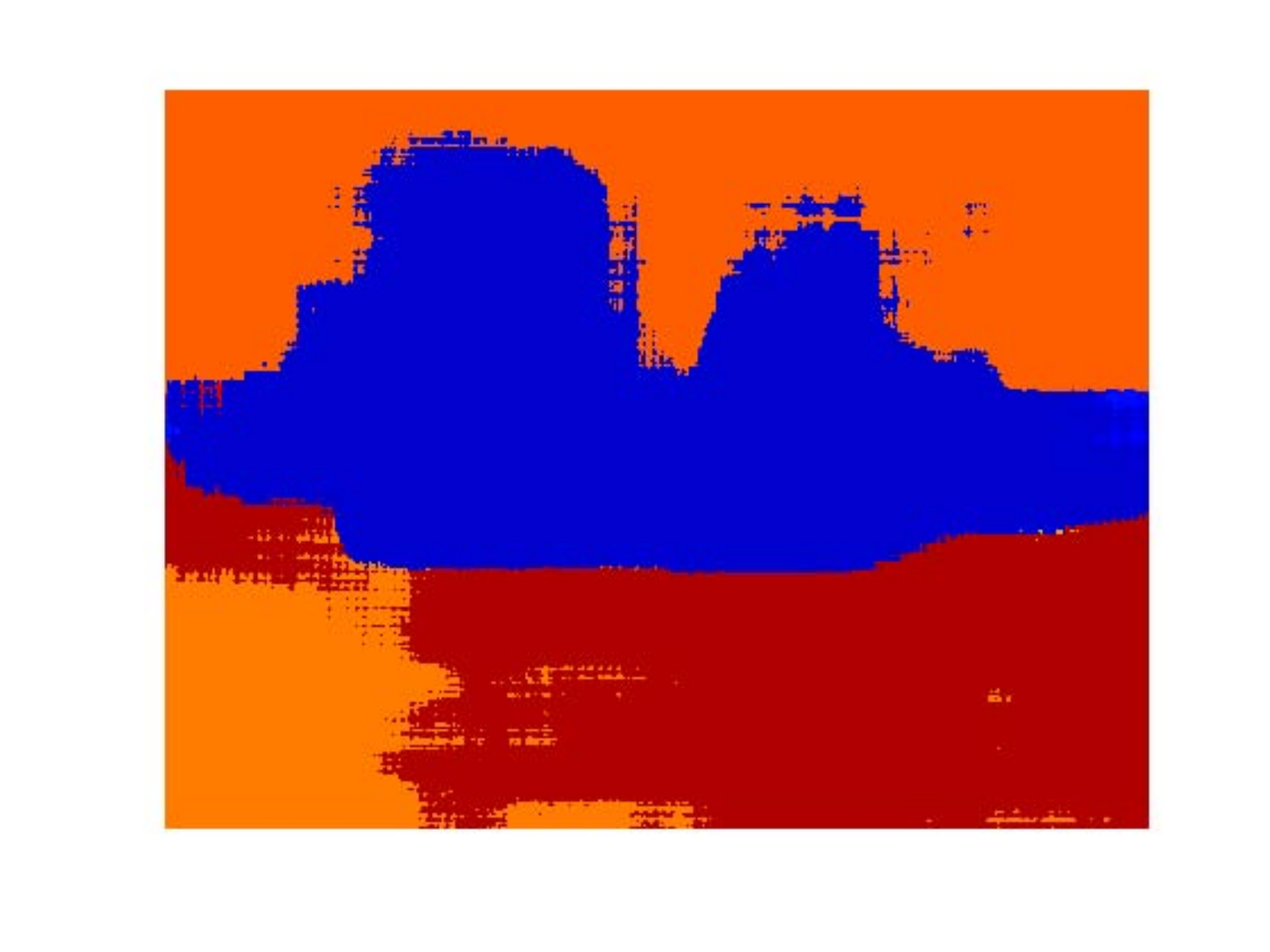}}
	& \imagetop{\includegraphics[width=0.19\linewidth]{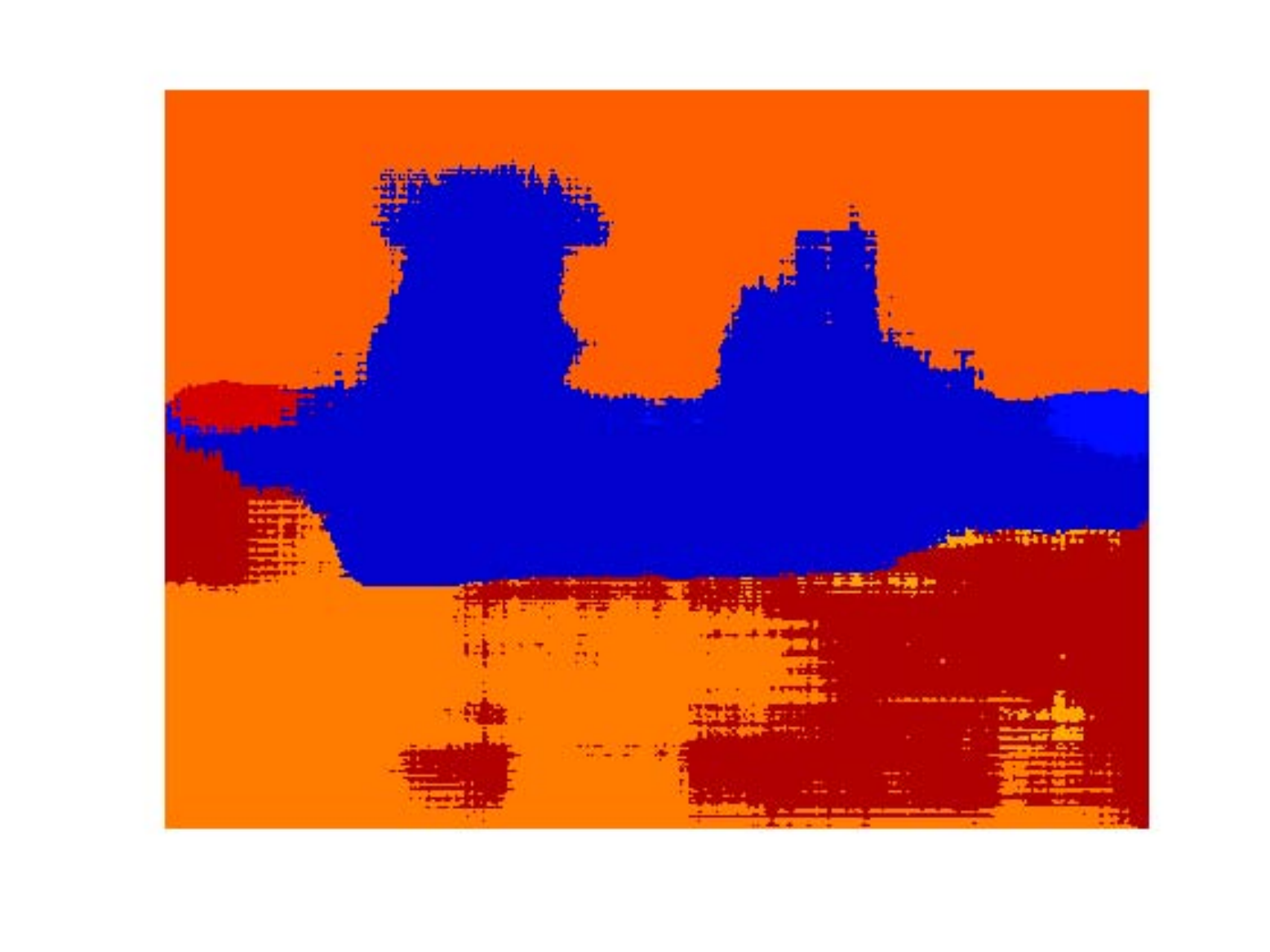}}
	& \imagetop{\includegraphics[width=0.07\linewidth]{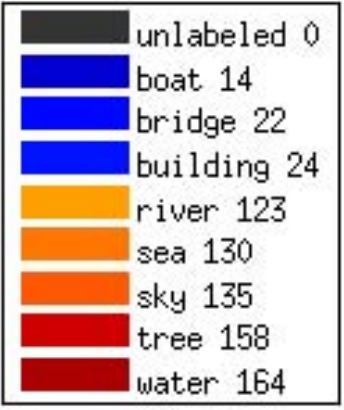}}	\\
	\vspace{-0.2cm}
&  \imagetop{\includegraphics[width=0.19\linewidth]{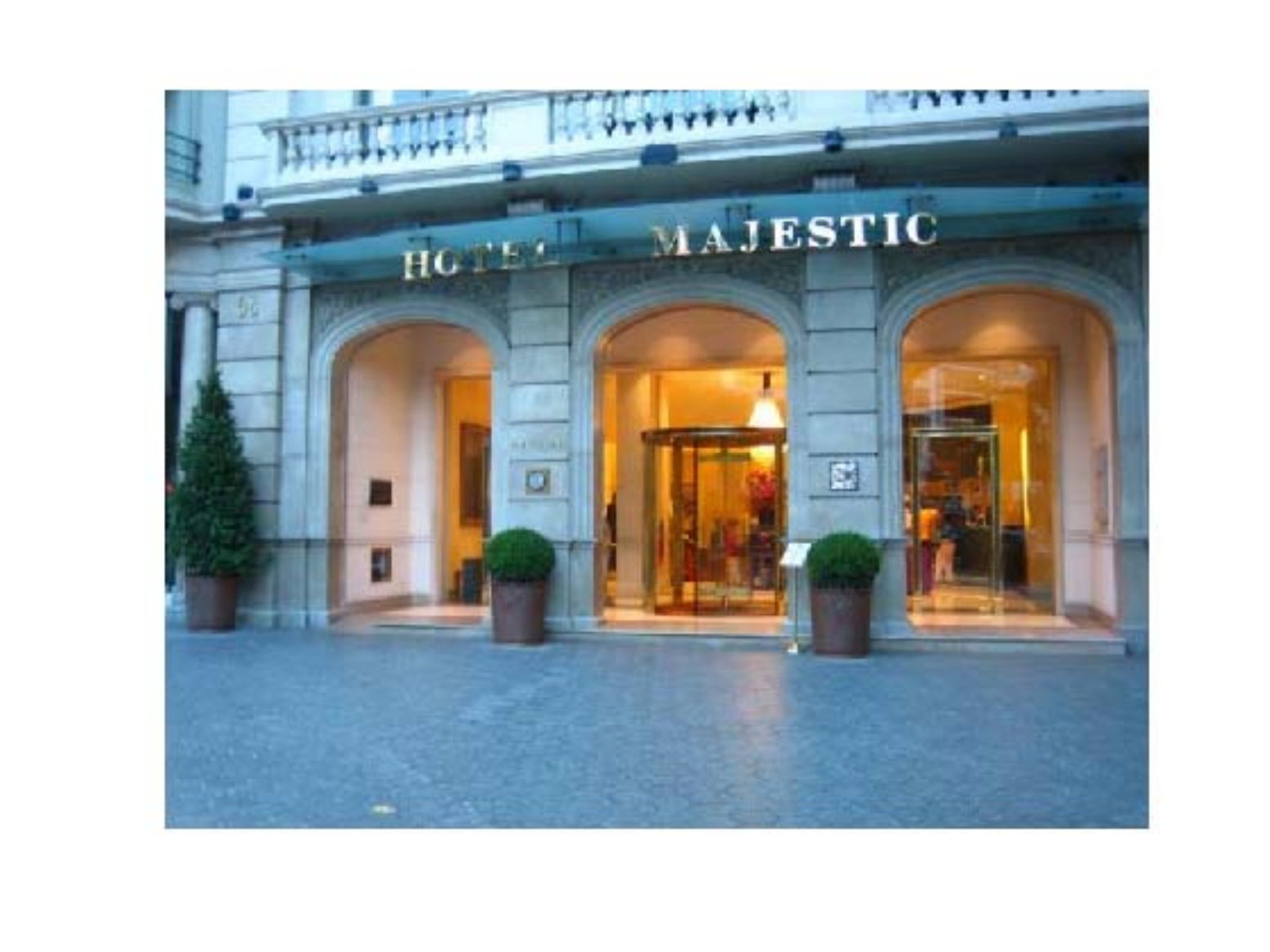}}
	& \imagetop{\includegraphics[width=0.19\linewidth]{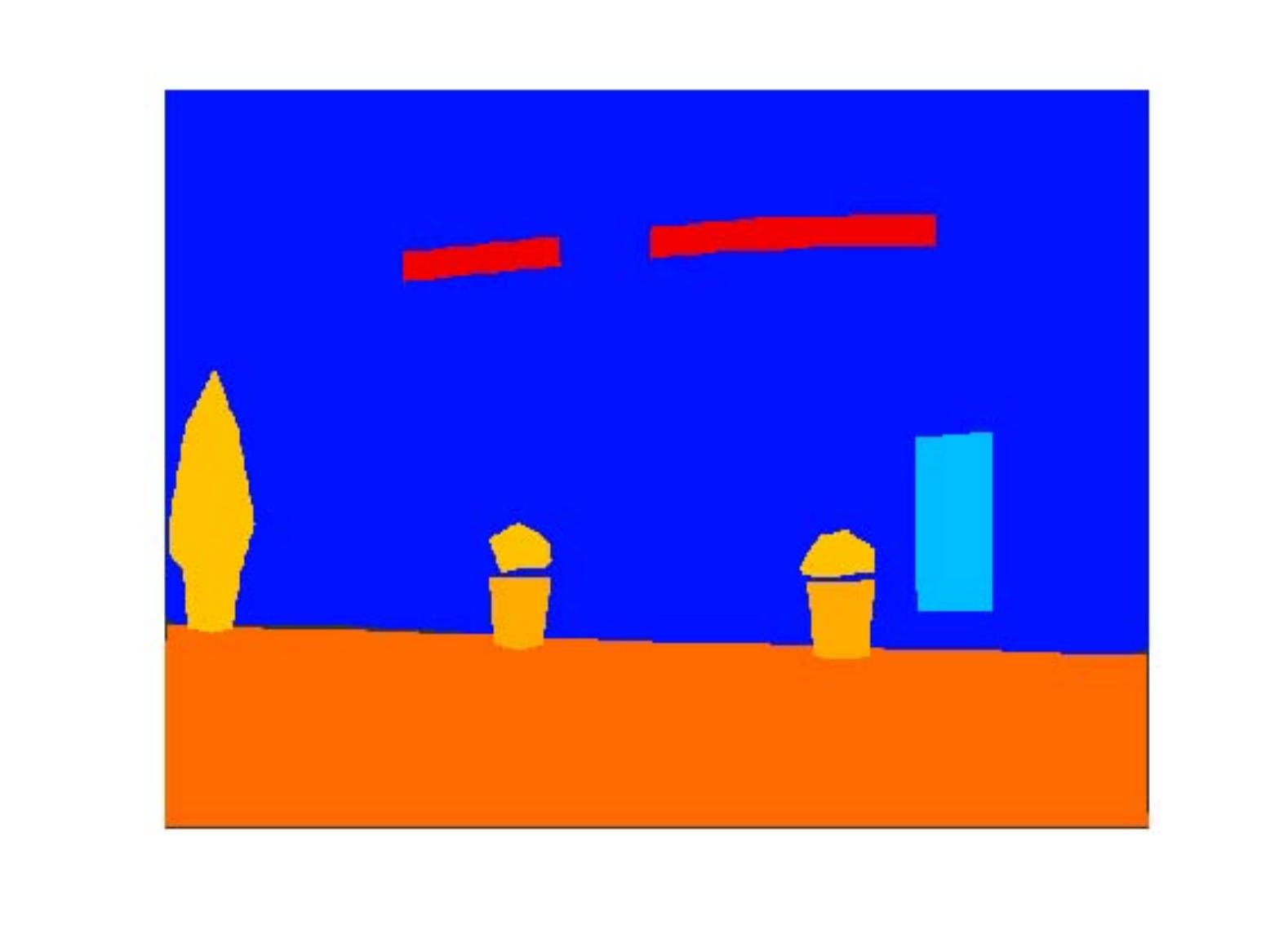}}
	& \imagetop{\includegraphics[width=0.19\linewidth]{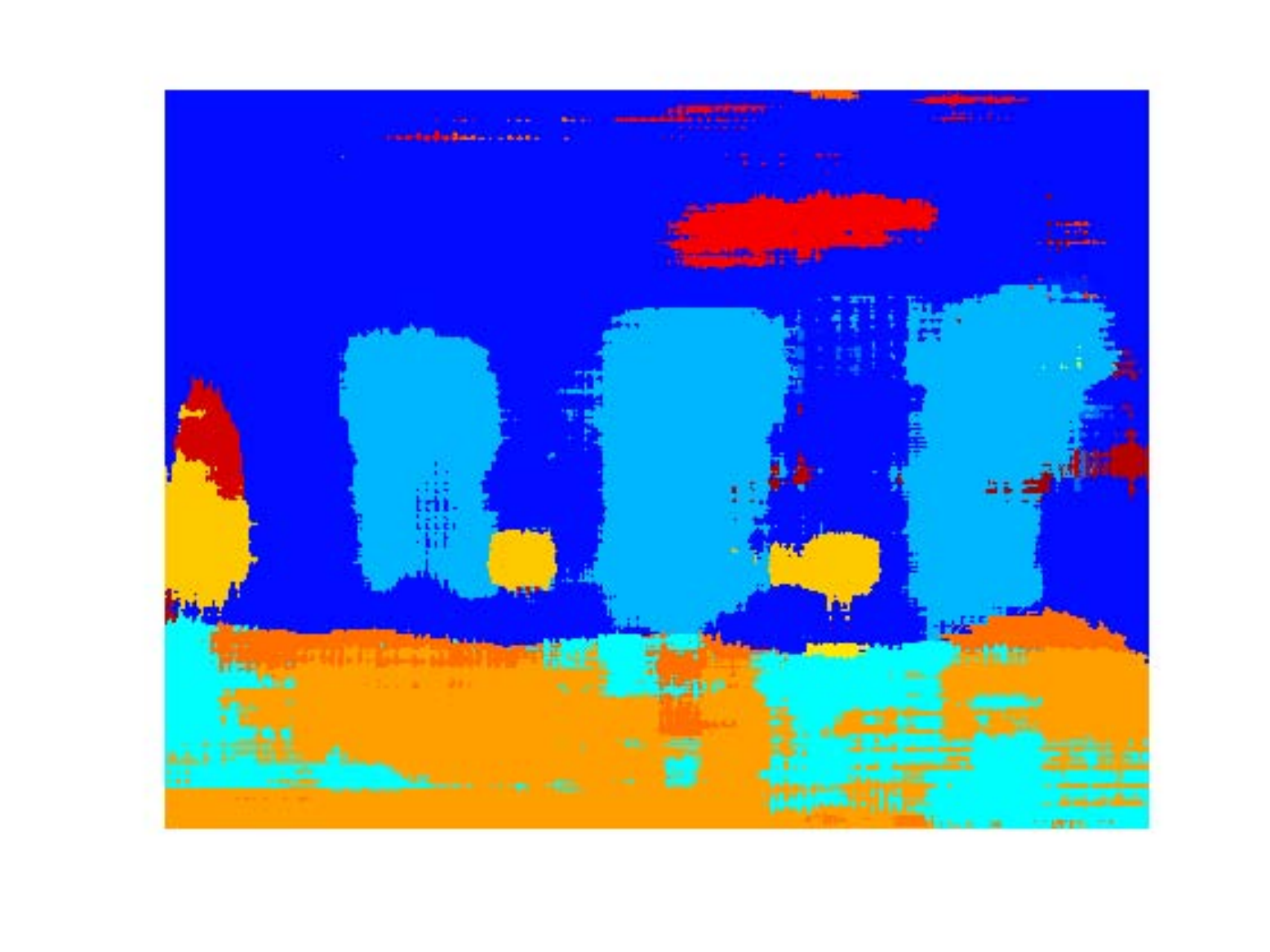}}
	& \imagetop{\includegraphics[width=0.19\linewidth]{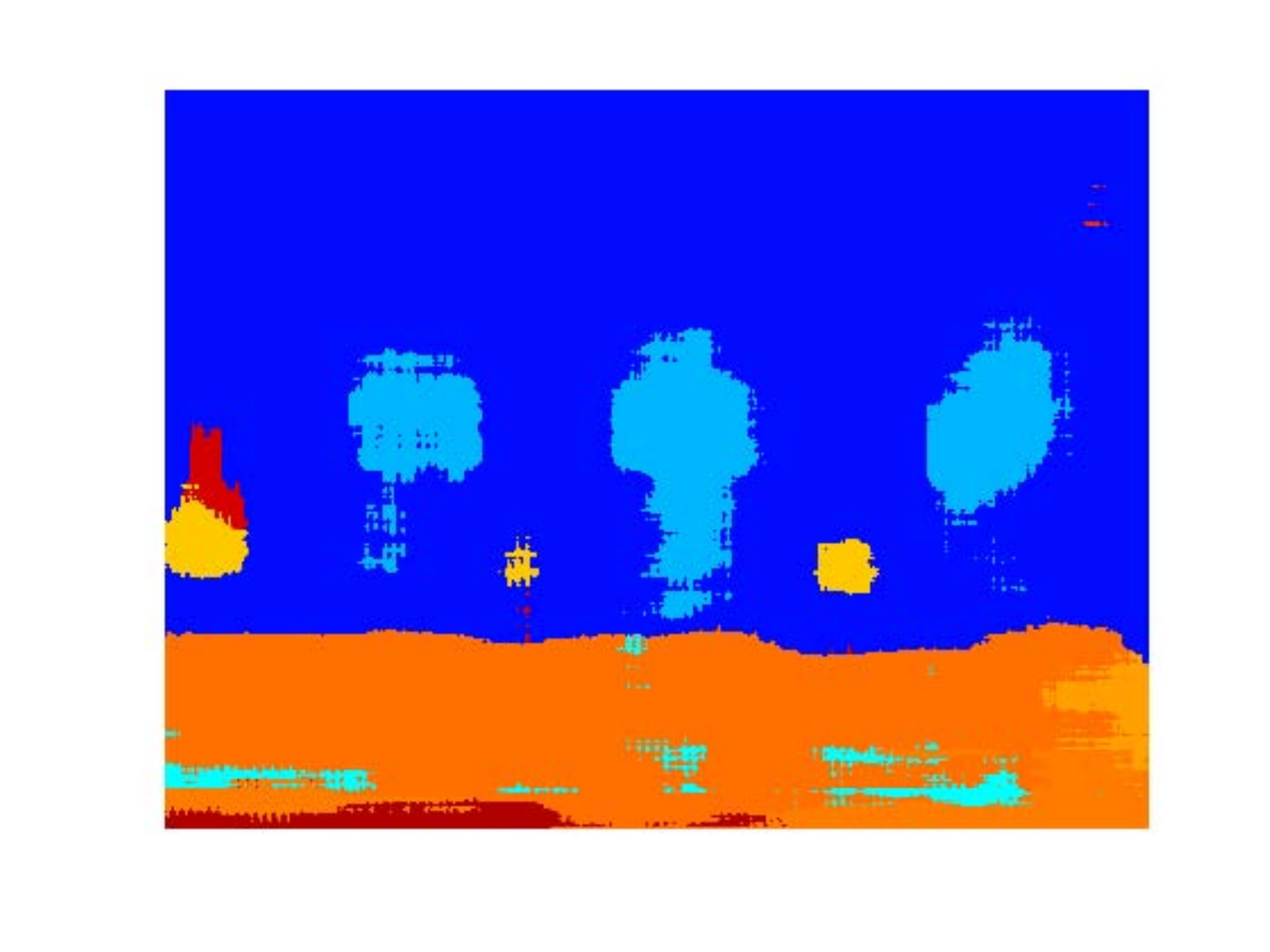}}
	& \imagetop{\includegraphics[width=0.07\linewidth]{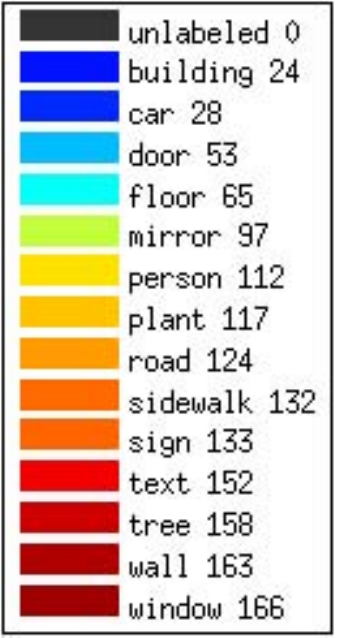}}	\\
	\vspace{-0.2cm}
& \imagetop{\includegraphics[width=0.19\linewidth]{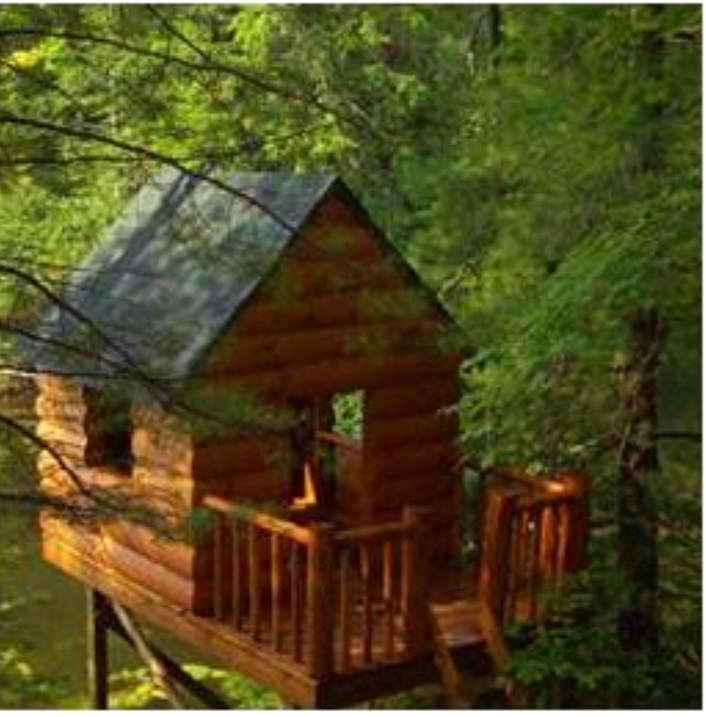}}
	& \imagetop{\includegraphics[width=0.19\linewidth]{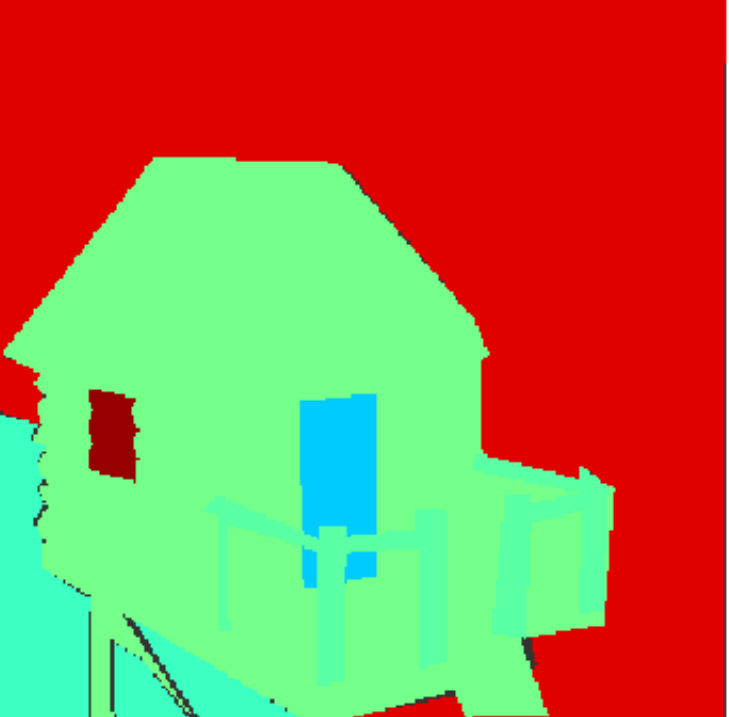}}
	& \imagetop{\includegraphics[width=0.19\linewidth]{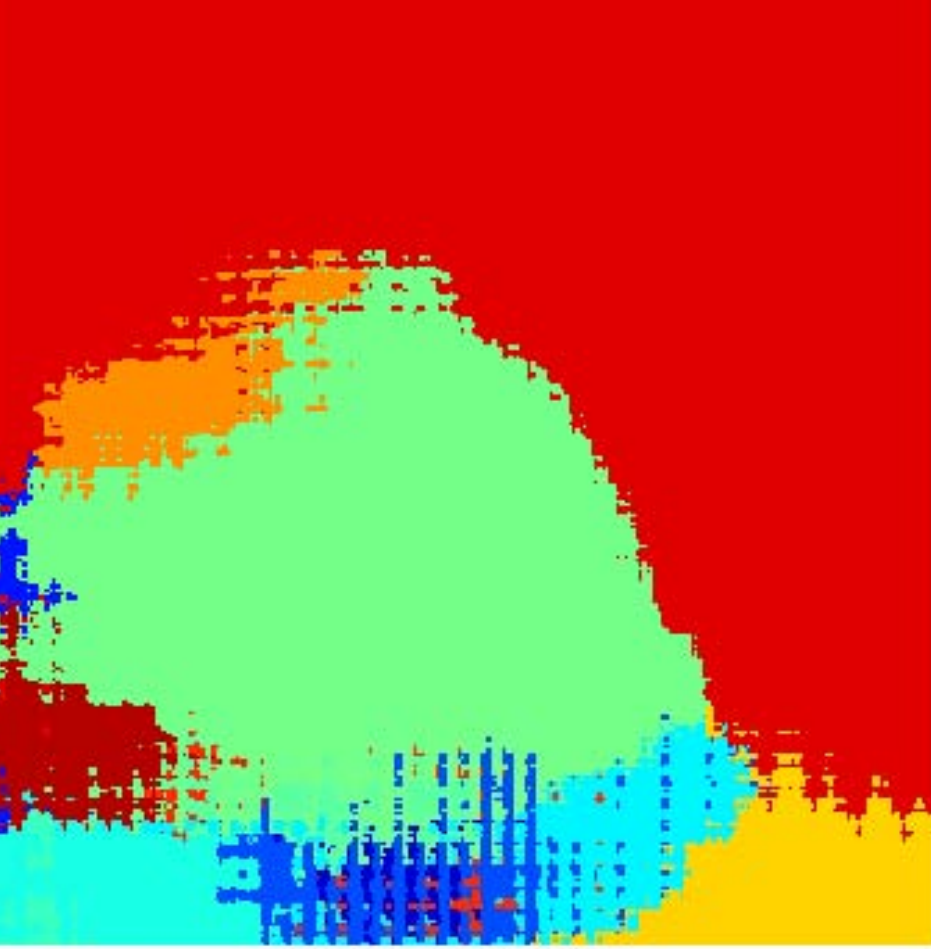}}
	& \imagetop{\includegraphics[width=0.19\linewidth]{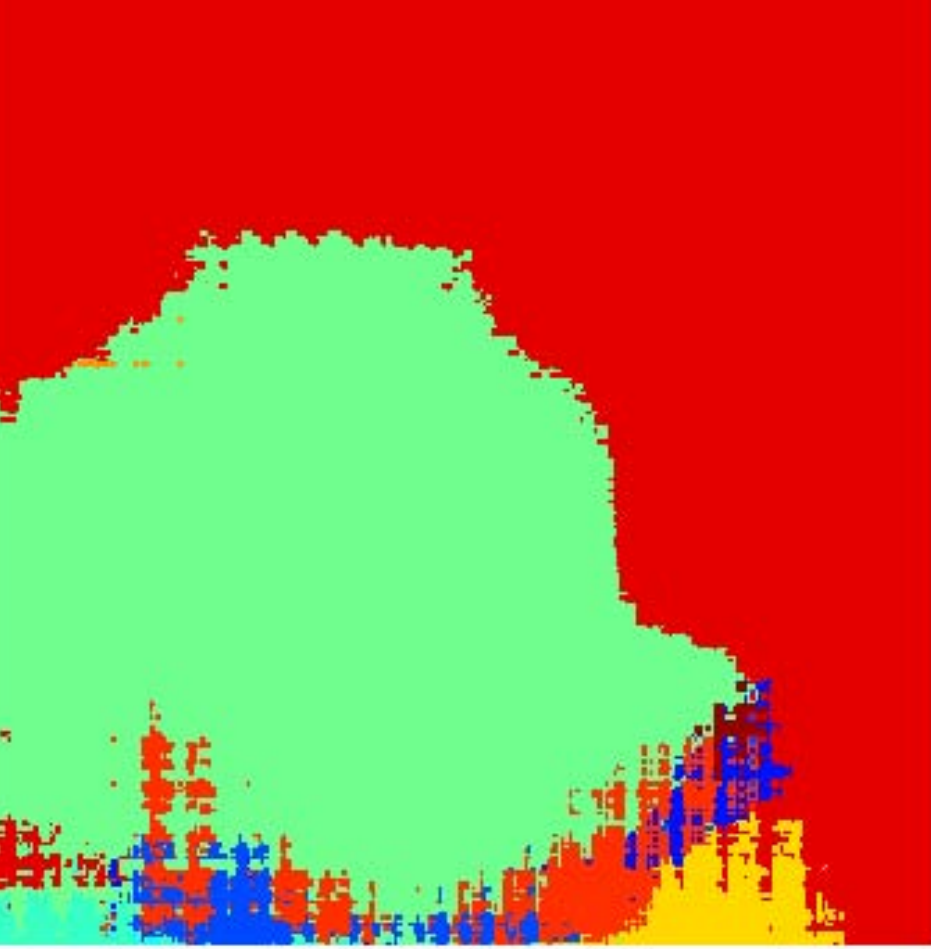}}
	& \imagetop{\includegraphics[width=0.07\linewidth]{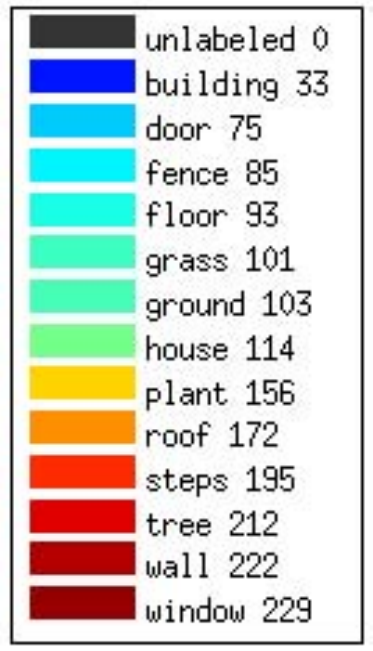}}	\\
	\vspace{-0.2cm}
& \imagetop{\includegraphics[width=0.19\linewidth]{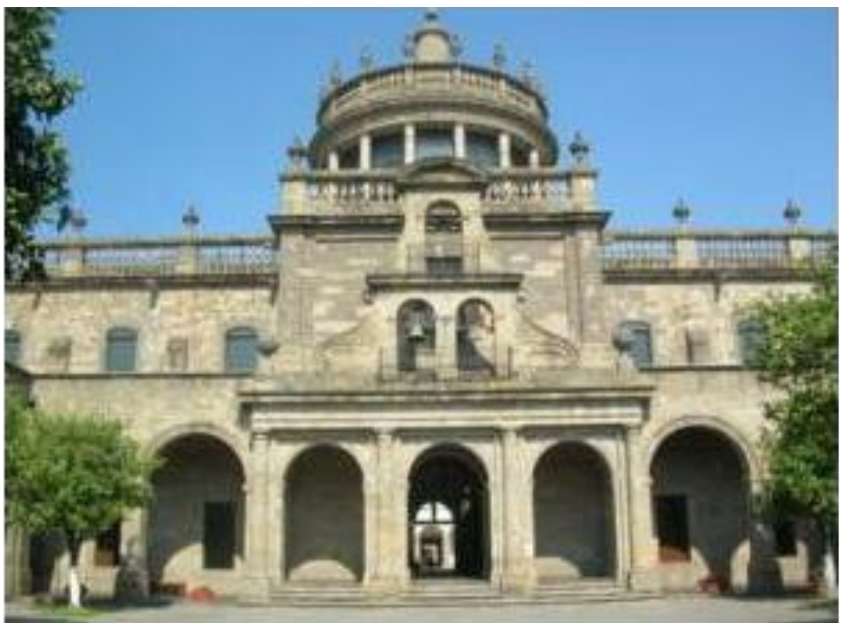}}
	& \imagetop{\includegraphics[width=0.19\linewidth]{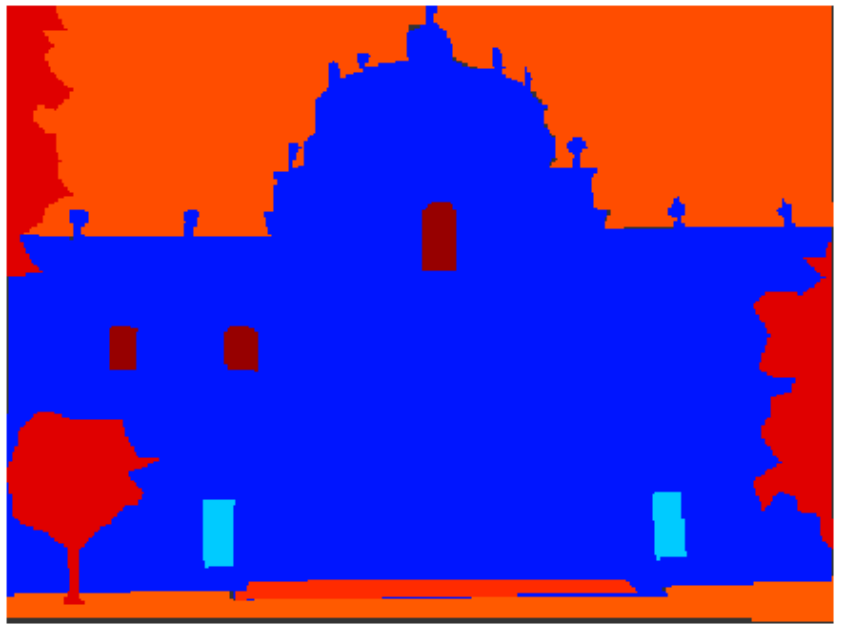}}
	& \imagetop{\includegraphics[width=0.19\linewidth]{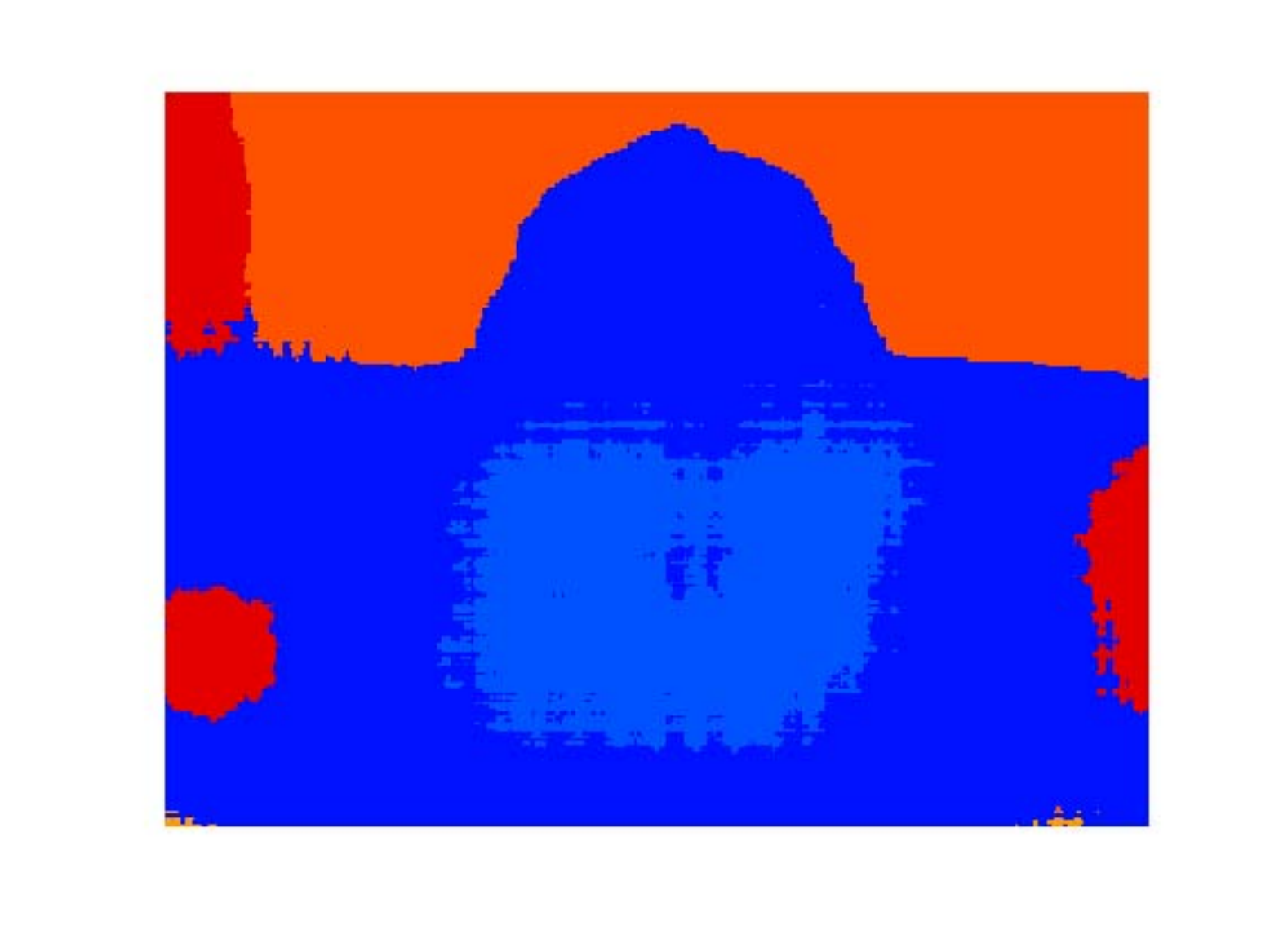}}
	& \imagetop{\includegraphics[width=0.19\linewidth]{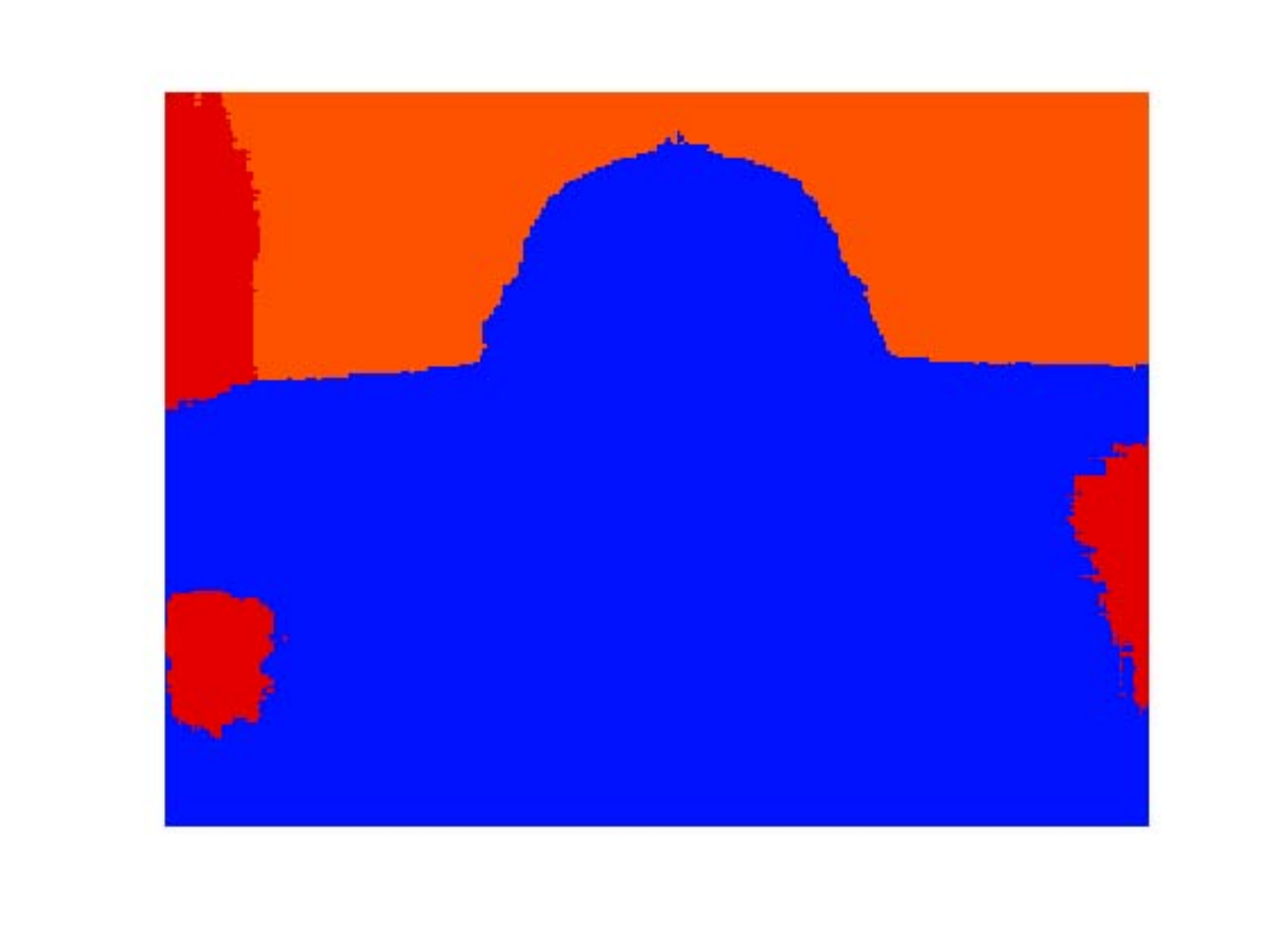}}
	& \imagetop{\includegraphics[width=0.07\linewidth]{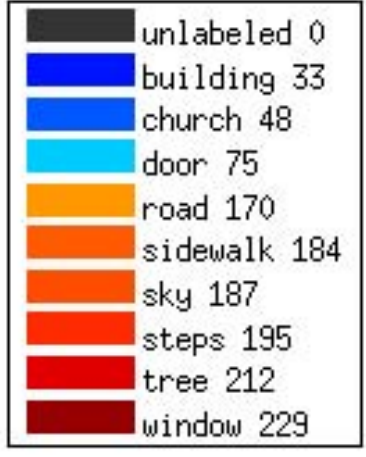}}	\\
	\vspace{-0.2cm}
&\imagetop{\includegraphics[width=0.19\linewidth]{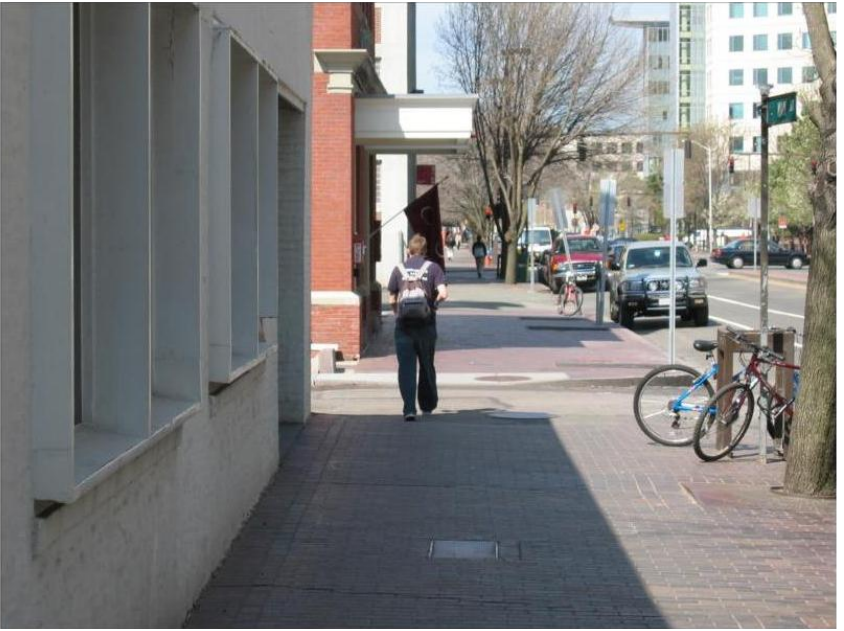}}
	& \imagetop{\includegraphics[width=0.19\linewidth]{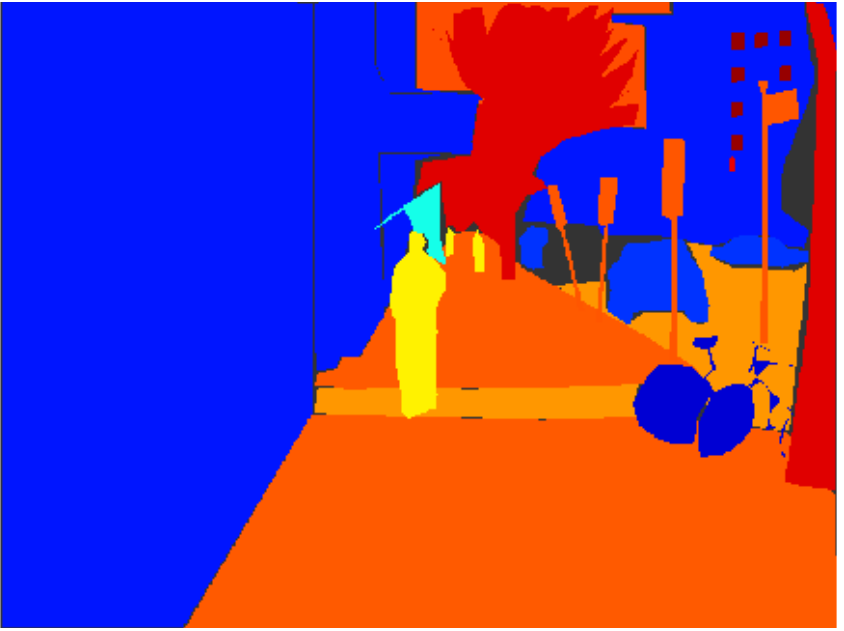}}
	& \imagetop{\includegraphics[width=0.19\linewidth]{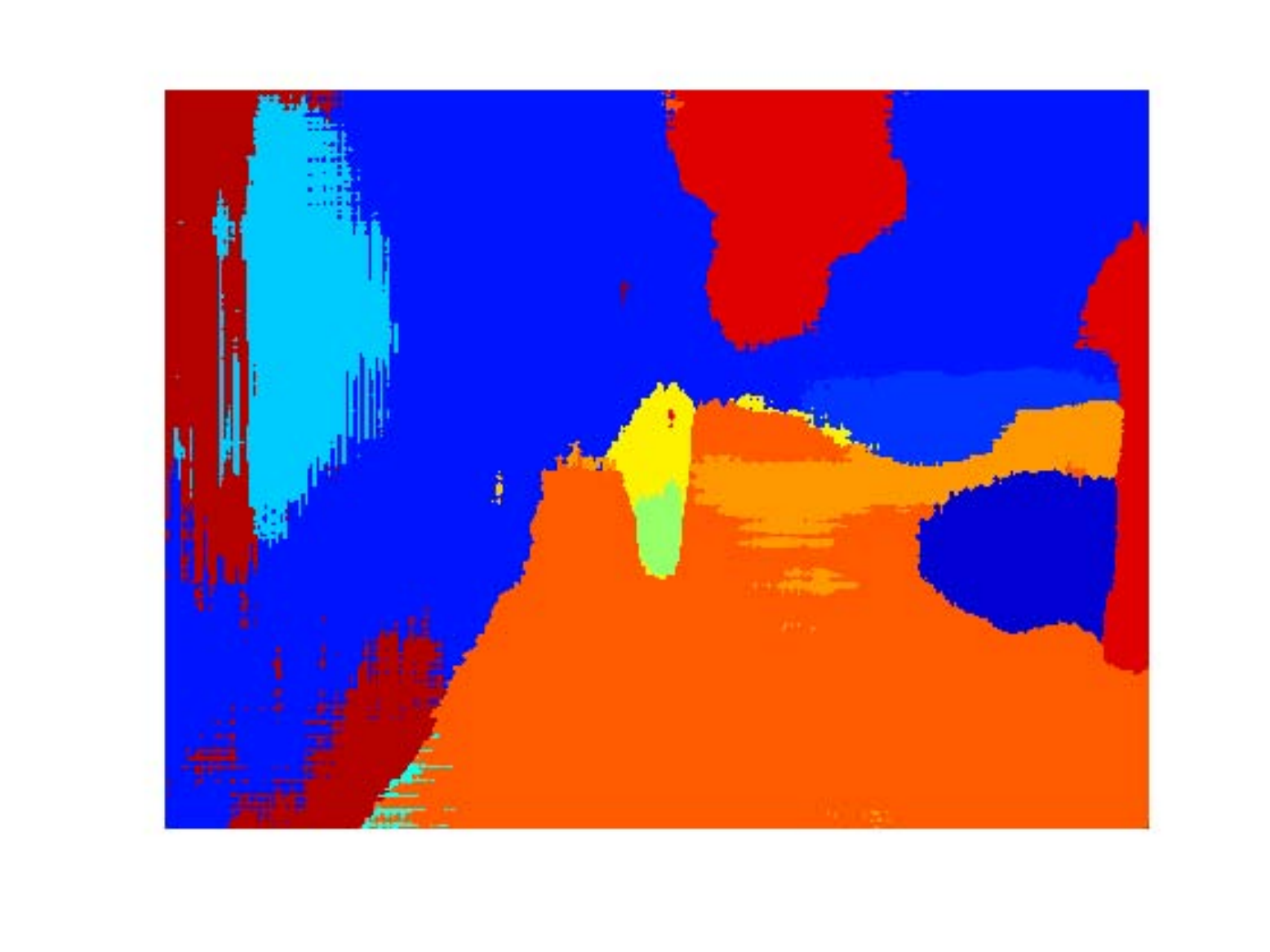}}
	& \imagetop{\includegraphics[width=0.19\linewidth]{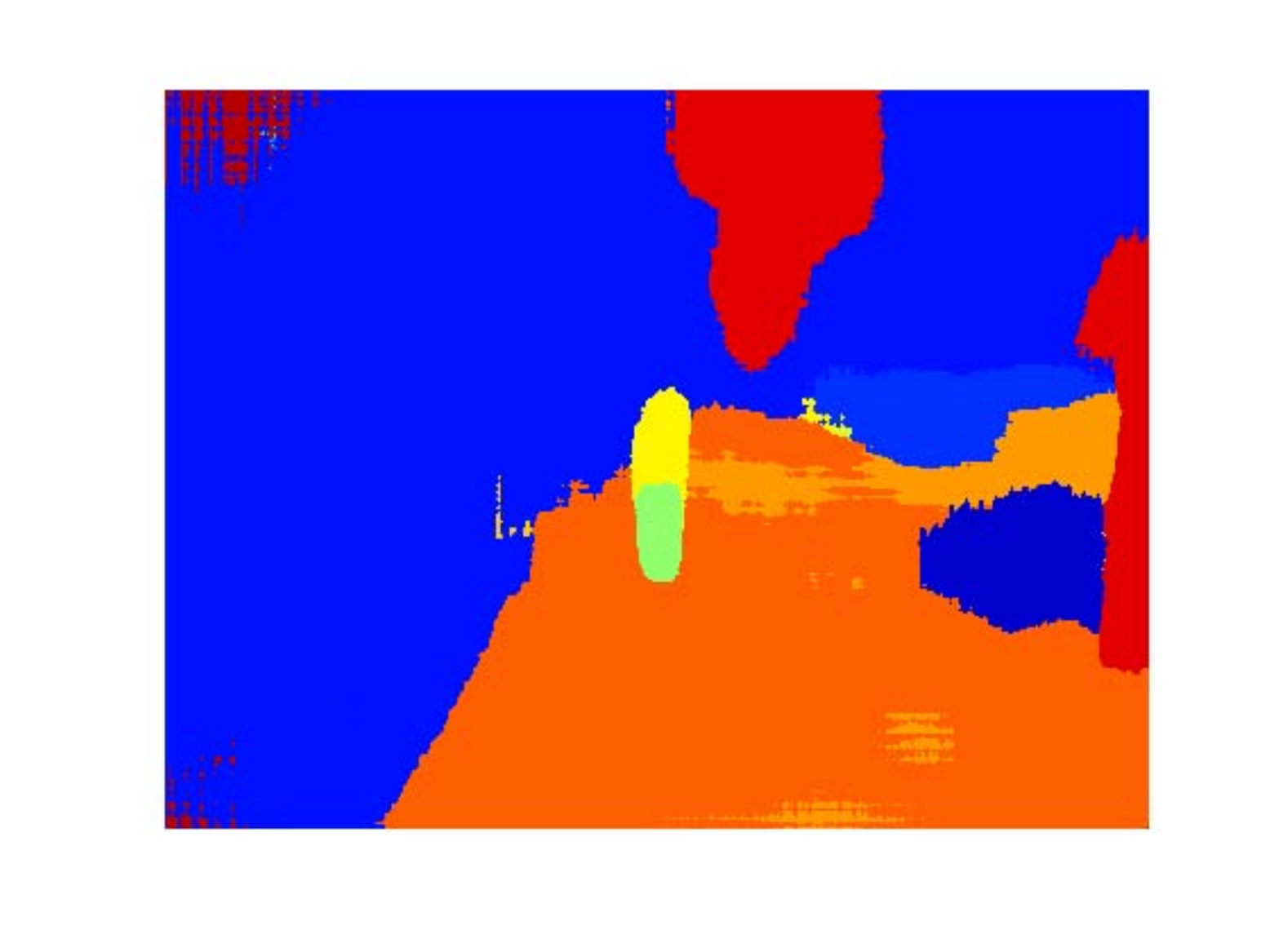}}
	& \imagetop{\includegraphics[width=0.07\linewidth]{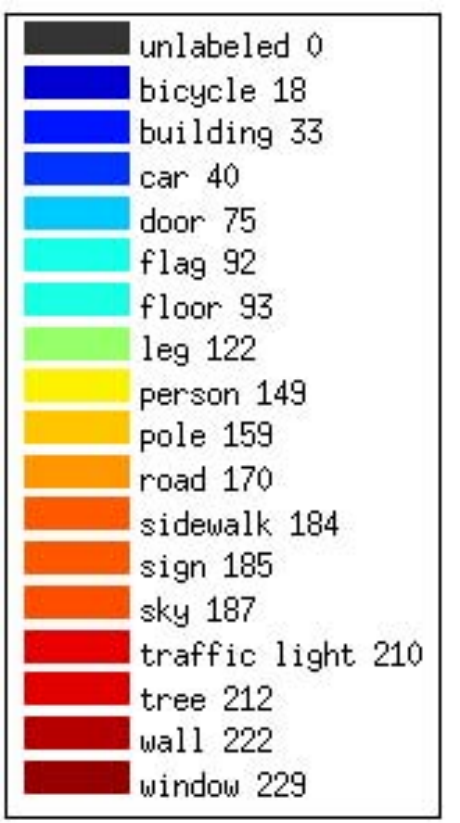}}	\\
	\vspace{-0.2cm}
	\end{tabular}
	\vspace{+2pt}
	\caption{Some example scene labeling results on the (top three rows) Barcelona and (bottom three rows) LM+Sun datasets by the proposed OverFeat+S+LC ($R=129$) and the OverFeat baseline models.  Best viewed in color.}\label{fig:result_lmsun}
\end{figure*}

We show our results on the Barcelona and LM+Sun datasets in Tables \ref{tbl:barcelona_result} and \ref{tbl:lmsun_result}. Images in neither datasets have meaningful scene names, so we only experimented on creating the label hierarchy with semantic context from label map statistics. The baseline model is chosen as the OverFeat model. On the Barcelona dataset, we first tried creating the label hierarchy with the original label maps, where about fewer than half of the pixels are unlabeled. This is denoted as OverFeat+S+LC (R=129) (Naive) in Table \ref{tbl:barcelona_result}. As expected, since the clustering result is not very reliable, the improvement is marginal. Then we used the baseline model to predict what the unlabeled pixels are and created a new label hierarchy with the new label maps. Both results of the sequential and hierarchical finetuning methods based on the new label hierarchy outperform the baseline model and the state-of-the-art methods by a large margin on the two datasets.  
Some qualitative results are shown in Fig. \ref{fig:result_lmsun}.

Although we cannot avoid errors in the new label maps introduced by the baseline model,  our method can still benefit a lot from finetuning on the subclasses. Two reasons may explain it. On the one side, we take the histogram of the patch for clustering. So the baseline model is only expected to find out what categories are there and how many pixels belong to each category, but it does not need to accurately localize the boundaries. On the other side, this can be viewed as utilizing the additional context information provided by the baseline model. With the training by either the sequential strategy or the  hierarchical one, the context information obtained from the baseline model is also incorporated into the learning process.

\begin{table}[t]
\centering
\begin{tabular}{|l|c|c|}
\hline
~~~~~~~~~~~~~~~~~~Method & Per-pixel & Per-class \\ \hline \hline
\cite{tighe2010superparsing}    & 0.669     & 0.076     \\
\cite{farabet2013learning}     & 0.678     & 0.095     \\
\cite{shuai2015dag} & 0.746 & \textbf{0.246} \\
\hline \hline
OverFeat baseline             & 0.742     & 0.165     \\
 OverFeat+S+LC ($R=129$) (Naive)  & 0.745 & 0.169 \\
 OverFeat+S+LC ($R=129$)   & \textbf{0.758}     & 0.172     \\
 OverFeat+H+LC ($R=129$)  & 0.756     & 0.19     \\ \hline
\end{tabular}
\caption{Per-pixel and per-class accuracies on the Barcelona dataset by different methods. (The best accuracy is marked in bold.) S = Sequantial strategy, H = Hierarchical loss and LC = Label map Clustering. OverFeat+S+LC ($R=129$) (Naive) is finetuned with subclasses directly created from the original label maps, while our other methods are finetuned with subclasses created from updated label maps. }\label{tbl:barcelona_result}
\end{table}

\begin{table}[t]
\centering
\begin{tabular}{|l|c|c|}
\hline
~~~~~~~~~~~~~~~~~~Method & Per-pixel & Per-class \\ \hline \hline
\cite{tighe2013finding}    & 0.549     & 0.071     \\
\cite{tighe2010superparsing_jnl}    & 0.651     & 0.152     \\
\cite{yang2014context}                  & 0.606     & 0.18     \\
\cite{george2015image} & 0.612 & 0.16\\
\hline \hline
OverFeat baseline             & 0.69     & 0.168     \\
 OverFeat+S+LC ($R=129$)  & 0.699 & \textbf{0.185} \\
  OverFeat+H+LC ($R=129$)   & \textbf{0.701}     & \textbf{0.185}     \\ \hline 
\end{tabular}
\caption{Per-pixel and per-class accuracies on the LM+Sun dataset by different methods. (The best accuracy entries are marked in bold. S = Sequantial strategy, H = Hierarchical loss and LC = Label map Clustering.}\label{tbl:lmsun_result}
\end{table}

\section{Discussion}
We would like to understand more why the accuracy of scene labeling can be improved by automatically creating label hierarchies from semantic context. The intuition is that one could  increase the appearance consistency within each subclass, which makes feature learning easier, such that the learned neurons can better detect meaningful visual patterns. 

One could also understand it as deep feature representations being better learned with richer prediction as supervisory signals. This has been verified by many successful cases of deep learning. 
In scene labeling, as the original classes are expanded to subclasses, the prediction task becomes more challenging. However, on the other hand, it also implies that the information carried by each training sample increases, since it requires extra semantic context (scene names and label map statistics) to obtain the ground-truth labels of subclasses. More challenging prediction can also reduce the overfitting problem which is often encountered in training neural networks.

\section{Conclusion}
In this paper, we exploited using semantic context in scene labeling to design novel supervisions for learning more discriminative deep feature representations. Two types of semantic context, scene names of images and label map statistics of image patches, were explored to create label hierarchies with subclasses. Such label hierarchies provide guided supervision for training deep models without the need of additional manual labeling. The lower intra-class variation between newly created subclasses decreases the ambiguity of training labels. Two training strategies, the sequential fine-tuning and hierarchical label fine-tuning strategies, are proposed to utilize the proposed label hierarchies for training CNNs. Extensive experiments on the SIFTFlow, Stanford background, Barcelona and LM+Sun datasets demonstrate the effectiveness of the proposed label hierarchy and corresponding training strategies.





\section*{Acknowledgement}
This work is supported by SenseTime Group Limited, Research Grants Council of Hong Kong (Project Number CUHK-14206114, CUHK14205615, CUHK14207814, CUHK14203015, CUHK417011 and CUHK14239816) and National Natural Science Foundation of China (61371192, 61301269).

\section*{Reference}

\bibliographystyle{elsarticle-harv} 
\bibliography{egbib}

\begin{thebibliography}{52}
\expandafter\ifx\csname natexlab\endcsname\relax\def\natexlab#1{#1}\fi
\expandafter\ifx\csname url\endcsname\relax
  \def\url#1{\texttt{#1}}\fi
\expandafter\ifx\csname urlprefix\endcsname\relax\def\urlprefix{URL }\fi

\bibitem[{Barinova et~al.(2010)Barinova, Lempitsky, Tretiak, and
  Kohli}]{barinova2010geometric}
Barinova, O., Lempitsky, V., Tretiak, E., Kohli, P., 2010. Geometric image
  parsing in man-made environments. In: European Conference on Computer Vision.

\bibitem[{Caesar et~al.(2016)Caesar, Uijlings, and Ferrari}]{caesar2016region}
Caesar, H., Uijlings, J., Ferrari, V., 2016. Region-based semantic segmentation
  with end-to-end training. In: European Conference on Computer Vision.
  Springer.

\bibitem[{Chatfield et~al.(2014)Chatfield, Simonyan, Vedaldi, and
  Zisserman}]{chatfieldreturn}
Chatfield, K., Simonyan, K., Vedaldi, A., Zisserman, A., 2014. Return of the
  devil in the details: Delving deep into convolutional nets.

\bibitem[{Chen et~al.(2015)Chen, Papandreou, Kokkinos, Murphy, and
  Yuille}]{chen2014semantic}
Chen, L.-C., Papandreou, G., Kokkinos, I., Murphy, K., Yuille, A.~L., 2015.
  Semantic image segmentation with deep convolutional nets and fully connected
  crfs. In: International Conference on Learning Representations.

\bibitem[{Dai et~al.(2015)Dai, He, and Sun}]{dai2015boxsup}
Dai, J., He, K., Sun, J., 2015. Boxsup: Exploiting bounding boxes to supervise
  convolutional networks for semantic segmentation. In: Proceedings of the IEEE
  International Conference on Computer Vision.

\bibitem[{Dauphin et~al.(2014)Dauphin, Pascanu, Gulcehre, Cho, Ganguli, and
  Bengio}]{dauphin2014identifying}
Dauphin, Y.~N., Pascanu, R., Gulcehre, C., Cho, K., Ganguli, S., Bengio, Y.,
  2014. Identifying and attacking the saddle point problem in high-dimensional
  non-convex optimization. In: Advances in Neural Information Processing
  Systems. pp. 2933--2941.

\bibitem[{Eigen and Fergus(2015)}]{eigen2015predicting}
Eigen, D., Fergus, R., 2015. Predicting depth, surface normals and semantic
  labels with a common multi-scale convolutional architecture. Proceedings of
  the IEEE International Conference on Computer Vision.

\bibitem[{Farabet et~al.(2013)Farabet, Couprie, Najman, and
  LeCun}]{farabet2013learning}
Farabet, C., Couprie, C., Najman, L., LeCun, Y., 2013. Learning hierarchical
  features for scene labeling. Pattern Analysis and Machine Intelligence, IEEE
  Transactions on, 1915--1929.

\bibitem[{George(2015)}]{george2015image}
George, M., 2015. Image parsing with a wide range of classes and scene-level
  context. In: Proceedings of the IEEE Conference on Computer Vision and
  Pattern Recognition. pp. 3622--3630.

\bibitem[{Girshick et~al.(2014)Girshick, Donahue, Darrell, and
  Malik}]{rirshick2014rich}
Girshick, R., Donahue, J., Darrell, T., Malik, J., 2014. Rich feature
  hierarchies for accurate object detection and semantic segmentation.

\bibitem[{Gould et~al.(2009)Gould, Fulton, and Koller}]{gould2009decomposing}
Gould, S., Fulton, R., Koller, D., 2009. Decomposing a scene into geometric and
  semantically consistent regions. In: International Conference on Computer
  Vision.

\bibitem[{Grangier et~al.(2009)Grangier, Bottou, and
  Collobert}]{grangier2009deep}
Grangier, D., Bottou, L., Collobert, R., 2009. Deep convolutional networks for
  scene parsing. In: International Conference on Machine Learning 2009 Deep
  Learning Workshop. Citeseer.

\bibitem[{He et~al.(2015{\natexlab{a}})He, Zhang, Ren, and Sun}]{he2015deep}
He, K., Zhang, X., Ren, S., Sun, J., 2015{\natexlab{a}}. Deep residual learning
  for image recognition. arXiv preprint arXiv:1512.03385.

\bibitem[{He et~al.(2015{\natexlab{b}})He, Zhang, Ren, and Sun}]{he2015delving}
He, K., Zhang, X., Ren, S., Sun, J., 2015{\natexlab{b}}. Delving deep into
  rectifiers: Surpassing human-level performance on imagenet classification.
  In: International Conference on Computer Vision.

\bibitem[{He et~al.(2004)He, Zemel, and Carreira-Perpindn}]{he2004multiscale}
He, X., Zemel, R.~S., Carreira-Perpindn, M., 2004. Multiscale conditional
  random fields for image labeling. In: Proceedings of the IEEE Conference on
  Computer Vision and Pattern Recognition.

\bibitem[{Ioffe and Szegedy(2015)}]{ioffe2015batch}
Ioffe, S., Szegedy, C., 2015. Batch normalization: Accelerating deep network
  training by reducing internal covariate shift. arXiv preprint
  arXiv:1502.03167.

\bibitem[{Joulin et~al.(2016)Joulin, van~der Maaten, Jabri, and
  Vasilache}]{joulin2015learning}
Joulin, A., van~der Maaten, L., Jabri, A., Vasilache, N., 2016. Learning visual
  features from large weakly supervised data. European Conference on Computer
  Vision.

\bibitem[{Ladicky et~al.(2010)Ladicky, Russell, Kohli, and
  Torr}]{ladicky2010graph}
Ladicky, L., Russell, C., Kohli, P., Torr, P.~H., 2010. Graph cut based
  inference with co-occurrence statistics. In: European Conference on Computer
  Vision.

\bibitem[{Lempitsky et~al.(2011)Lempitsky, Vedaldi, and
  Zisserman}]{lempitsky2011pylon}
Lempitsky, V., Vedaldi, A., Zisserman, A., 2011. Pylon model for semantic
  segmentation. In: Advances in Neural Information Processing Systems.

\bibitem[{Li et~al.(2014)Li, Zhao, and Wang}]{li2014highly}
Li, H., Zhao, R., Wang, X., 2014. Highly efficient forward and backward
  propagation of convolutional neural networks for pixelwise classification.
  arXiv preprint arXiv:1412.4526.

\bibitem[{Lim et~al.(2013)Lim, Zitnick, and Doll{\'a}r}]{lim2013sketch}
Lim, J.~J., Zitnick, C.~L., Doll{\'a}r, P., 2013. Sketch tokens: A learned
  mid-level representation for contour and object detection. In: Proceedings of
  the IEEE Conference on Computer Vision and Pattern Recognition. pp.
  3158--3165.

\bibitem[{Liu et~al.(2011)Liu, Yuen, and Torralba}]{liu2011nonparametric}
Liu, C., Yuen, J., Torralba, A., 2011. Nonparametric scene parsing via label
  transfer. Pattern Analysis and Machine Intelligence, IEEE Transactions on
  33~(12), 2368--2382.

\bibitem[{Liu et~al.(2008)Liu, Yuen, Torralba, Sivic, and
  Freeman}]{liu2008sift}
Liu, C., Yuen, J., Torralba, A., Sivic, J., Freeman, W.~T., 2008. Sift flow:
  Dense correspondence across different scenes. In: European Conference on
  Computer Vision.

\bibitem[{Liu et~al.(2015)Liu, Li, Luo, Loy, and Tang}]{liu2015semantic}
Liu, Z., Li, X., Luo, P., Loy, C.-C., Tang, X., 2015. Semantic image
  segmentation via deep parsing network. In: Proceedings of the IEEE
  International Conference on Computer Vision.

\bibitem[{Long et~al.(2014)Long, E.Shelhamer, and Darrell}]{long2014fcn}
Long, J., E.Shelhamer, Darrell, T., 2014. Fully convolutional networks for
  semantic segmentation. Proceedings of the IEEE Conference on Computer Vision
  and Pattern Recognition.

\bibitem[{Martens and Sutskever(2012)}]{martens2012training}
Martens, J., Sutskever, I., 2012. Training deep and recurrent networks with
  hessian-free optimization. In: Neural Networks: Tricks of the Trade.
  Springer, pp. 479--535.

\bibitem[{Noh et~al.(2015)Noh, Hong, and Han}]{noh2015learning}
Noh, H., Hong, S., Han, B., 2015. Learning deconvolution network for semantic
  segmentation. In: Proceedings of the IEEE International Conference on
  Computer Vision.

\bibitem[{Pinheiro and Collobert(2014)}]{pinheiro2013recurrent}
Pinheiro, P. H.~O., Collobert, R., 2014. Recurrent convolutional neural
  networks for scene labeling. In: International Conference on Machine
  Learning.

\bibitem[{Qi et~al.(2015)Qi, Shi, Liu, Liao, and Jia}]{qi2015semantic}
Qi, X., Shi, J., Liu, S., Liao, R., Jia, J., 2015. Semantic segmentation with
  object clique potential. In: Proceedings of the IEEE International Conference
  on Computer Vision.

\bibitem[{Sermanet et~al.(2013)Sermanet, Eigen, Zhang, Mathieu, Fergus, and
  LeCun}]{sermanet2013overfeat}
Sermanet, P., Eigen, D., Zhang, X., Mathieu, M., Fergus, R., LeCun, Y., 2013.
  Overfeat: Integrated recognition, localization and detection using
  convolutional networks. International Conference on Learning Representations.

\bibitem[{Sharma et~al.(2015)Sharma, Tuzel, and Jacobs}]{sharma2015deep}
Sharma, A., Tuzel, O., Jacobs, D.~W., 2015. Deep hierarchical parsing for
  semantic segmentation. Proceedings of the IEEE Conference on Computer Vision
  and Pattern Recognition.

\bibitem[{Shotton et~al.(2006)Shotton, Winn, Rother, and
  Criminisi}]{shotton2006textonboost}
Shotton, J., Winn, J., Rother, C., Criminisi, A., 2006. Textonboost: Joint
  appearance, shape and context modeling for multi-class object recognition and
  segmentation. In: European Conference on Computer Vision.

\bibitem[{Shuai et~al.(2016)Shuai, Zuo, Wang, and Wang}]{shuai2015dag}
Shuai, B., Zuo, Z., Wang, G., Wang, B., 2016. Dag-recurrent neural networks for
  scene labeling. Proceedings of the IEEE Conference on Computer Vision and
  Pattern Recognition.

\bibitem[{Singh and Kosecka(2013)}]{singh2013nonparametric}
Singh, G., Kosecka, J., 2013. Nonparametric scene parsing with adaptive feature
  relevance and semantic context. In: Proceedings of the IEEE Conference on
  Computer Vision and Pattern Recognition.

\bibitem[{Socher et~al.(2011)Socher, Lin, Manning, and Ng}]{socher2011parsing}
Socher, R., Lin, C.~C., Manning, C., Ng, A.~Y., 2011. Parsing natural scenes
  and natural language with recursive neural networks. In: International
  Conference on Machine Learning.

\bibitem[{Srivastava et~al.(2014)Srivastava, Hinton, Krizhevsky, Sutskever, and
  Salakhutdinov}]{srivastava2014dropout}
Srivastava, N., Hinton, G., Krizhevsky, A., Sutskever, I., Salakhutdinov, R.,
  2014. Dropout: A simple way to prevent neural networks from overfitting. The
  Journal of Machine Learning Research 15~(1), 1929--1958.

\bibitem[{Srivastava et~al.(2015)Srivastava, Greff, and
  Schmidhuber}]{srivastava2015training}
Srivastava, R.~K., Greff, K., Schmidhuber, J., 2015. Training very deep
  networks. In: Advances in Neural Information Processing Systems. pp.
  2368--2376.

\bibitem[{Sutskever et~al.(2013)Sutskever, Martens, Dahl, and
  Hinton}]{sutskever2013importance}
Sutskever, I., Martens, J., Dahl, G., Hinton, G., 2013. On the importance of
  initialization and momentum in deep learning. In: International Conference on
  Machine Learning. pp. 1139--1147.

\bibitem[{Szegedy et~al.(2015)Szegedy, Liu, Jia, Sermanet, Reed, Anguelov,
  Erhan, Vanhoucke, and Rabinovich}]{szegedy2014going}
Szegedy, C., Liu, W., Jia, Y., Sermanet, P., Reed, S., Anguelov, D., Erhan, D.,
  Vanhoucke, V., Rabinovich, A., 2015. Going deeper with convolutions. In:
  Proceedings of the IEEE Conference on Computer Vision and Pattern
  Recognition.

\bibitem[{Tighe and Lazebnik(2010)}]{tighe2010superparsing}
Tighe, J., Lazebnik, S., 2010. Superparsing: scalable nonparametric image
  parsing with superpixels. In: European Conference on Computer Vision.

\bibitem[{Tighe and Lazebnik(2012)}]{tighe2010superparsing_jnl}
Tighe, J., Lazebnik, S., 2012. Superparsing: scalable nonparametric image
  parsing with superpixels. International Journal of Computer Vision.

\bibitem[{Tighe and Lazebnik(2013)}]{tighe2013finding}
Tighe, J., Lazebnik, S., 2013. Finding things: Image parsing with regions and
  per-exemplar detectors. In: Proceedings of the IEEE Conference on Computer
  Vision and Pattern Recognition. pp. 3001--3008.

\bibitem[{Vedaldi and Fulkerson(2008)}]{vedaldi08vlfeat}
Vedaldi, A., Fulkerson, B., 2008. {VLFeat}: An open and portable library of
  computer vision algorithms. http://www.vlfeat.org/.

\bibitem[{Wang et~al.(2016)Wang, Li, Ouyang, and Wang}]{wang2016learnable}
Wang, Z., Li, H., Ouyang, W., Wang, X., 2016. Learnable histogram: Statistical
  context features for deep neural networks. In: European Conference on
  Computer Vision. Springer.

\bibitem[{Xiao et~al.(2015)Xiao, Xia, Yang, Huang, and Wang}]{xiao2015learning}
Xiao, T., Xia, T., Yang, Y., Huang, C., Wang, X., 2015. Learning from massive
  noisy labeled data for image classification. In: Proceedings of the IEEE
  Conference on Computer Vision and Pattern Recognition. pp. 2691--2699.

\bibitem[{Xie et~al.(2015)Xie, Yang, Wang, and Lin}]{xie2015hyper}
Xie, S., Yang, T., Wang, X., Lin, Y., 2015. Hyper-class augmented and
  regularized deep learning for fine-grained image classification. In:
  Proceedings of the IEEE Conference on Computer Vision and Pattern
  Recognition. pp. 2645--2654.

\bibitem[{Yan et~al.(2015)Yan, Zhang, Piramuthu, Jagadeesh, DeCoste, Di, and
  Yu}]{yan2015hd}
Yan, Z., Zhang, H., Piramuthu, R., Jagadeesh, V., DeCoste, D., Di, W., Yu, Y.,
  2015. Hd-cnn: hierarchical deep convolutional neural networks for large scale
  visual recognition. In: Proceedings of the IEEE International Conference on
  Computer Vision. pp. 2740--2748.

\bibitem[{Yang et~al.(2014)Yang, Price, Cohen, and Yang}]{yang2014context}
Yang, J., Price, B., Cohen, S., Yang, M.-H., 2014. Context driven scene parsing
  with attention to rare classes. In: Proceedings of the IEEE Conference on
  Computer Vision and Pattern Recognition.

\bibitem[{Yao et~al.(2012)Yao, Fidler, and Urtasun}]{yao2012describing}
Yao, J., Fidler, S., Urtasun, R., 2012. Describing the scene as a whole: Joint
  object detection. Proceedings of the IEEE Conference on Computer Vision and
  Pattern Recognition.

\bibitem[{Yosinski et~al.(2014)Yosinski, Clune, Bengio, and
  Lipson}]{yosinski2014transferable}
Yosinski, J., Clune, J., Bengio, Y., Lipson, H., 2014. How transferable are
  features in deep neural networks? In: Advances in Neural Information
  Processing Systems.

\bibitem[{Zeiler and Fergus(2014)}]{zeiler2014visualizing}
Zeiler, M.~D., Fergus, R., 2014. Visualizing and understanding convolutional
  networks. In: European Conference on Computer Vision.

\bibitem[{Zheng et~al.(2015)Zheng, Jayasumana, Romera-Paredes, Vineet, Su, Du,
  Huang, and Torr}]{zheng2015conditional}
Zheng, S., Jayasumana, S., Romera-Paredes, B., Vineet, V., Su, Z., Du, D.,
  Huang, C., Torr, P., 2015. Conditional random fields as recurrent neural
  networks. In: Proceedings of the IEEE International Conference on Computer
  Vision.

\end{thebibliography}





\end{document}